\documentclass[10pt,twocolumn,letterpaper]{article}
\pdfoutput=1
\usepackage{cvpr}
\usepackage{times}
\usepackage{epsfig}
\usepackage{graphicx}
\usepackage{amsmath}
\usepackage{amssymb}
\usepackage{caption}
\usepackage{subfig}
\usepackage{bm}
\usepackage{bbold}
\usepackage{tabls}
\usepackage{enumitem}
\usepackage{gensymb}

\usepackage[pagebackref=true,breaklinks=true,letterpaper=true,colorlinks,bookmarks=false]{hyperref}

\cvprfinalcopy 

\ifcvprfinal\fi
\begin{document}

\title{An Adversarial Neuro-Tensorial Approach For Learning Disentangled Representations}

\author{
Mengjiao Wang$^1$ \hspace{1pt} Zhixin Shu$^2$ \hspace{1pt} Shiyang Cheng$^1$ \hspace{1pt} Yannis Panagakis$^1$ \hspace{1pt} Dimitris Samaras$^2$ \hspace{1pt} Stefanos Zafeiriou$^1$ 
\\
$^1$ Imperial College London \quad
$^2$ Stony Brook University 
\\
{\tt\small $^1$ \{m.wang15,shiyang.cheng11,i.panagakis,s.zafeiriou\}@imperial.ac.uk}
{\tt\small $^2$ \{zhshu,samaras\}@cs.stonybrook.edu}
}

\maketitle

\begin{abstract}

Several factors contribute to the appearance of an object in a visual scene, including pose, illumination, and deformation, among others. Each factor accounts for a source of variability in the data, while the multiplicative interactions of these factors emulate the entangled variability, giving rise to the rich structure of visual object appearance.
Disentangling such unobserved factors from visual data is a challenging task, especially when the data have been captured in uncontrolled recording conditions (also referred to as ``in-the-wild") and label information is not available. 

In this paper, we propose the first unsupervised deep learning method (with pseudo-supervision) for disentangling multiple latent factors of variation in face images captured in-the-wild. To this end, we propose a deep latent variable model, where the multiplicative interactions of multiple latent factors of variation are explicitly modelled by means of multilinear (tensor) structure.
We demonstrate that the proposed approach indeed learns disentangled representations of  facial expressions and pose, which can be used in various applications, including face editing,
as well as 3D face reconstruction and classification of facial expression, identity and pose.

\end{abstract}
\vspace{-6mm}
\section{Introduction}
\label{intro}
\begin{figure*}[!thb]
\captionsetup[subfigure]{labelformat=empty, justification=centering,position=top}
\subfloat[Expression Editing]{\includegraphics[width=0.5\linewidth]{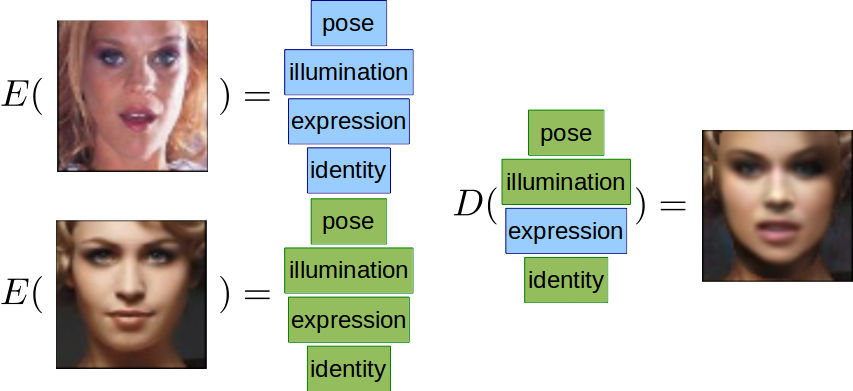}}
\subfloat[Pose Editing]{\includegraphics[width=0.5\linewidth]{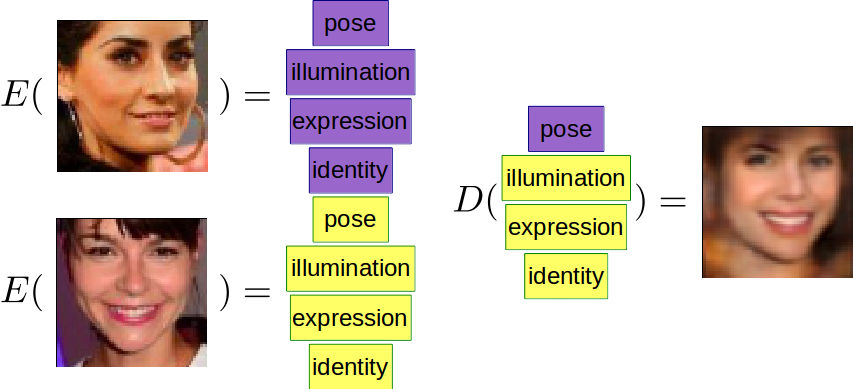}}
\caption{Given a single in-the-wild image, our network learns disentangled representations for pose, illumination, expression and identity. Using these representations, we are able to manipulate the image and edit the pose or expression.}
\label{editing}
\vspace{-10pt}
\end{figure*}

The appearance of visual objects is significantly
affected by multiple factors of variability such as, for example, pose, illumination, identity, and expression in case of faces. Each factor accounts for a source of variability in the data, while their complex interactions give rise to the observed entangled variability. Discovering the modes of variation, or in other words disentangling the latent factors of variations in visual data, is a very important problem in the intersection of statistics, machine learning, and computer vision. 

Factor analysis~\cite{fabrigar2011exploratory} and the closely related Principal Component Analysis (PCA)~\cite{hotelling1933analysis}  are probably  the most popular  statistical methods that find a single mode of variation explaining the data. Nevertheless, visual appearance (e.g., facial appearance) is affected by several  modes of variations. Hence, methods such as PCA are not able to identify such multiple factors of variation. For example, when PCA is applied to facial images, the first 
principal component captures  both pose and expressions variations.

An early approach for learning different modes of variation in the data is TensorFaces \cite{vasilescu2002multilinear}. In particular,  TensorFaces is a strictly supervised method as it not only requires the facial data to be labelled (e.g., in terms of expression, identity, illumination etc.) but the data tensor must also contain all samples in all different variations. This is the primary reason that the use of such tensor decompositions is still limited to databases that have been captured in a strictly controlled environment, such as the Weizmann face database \cite{vasilescu2002multilinear}. 

Recent unsupervised tensor decompositions methods \cite{tang2013tensor,wang2017learning}  automatically discover the modes of variation in unlabelled data. In particular, the most recent one \cite{wang2017learning} assumes that the original visual data have been produced by a hidden multilinear structure and the aim of the unsupervised tensor decomposition is to discover both the underlying multilinear structure, as well as the corresponding weights (coefficients) that best explain the data. Special instances of the unsupervised tensor decomposition are the Shape-from-Shading (SfS) decompositions in \cite{kemelmacher2013internet,snape2015automatic} and the multilinear decompositions for 3D face description in \cite{wang2017learning}. In \cite{wang2017learning}, it is shown that the method indeed can be used to learn representations where many modes of variation have been disentangled (e.g., identity, expression and illumination etc.). Nevertheless, the method in \cite{wang2017learning} is not able to find pose variations and bypasses this problem by applying it to faces which have been frontalised by applying a warping function (e.g., piece-wise affine warping \cite{matthews2004active}). 

Another promising line of research for  discovering  latent representations is  unsupervised Deep Neural Networks (DNNs). Unsupervised DNNs architectures include the Auto-Encoders (AE) \cite{bengio2013representation}, as well as the Generative Adversarial Networks (GANs) \cite{goodfellow2014generative} or adversarial versions of AE, e.g., the Adversarial Auto-Encoders (AAE)  \cite{makhzani2015adversarial}. Even though GANs, as well as AAEs, provide very elegant frameworks for discovering powerful low-dimensional embeddings without having to align the faces, due to the complexity of the networks, unavoidably all modes of variation are multiplexed in the latent-representation. Only with the use of labels it is possible to model/learn the manifold over the latent representation, usually as a post-processing step \cite{shu2017neural}. 

In this paper, we show that it is possible to learn a disentangled representation of the human face captured in arbitrary recording conditions in an unsupervised manner\footnote{Our methodology uses the information produced by an automatic 3D face fitting procedure~\cite{booth20173d} but it does not make use of any labels in the training set.} by imposing a multilinear structure on the latent representation of an AAE \cite{shu2017neural}. To the best of our knowledge, this is the first time that unsupervised tensor decompositions have been combined with DNNs for learning disentangled representations.  We demonstrate the power of the proposed approach by showing expression/pose transfer using only the latent variable that is related to expression/pose. We also demonstrate that the disentangled low-dimensional embeddings are useful for many other applications, such as facial expression, pose, and identity recognition and clustering.  An example of the proposed approach is given in Fig.~\ref{editing}. In particular, the left pair of images have been decomposed, using the encoder of the proposed neural network $E(\cdot)$, into many different latent representations including latent representations for pose, illumination, identity and expression. Since our framework has learned a disentangled representation we can easily transfer the expression by only changing the latent variable related to expression and passing the latent vector into the decoder of our neural network $D(\cdot)$. Similarly, we can transfer the pose merely by changing the latent variable related to pose. 

\section{Related Work}
\begin{figure*}[!thb]
\centering
\includegraphics[width=0.919\linewidth]{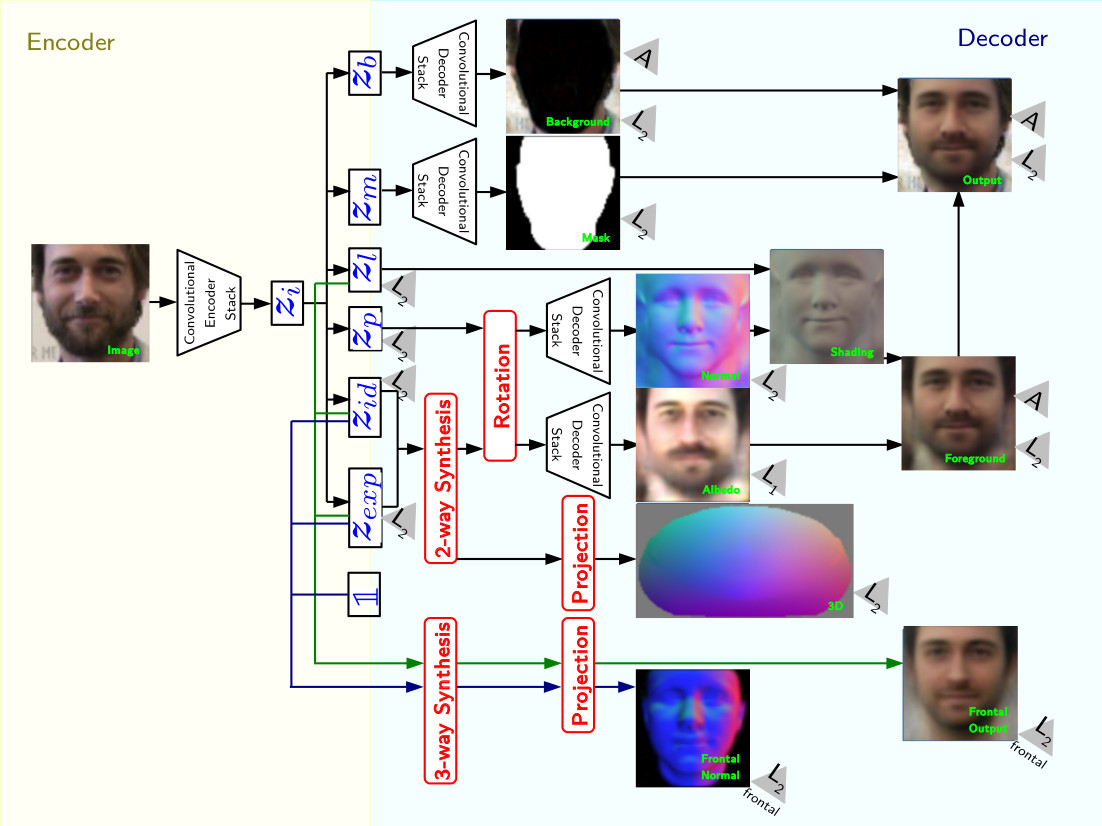}
\caption{Our network is an end-to-end trained auto-encoder. The encoder $E$ extracts latent variables corresponding to illumination, pose, expression and identity from the input image $\bm{x}$. These latent variables are then fed into the decoder $D$ to reconstruct the image. We impose a multilinear structure and enforce the disentangling of variations. The grey triangles represent the losses:  adversarial loss $A$, $L_1$ and $L_2$ losses. }
\label{network}
\vspace{-10pt}
\end{figure*}

Learning disentangled representations that explain multiple factors of variation in the data as disjoint latent dimensions is desirable in several machine learning, computer vision, and graphics tasks.

Indeed, bilinear factor analysis models~\cite{Tenenbaum:2000:SSC:1121517.1121518} have been employed for disentangling two factors of variation (e.g., head pose and facial identity) in the data.  Identity, expression, pose, and illumination variations are disentangled in~\cite{vasilescu2002multilinear} by applying Tucker decomposition (also known as multilinear Singular Value Decomposition (SVD) \cite{de2000multilinear}) into a carefully constructed tensor  through label information. Interestingly, the modes of variation in well aligned images can be  recovered via a multilinear matrix factorization ~\cite{wang2017learning} without any supervision. However, inference in \cite{wang2017learning}  might be ill-posed. 

More recently, both supervised and unsupervised deep learning methods have been developed for disentangled representations learning. Transforming auto-encoders~\cite{hinton2011transforming} is among the earliest methods for disentangling latent factors by means of auto-encoder capsules. In~\cite{desjardins2012disentangling} hidden factors of variation are disentangled via inference in a variant of the restricted Boltzmann machine. Disentangled representations of input images are obtained by the hidden layers of deep networks in~\cite{cheung2014discovering} and through a higher-order Boltzmann machine in~\cite{pmlr-v32-reed14}. The Deep Convolutional Inverse Graphics Network~\cite{kulkarni2015deep} learns a representation that is disentangled with respect to transformations such as out-of-plane rotations and lighting variations. Methods in \cite{chen2016infogan,mathieu2016disentangling,ijcai2017-404,tewari17MoFA,tran2017disentangled} extract disentangled and interpretable visual representations by employing adversarial training. The method in \cite{shu2017neural} disentangles the latent representations of illumination, surface normals, and albedo of face images using an image rendering pipeline. Trained with pseudo-supervision, \cite{shu2017neural} undertakes multiple image editing tasks by manipulating the relevant latent representations. Nonetheless, this editing approach still requires expression labelling, as well as sufficient sampling of a specific expression. 

Here, the proposed network is able to edit the expression of a face image given another single in-the-wild face image of arbitrary expression. Furthermore, we are able to edit the pose of a face in the image which is not possible in~\cite{shu2017neural}.

\section{Proposed Method}

In this section, we will introduce the main multilinear models used to describe three different image modalities, namely texture, 3D shape and 3D surface normals. To this end, we assume that for each different modality there is a different core tensor but all modalities share the same latent representation of weights regarding identity and expression. During training all the core tensors inside the network are randomly initialised and learnt end-to-end. In the following, we assume that we have a set of $n$ facial images (e.g., in the training batch) and their corresponding 3D facial shape, as well as their normals per pixel (the 3D shape and normals have been produced by fitting a 3D model on the 2D image, e.g.,~\cite{booth20173d}).  

\subsection{Facial Texture}

The main assumption here follows from ~\cite{wang2017learning}. That is, the rich structure of visual data is a result of multiplicative interactions of hidden (latent) factors and hence  the underlying multilinear structure, as well as the corresponding weights (coefficients) that best explain the data can be recovered using the unsupervised tensor decomposition \cite{wang2017learning}.
Indeed, following~\cite{wang2017learning},  disentangled representations can be learnt (e.g., identity, expression, and illumination, etc.) from frontalised facial images. The frontalisation process is performed by applying a piecewise affine transform using the sparse shape recovered by a face alignment process. Inevitably, this process suffers from warping artifacts. Therefore, rather than applying any warping process, we perform the multilinear decomposition only on near frontal faces, which can be automatically detected during the 3D face fitting stage. In particular, assuming a near frontal facial image rasterised in a vector $\bm{x}_{f} \in \mathbb{R}^{k_{x} \times 1}$, given a core tensor \vspace{0.1pt} $\mathcal{Q} \in \mathbb{R}^{k_x \times k_l \times k_{exp} \times k_{id}}$ \footnote{
Tensors notation: Tensors (i.e., multidimensional arrays) are 
and denoted by calligraphic letters, e.g., $\bm{\mathcal{X}}$. The \textit{mode-$m$ matricisation} of a tensor $\bm{\mathcal{X}} \in
\mathbb{R}^{I_1 \times I_2 \times \cdots \times I_M}$ maps
$\bm{\mathcal{X}}$ to a matrix $\mathbf{X}_{(m)} \in \mathbb{R}^{I_{m}
\times \bar{I}_{m}}$.  The \textit{mode-$m$ vector product} of a tensor  $\bm{\mathcal{X}} \in
\mathbb{R}^{I_{1}\times I_{2}\times \ldots \times I_{M}}$ with a
vector $\mathbf{x} \in \mathbb{R}^{I_m}$, denoted by
$\bm{\mathcal{X}} \times_{n} \mathbf{x} \in \mathbb{R}^{I_{1}\times
I_{2}\times\cdots\times I_{n-1}  \times I_{n+1} \times
\cdots \times I_{N}} $.

The \textit{Kronecker product} is denoted by $\otimes$ and the \textit{Khatri-Rao} (i.e., column-wise Kronecker product) product is denoted by $\odot$. More details on tensors and multilinear operators can be found in \cite{Kolda2008}.}, this can be decomposed as \vspace{0.1pt}
\begin{equation}
\begin{aligned}
\bm{x}_{f} &= \mathcal{Q} \times_2\bm{z}_{l} \times_3 \bm{z}_{exp} \times_4 \bm{z}_{id},
\end{aligned}
\end{equation}
where $\bm{z}_{l} \in \mathbb{R}^{k_l}, \bm{z}_{exp} \in \mathbb{R}^{k_{exp}}$ and $\bm{z}_{id} \in \mathbb{R}^{k_{id}}$ are the weights that correspond to illumination, expression and identity respectively. The equivalent form in case that we have a number of images in the batch stacked in the columns of a matrix $\mathbf{X}_{f} \in \mathbb{R}^{k_x \times n}$ is 
\begin{equation}
\bm{X}_{f} = \bm{Q}_{(1)} (\bm{Z}_{l} \odot \bm{Z}_{exp} \odot \bm{Z}_{id}), 
\label{eq:xf}
\end{equation}
where $\bm{Q}_{(1)}$ is a mode-1 matricisation of tensor $\mathcal{Q}$  and $\bm{Z}_{l}$, $\bm{Z}_{exp}$ and $\bm{Z}_{id}$
are the corresponding matrices that gather the weights of the decomposition for all images in the batch. That is, $\bm{Z}_{exp} \in \mathbb{R}^{k_{exp} \times n}$ stacks the $n$ latent variables of expressions of the images, $\bm{Z}_{id} \in \mathbb{R}^{k_{id} \times n}$ stacks the $n$ latent variables of identity and $\bm{Z}_{l} \in \mathbb{R}^{k_{l} \times n}$ stacks the $n$ latent variables of illumination.


\subsection{3D Facial Shape}

It is quite common to use a bilinear model for disentangling identity and expression in 3D facial shape~\cite{bolkart2016robust}. Hence, for 3D shape we assume that there is a different core tensor $\mathcal{B} \in \mathbb{R}^{k_{3d} \times k_{exp} \times k_{id}}$ and each 3D facial shape 
$\bm{x}_{3d} \in \mathbb{R}^{k_{3d}}$ can be decomposed as:
\begin{equation}
\bm{x}_{3d} = \mathcal{B} \times_2 \bm{z}_{exp} \times_3 \bm{z}_{id},
\end{equation}
where $\bm{z}_{exp}$ and $\bm{z}_{id}$ are exactly the same weights as in the texture decomposition (\ref{eq:xf}). The tensor decomposition for the $n$ images in the batch is therefore written as as
\begin{equation}
\bm{X}_{3d} = \bm{B}_{(1)} (\bm{Z}_{exp} \odot \bm{Z}_{id}),
\label{eq:x3d}
\end{equation}
where $\bm{B}_{(1)}$ is a mode-1 matricization of tensor $\mathcal{B}$. 

\subsection{Facial Normals}

The tensor decomposition we opted to use for facial normals was exactly the same as the texture, hence we can  use the same core tensor and weights. The difference is that since facial normals do not depend on illumination parameters (assuming a Lambertian illumination model), we just need to replace the illumination weights with a constant\footnote{This is also the way that normals are computed in~\cite{wang2017learning} up to a scaling factor}. Thus, the decomposition for normals can be written as
\begin{equation}
\bm{X}_{N} = \bm{Q}_{(1)} (\frac{1}{k_l} \bm{\mathfrak{1}} \odot \bm{Z}_{exp} \odot \bm{Z}_{id}),
\label{eq:xn}
\end{equation}
where $\bm{\mathfrak{1}}$ is a matrix of ones. 

\subsection{3D Facial Pose}
Finally, we define another latent variable regarding 3D pose. This latent variable $\bm{z}_p \in \mathbb{R}^{9}$ represents a 3D rotation. We denote by $\bm{x}^i \in \mathbb{R}^{k_x}$ an image at index $i$. The indexing is denoted in the following by the superscript. The corresponding $\bm{z}_p^i$ can be reshaped into a rotation matrix $\bm{R}^i \in \mathbb{R}^{3 \times 3}$. As proposed in~\cite{Worrall_2017_ICCV}, we apply this rotation to the feature of the image $\bm{x}^i$ created by 2-way synthesis (explained in Section~\ref{networkarch}). This feature vector is the $i$-th column of the feature matrix resulting from the 2-way synthesis $(\bm{Z}_{exp} \odot \bm{Z}_{id}) \in \mathbb{R}^{k_{exp}k_{id} \times n}$. We denote this feature vector corresponding to a single image as $(\bm{Z}_{exp} \odot \bm{Z}_{id})^i \in \mathbb{R}^{k_{exp}k_{id}}$. 
Next $(\bm{Z}_{exp} \odot \bm{Z}_{id})^i$ is reshaped into a $3 \times \frac{k_{exp}k_{id}}{3}$ matrix and left-multiplied by $\bm{R}^i$. After another round of vectorisation, the resulting feature $\in \mathbb{R}^{k_{exp}k_{id}}$ becomes the input of the decoders for normal and albedo. This transformation from feature vector $(\bm{Z}_{exp} \odot \bm{Z}_{id})^i$ to the rotated feature is called \textbf{rotation}.

\subsection{Network Architecture}
\label{networkarch}
We incorporate the structure imposed by Equations \eqref{eq:xf}, \eqref{eq:x3d} and \eqref{eq:xn} into an auto-encoder network, see Figure~\ref{network}. For some matrices $\bm{Y}_i \in \mathbb{R}^{k_{yi} \times n}$, we refer to the operation $\bm{Y}_1 \odot \bm{Y}_2 \in \mathbb{R}^{k_{y1}k_{y2} \times n}$ as \textbf{2-way synthesis} and $\bm{Y}_1 \odot \bm{Y}_2 \odot \bm{Y}_3 \in \mathbb{R}^{k_{y1}k_{y2}k_{y3} \times n}$ as \textbf{3-way synthesis}. The multiplication of a feature matrix by $\bm{B}_{(1)}$ or $\bm{Q}_{(1)}$, mode-1 matricisations of tensors $\mathcal{B}$ and $\mathcal{Q}$, is referred to as \textbf{projection} and can be represented by an unbiased fully-connected layer.

Our network follows the architecture of~\cite{shu2017neural}. The encoder $E$ receives an input image $\bm{x}$ and the convolutional encoder stack first encodes it into $\bm{z}_{i}$, an intermediate latent variable vector  of size $128 \times 1$. $\bm{z}_{i}$ is then transformed into latent codes for background $\bm{z}_{b}$, mask $\bm{z}_{m}$, illumination $\bm{z}_{l}$, pose $\bm{z}_{p}$, identity $\bm{z}_{id}$ and expression $\bm{z}_{exp}$ via fully-connected layers.
\begin{equation}
E(\bm{x}) = [\bm{z}_{b}, \bm{z}_{m}, \bm{z}_{l}, \bm{z}_{p}, \bm{z}_{id}, \bm{z}_{exp} ]^T.
\label{eq:encode}
\end{equation}

The decoder $D$ takes in the latent codes as input. $\bm{z}_{b}$ and $\bm{z}_{m}$ ($128 \times 1$ vectors) are directly passed into convolutional decoder stacks to estimate background and face mask respectively.
The remaining latent variables follow 3 streams:
\begin{enumerate}[leftmargin=*]
\item $\bm{z}_{exp}$ ($15 \times 1$ vector) and $\bm{z}_{id}$ ($80 \times 1$ vector) are joined by 2-way synthesis and projection to estimate facial shape $\hat{\bm{x}_{3d}}$. 
\item The result of 2-way synthesis of $\bm{z}_{exp}$ and $\bm{z}_{id}$ is rotated using $\bm{z}_{p}$. The rotated feature is passed into 2 different convolutional decoder stacks: one for normal estimation and another for albedo. 
Using the estimated normal map, albedo, illumination component $\bm{z}_{l}$, 
mask and background, we render a reconstructed image $\hat{\bm{x}}$. 
\item $\bm{z}_{exp}$, $\bm{z}_{id}$ and $\bm{z}_{l}$ are combined by a 3-way synthesis and projection to estimate frontal normal map and a frontal reconstruction of the image. 
\end{enumerate}
Streams 1 and 3 drive the disentangling of expression and identity components, while stream 2 focuses on the reconstruction of the image by adding the pose components.

\begin{equation}
D(\bm{z}_{b}, \bm{z}_{m}, \bm{z}_{l}, \bm{z}_{p}, \bm{z}_{id}, \bm{z}_{exp}) = \hat{\bm{x}}.
\label{eq:decode}
\end{equation}

Our input images are aligned and cropped facial images from the CelebA database~\cite{liu2015faceattributes} of size $64 \times 64$, so $k_x = 3 \times 64 \times 64$. 
$k_{3d} = 3 \times 9375$, $k_{l} = 9$, $k_{id} = 80$ and $k_{exp} = 15$.
More details on the network such as the convolutional encoder stacks and decoder stacks can be found in the supplementary material.

\begin{figure*}[!thb]
\centering
\includegraphics[width=0.919\linewidth]{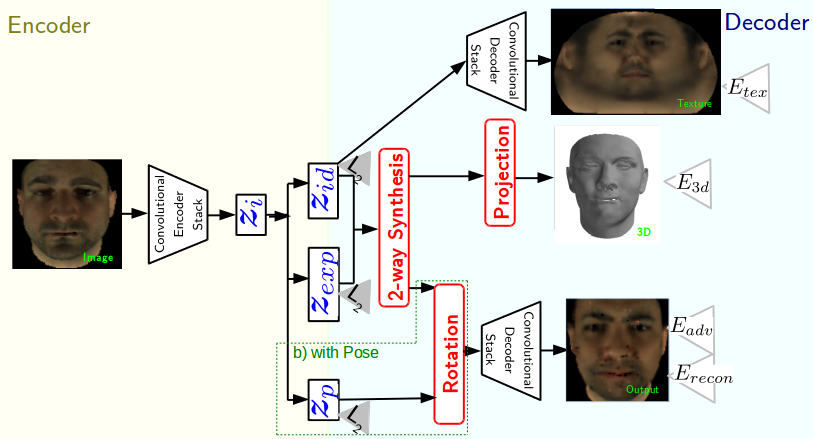}
\caption{Our proof-of-concept network is an end-to-end trained auto-encoder. The encoder $E$ extracts latent variables corresponding to  expression and identity from the input image $\bm{x}$. These latent variables are then fed into the decoder $D$ to reconstruct the image. A separate stream also reconstructs facial texture from $\bm{z}_{id}$. We impose a multilinear structure and enforce the disentanglement of variations. In the extended version b) the encoder also extracts a latent variable corresponding to pose. The decoder takes in this information and reconstructs an image containing pose variations.}
\label{simplenetwork}
\vspace{-10pt}
\end{figure*}

\subsection{Training}
We use in-the-wild face images for training. Hence,
we only have access to the image itself ($\bm{x}$) while ground truth labelling for pose, illumination, normal, albedo, expression, identity or 3D shape is unavailable.  The main loss function is the reconstruction loss of the image $x$:
\begin{equation}
E_x = E_{recon} + \lambda_{adv} E_{adv} + \lambda_{veri} E_{veri},
\end{equation}
where $\hat{\bm{x}}$ is the reconstructed image, $E_{recon} = \| \bm{x} - \hat{\bm{x}} \|^2_2$ is the reconstruction loss,  $\lambda_{adv}$ and $lambda_{adv}$ are regularisation weights, $E_{adv}$ represents the adversarial loss and $E_{veri}$ the verification loss. We use the pre-trained verification network $\mathcal{V}$~\cite{DBLP:journals/corr/WuHS15} to find face embeddings of our images $\bm{x}$ and $\hat{\bm{x}}$. As both images are supposed to represent the same person, we minimise the cosine distance between the embeddings: $E_{veri} = 1 - cos(\mathcal{V}(\bm{x}), \mathcal{V}(\hat{\bm{x}}))$. Simultaneously, a discriminative network $\mathcal{D}$ is trained to distinguish between the generated and real images~\cite{goodfellow2014generative}. We incorporate the discriminative information by following the auto-encoder loss distribution matching
approach of~\cite{berthelot2017began}. The discriminative network $\mathcal{D}$ is itself an auto-encoder trying to reconstruct the input image $\bm{x}$ so the adversarial loss is $E_{adv} = \|\hat{\bm{x}} - \mathcal{D}(\hat{\bm{x}})\|_1$. $\mathcal{D}$ is trained to minimise $\|\bm{x} - \mathcal{D}(\bm{x})\|_1 - k_t \|\hat{\bm{x}} - \mathcal{D}(\hat{\bm{x}})\|_1$. 
 
As fully unsupervised training often results in semantically meaningless latent representations, Shu et al.~\cite{shu2017neural} proposed to train with “pseudo ground truth” values for normals, lighting and 3D facial shape. We adopt here this technique and introduce further “pseudo ground truth” values for pose $\hat{\bm{x}_p}$, expression $\hat{\bm{x}_{exp}}$ and identity $\hat{\bm{x}_{id}}$. 
$\hat{\bm{x}_p}$, $\hat{\bm{x}_{exp}}$ and $\hat{\bm{x}_{id}}$ are obtained by fitting coarse face geometry to every image in the training set using a 3D Morphable Model~\cite{booth20173d}. We incorporated the constraints used in~\cite{shu2017neural} for illumination, normals and albedo. Hence, the following new objectives are introduced:
\begin{equation}
E_{p} = \| \bm{z}_p -  \hat{\bm{x}_p} \|^2_2,
\end{equation}
where $\hat{\bm{x}_p}$ is a 3D camera rotation matrix.

\begin{equation}
E_{exp} = \|  fc(\bm{z}_{exp}) -  \hat{\bm{x}_{exp}} \|^2_2,
\end{equation} 
where fc($\cdot$) is a fully-connected layer and $\hat{\bm{x}_{exp}} \in \mathbb{R}^{28}$ is a “pseudo ground truth” vector representing 3DMM expression components of the image $\bm{x}$.

\begin{equation}
E_{id} = \| fc(\bm{z}_{id}) -  \hat{\bm{x}_{id}} \|^2_2
\end{equation}
where fc($\cdot$) is a fully-connected layer and $\hat{\bm{x}_{id}} \in \mathbb{R}^{157}$ is a “pseudo ground truth” vector representing 3DMM identity components of the image $\bm{x}$.

\begin{figure*}[!thb]
\captionsetup[subfigure]{labelformat=empty, justification=centering,position=top}
{\def\arraystretch{0.5}\tabcolsep=1pt
\begin{tabular}{ lclc } 
\vspace{-9pt}
\subfloat[Original Image]{\includegraphics[width=0.1\linewidth, height=0.1\linewidth]{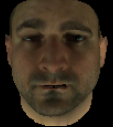}} 
\subfloat[Expression]{\includegraphics[width=0.1\linewidth, height=0.1\linewidth]{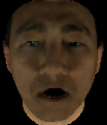}}
\subfloat[Our Recon]{\includegraphics[width=0.1\linewidth, height=0.1\linewidth]{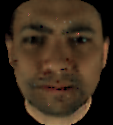}}
\subfloat[Our Exp Edit]{\includegraphics[width=0.1\linewidth, height=0.1\linewidth]{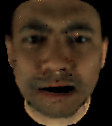}}&
\subfloat[Ground Truth]{\includegraphics[trim={1.5cm 0.3cm 1.5cm 0},clip, width=0.1\linewidth, height=0.1\linewidth]{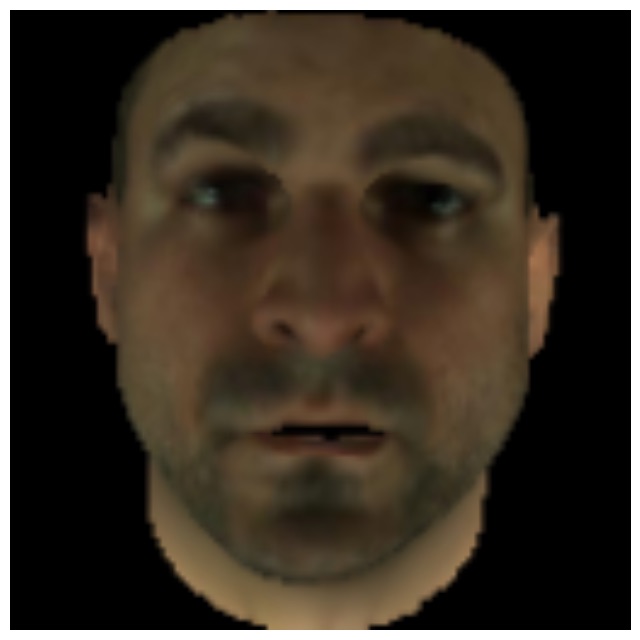}}
&
\subfloat[Original Image]{\includegraphics[width=0.1\linewidth, height=0.1\linewidth]{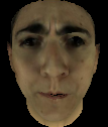}} 
\subfloat[Expression]{\includegraphics[width=0.1\linewidth, height=0.1\linewidth]{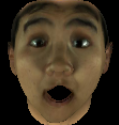}}
\subfloat[Our Recon]{\includegraphics[width=0.1\linewidth, height=0.1\linewidth]{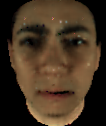}}
\subfloat[Our Exp Edit]{\includegraphics[width=0.1\linewidth, height=0.1\linewidth]{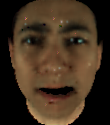}}&
\subfloat[Ground Truth]{\includegraphics[trim={1.5cm 0.3cm 1.5cm 0},clip, width=0.1\linewidth, height=0.1\linewidth]{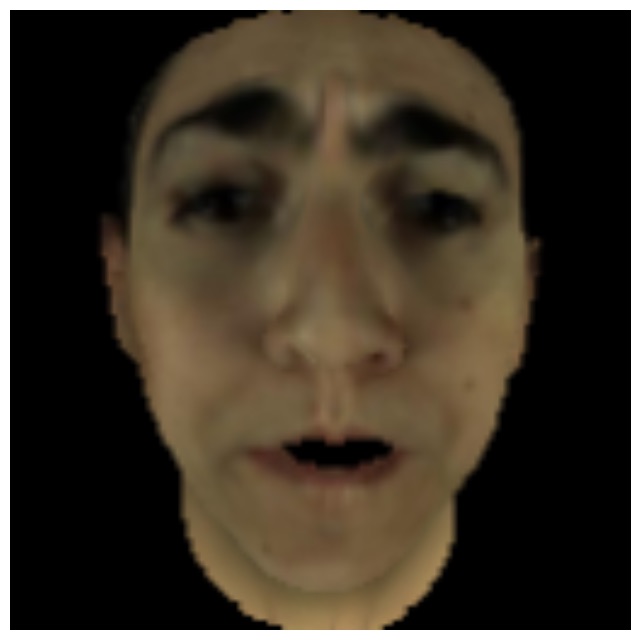}}
\\
\subfloat{\includegraphics[width=0.1\linewidth, height=0.1\linewidth]{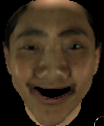}} 
\subfloat{\includegraphics[width=0.1\linewidth, height=0.1\linewidth]{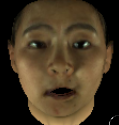}}
\subfloat{\includegraphics[width=0.1\linewidth, height=0.1\linewidth]{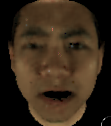}}
\subfloat{\includegraphics[width=0.1\linewidth, height=0.1\linewidth]{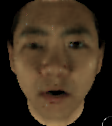}}&
\subfloat{\includegraphics[trim={1.5cm 0.3cm 1.5cm 0},clip, width=0.1\linewidth, height=0.1\linewidth]{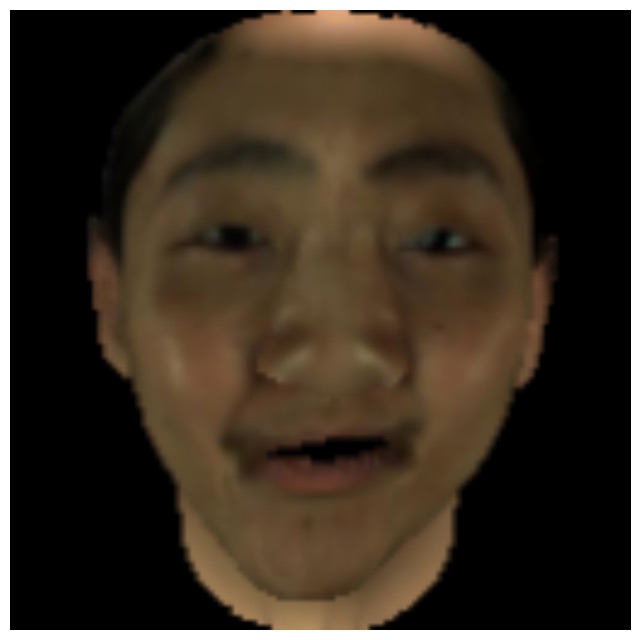}}
&
\subfloat{\includegraphics[width=0.1\linewidth, height=0.1\linewidth]{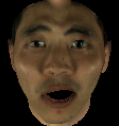}} 
\subfloat{\includegraphics[width=0.1\linewidth, height=0.1\linewidth]{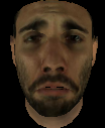}}
\subfloat{\includegraphics[width=0.1\linewidth, height=0.1\linewidth]{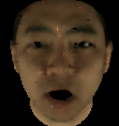}}
\subfloat{\includegraphics[width=0.1\linewidth, height=0.1\linewidth]{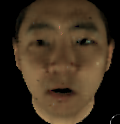}}&
\subfloat{\includegraphics[trim={1.5cm 0.3cm 1.5cm 0},clip, width=0.1\linewidth, height=0.1\linewidth]{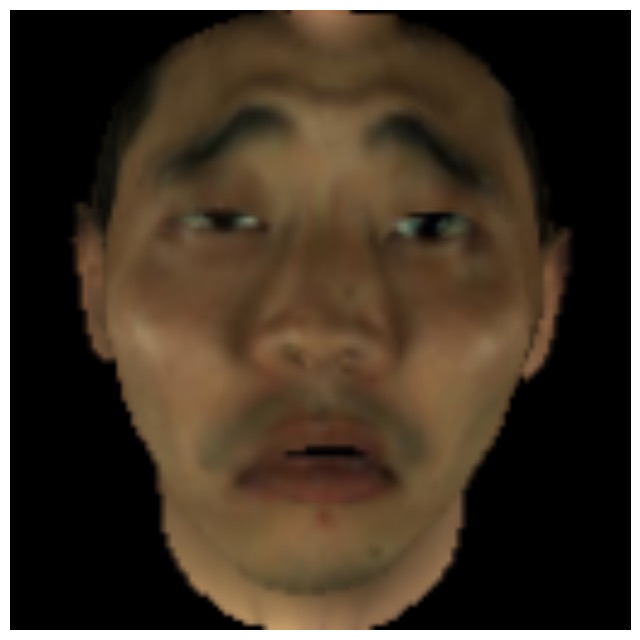}}\\
\end{tabular}
}
\caption{Our network is able to transfer the expression from one face to another by disentangling the expression components of the images. The ground truth has been computed using the ground truth texture with synthetic identity and expression components.}
    \label{synthexp}
\end{figure*}

\begin{figure*}[!thb]
\captionsetup[subfigure]{labelformat=empty, justification=centering,position=top}
{\def\arraystretch{0.5}\tabcolsep=1pt
\begin{tabular}{ p{1.1cm} c } 
Input &
\subfloat{\includegraphics[width=0.9\linewidth]{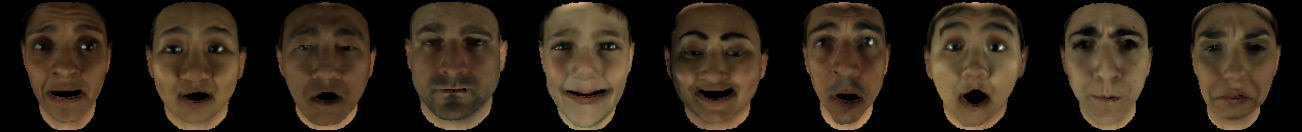}}
\\
Ground Truth &
\subfloat{\includegraphics[trim={4cm 2cm 4cm 1cm},clip,width=0.09\linewidth]{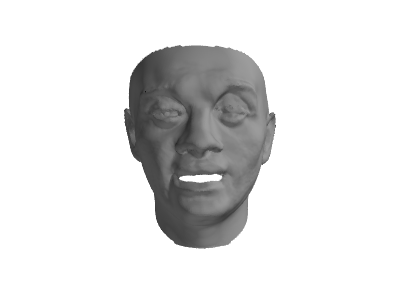}}
\subfloat{\includegraphics[trim={4cm 2cm 4cm 1cm},clip,width=0.09\linewidth]{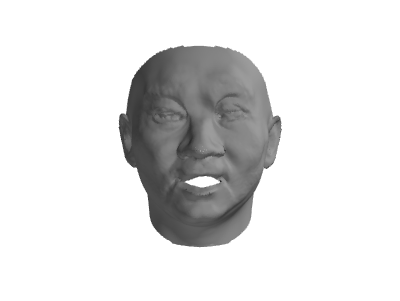}}
\subfloat{\includegraphics[trim={4cm 2cm 4cm 1cm},clip,width=0.09\linewidth]{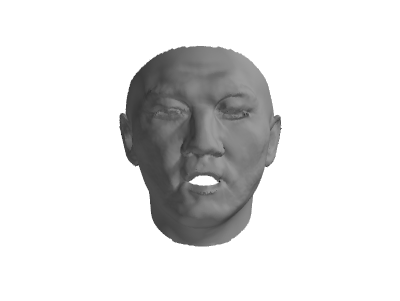}}
\subfloat{\includegraphics[trim={4cm 2cm 4cm 1cm},clip,width=0.09\linewidth]{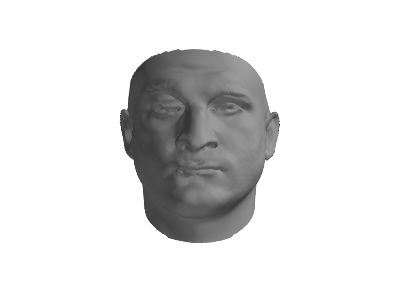}}
\subfloat{\includegraphics[trim={4cm 2cm 4cm 1cm},clip,width=0.09\linewidth]{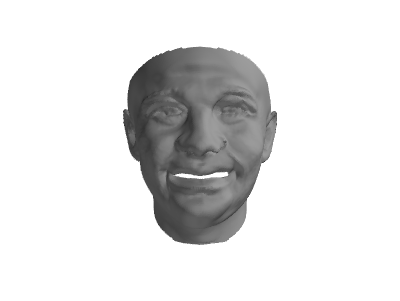}}
\subfloat{\includegraphics[trim={4cm 2cm 4cm 1cm},clip,width=0.09\linewidth]{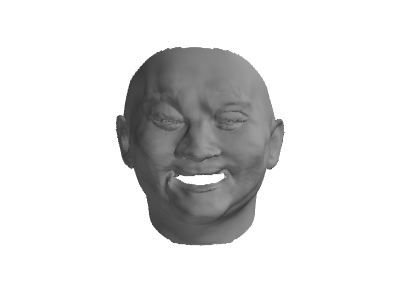}}
\subfloat{\includegraphics[trim={4cm 2cm 4cm 1cm},clip,width=0.09\linewidth]{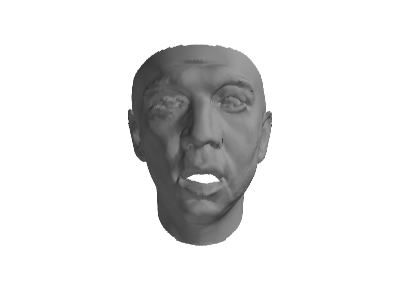}}
\subfloat{\includegraphics[trim={4cm 2cm 4cm 1cm},clip,width=0.09\linewidth]{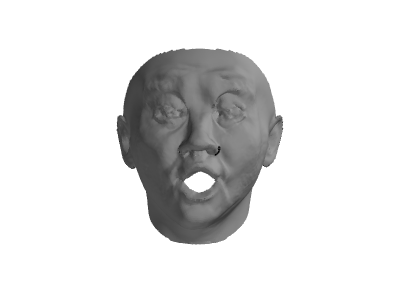}}
\subfloat{\includegraphics[trim={4cm 2cm 4cm 1cm},clip,width=0.09\linewidth]{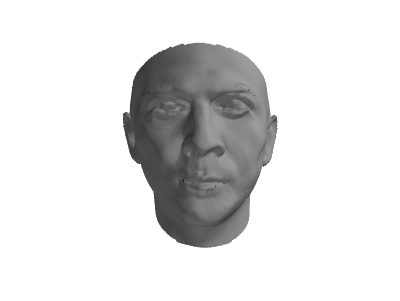}}
\subfloat{\includegraphics[trim={4cm 2cm 4cm 1cm},clip,width=0.09\linewidth]{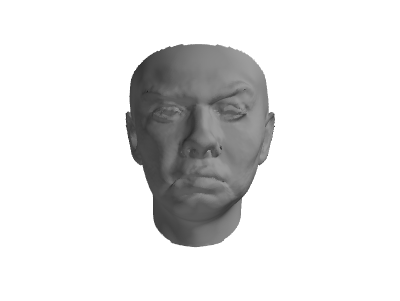}}
\\
Recons-truction &
\subfloat{\includegraphics[trim={4cm 2cm 4cm 1cm},clip,width=0.09\linewidth]{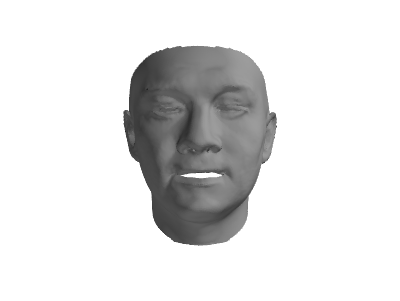}}
\subfloat{\includegraphics[trim={4cm 2cm 4cm 1cm},clip,width=0.09\linewidth]{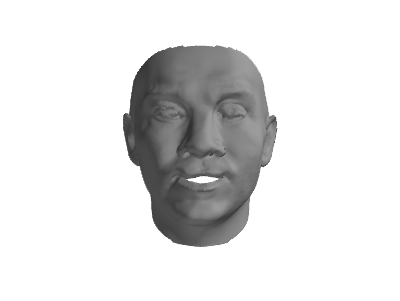}}
\subfloat{\includegraphics[trim={4cm 2cm 4cm 1cm},clip,width=0.09\linewidth]{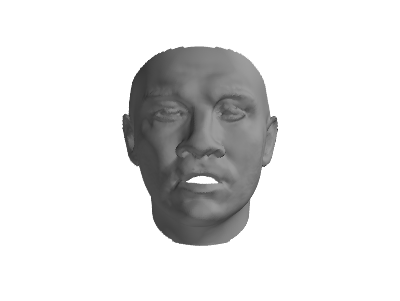}}
\subfloat{\includegraphics[trim={4cm 2cm 4cm 1cm},clip,width=0.09\linewidth]{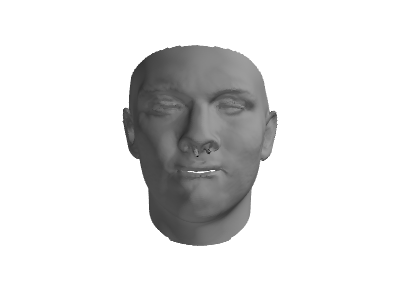}}
\subfloat{\includegraphics[trim={4cm 2cm 4cm 1cm},clip,width=0.09\linewidth]{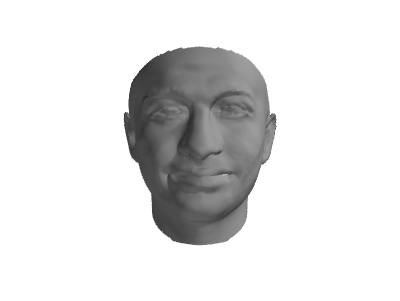}}
\subfloat{\includegraphics[trim={4cm 2cm 4cm 1cm},clip,width=0.09\linewidth]{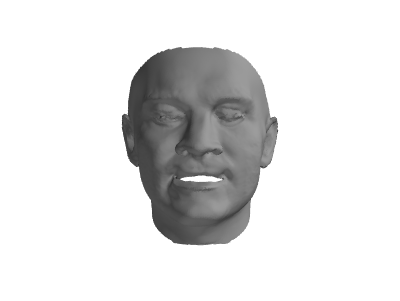}}
\subfloat{\includegraphics[trim={4cm 2cm 4cm 1cm},clip,width=0.09\linewidth]{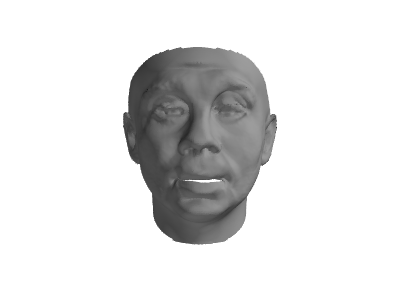}}
\subfloat{\includegraphics[trim={4cm 2cm 4cm 1cm},clip,width=0.09\linewidth]{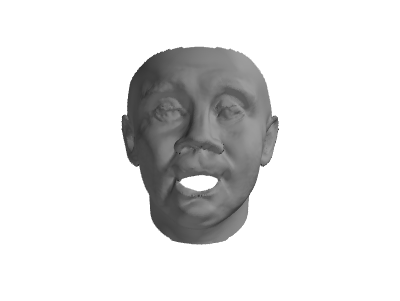}}
\subfloat{\includegraphics[trim={4cm 2cm 4cm 1cm},clip,width=0.09\linewidth]{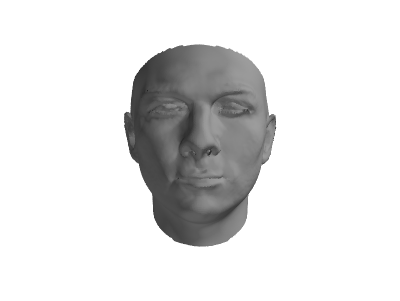}}
\subfloat{\includegraphics[trim={4cm 2cm 4cm 1cm},clip,width=0.09\linewidth]{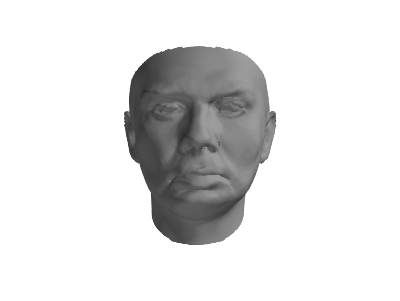}}
\end{tabular}
}
\caption{Given a single image, we infer meaningful expression and identity components to reconstruct a 3D mesh of the face. We compare the reconstruction (last row) against the ground truth ($2^{nd}$ row).}
    \label{synth3d}
    \vspace{-10pt}
\end{figure*}

\begin{figure*}[!thb]
\captionsetup[subfigure]{labelformat=empty, justification=centering,position=top}
{\def\arraystretch{0.5}\tabcolsep=1pt
\begin{tabular}{ p{1.1cm} c } 
Input &
\subfloat{\includegraphics[width=0.9\linewidth]{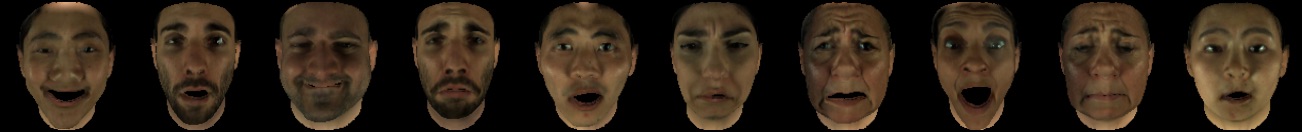}}
\\
Ground Truth &
\subfloat{\includegraphics[width=0.9\linewidth]{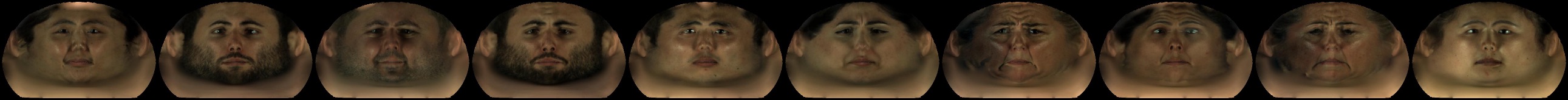}}
\\
Recons-truction &
\subfloat{\includegraphics[width=0.9\linewidth]{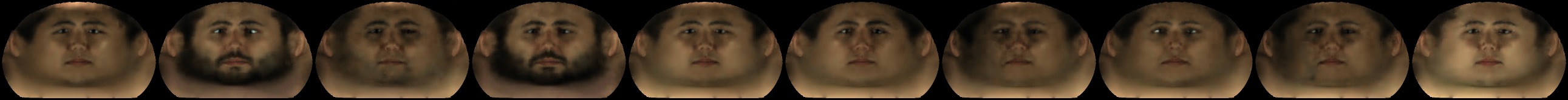}}
\end{tabular}
}
\caption{Given a single image, we infer the facial texture. We compare the reconstructed facial texture (last row) against the ground truth texture ($2^{nd}$ row).}
    \label{synthtex}
    \vspace{-10pt}
\end{figure*}

\begin{figure*}[!thb]
\captionsetup[subfigure]{labelformat=empty, justification=centering,position=top}
{\def\arraystretch{0.5}\tabcolsep=1pt
\begin{tabular}{ lclc } 
\vspace{-9pt}
\subfloat[Original Image]{\includegraphics[width=0.1\linewidth, height=0.1\linewidth]{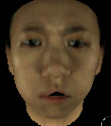}} 
\subfloat[Pose]{\includegraphics[width=0.1\linewidth, height=0.1\linewidth]{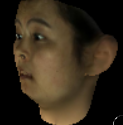}}
\subfloat[Our Recon]{\includegraphics[width=0.1\linewidth, height=0.1\linewidth]{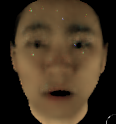}}
\subfloat[Our Pose Edit]{\includegraphics[width=0.1\linewidth, height=0.1\linewidth]{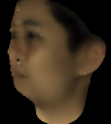}}&
\subfloat[Ground Truth]{\includegraphics[trim={1.5cm 0.3cm 1.5cm 0},clip, width=0.1\linewidth, height=0.1\linewidth]{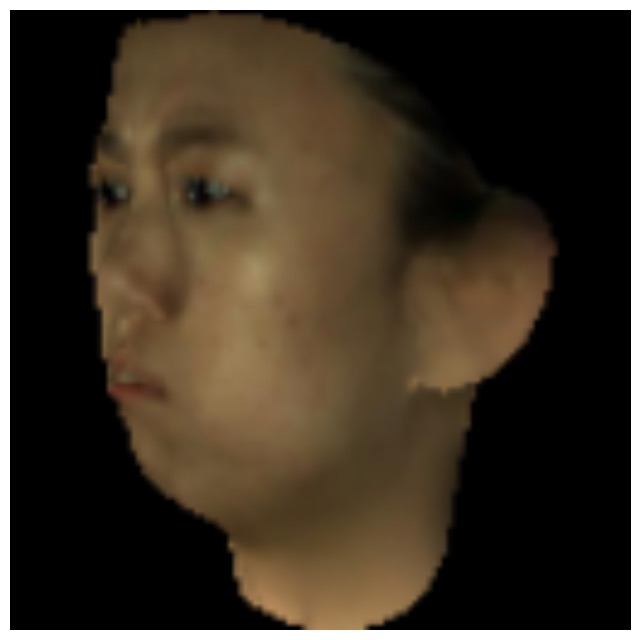}}
&
\subfloat[Original Image]{\includegraphics[width=0.1\linewidth, height=0.1\linewidth]{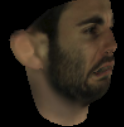}} 
\subfloat[Pose]{\includegraphics[width=0.1\linewidth, height=0.1\linewidth]{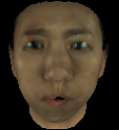}}
\subfloat[Our Recon]{\includegraphics[width=0.1\linewidth, height=0.1\linewidth]{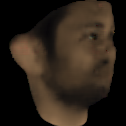}}
\subfloat[Our Pose Edit]{\includegraphics[width=0.1\linewidth, height=0.1\linewidth]{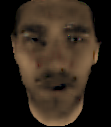}}&
\subfloat[Ground Truth]{\includegraphics[trim={1cm 0.3cm 1cm 0},clip, width=0.1\linewidth, height=0.1\linewidth]{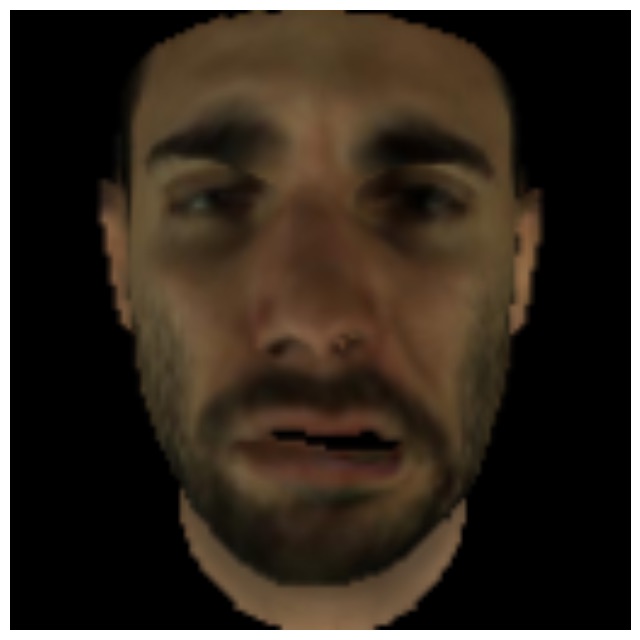}}
\\
\subfloat{\includegraphics[width=0.1\linewidth, height=0.1\linewidth]{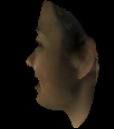}} 
\subfloat{\includegraphics[width=0.1\linewidth, height=0.1\linewidth]{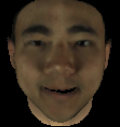}}
\subfloat{\includegraphics[width=0.1\linewidth, height=0.1\linewidth]{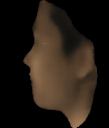}}
\subfloat{\includegraphics[width=0.1\linewidth, height=0.1\linewidth]{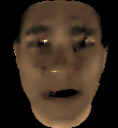}}&
\subfloat{\includegraphics[trim={1cm 0.3cm 1cm 0},clip, width=0.1\linewidth, height=0.1\linewidth]{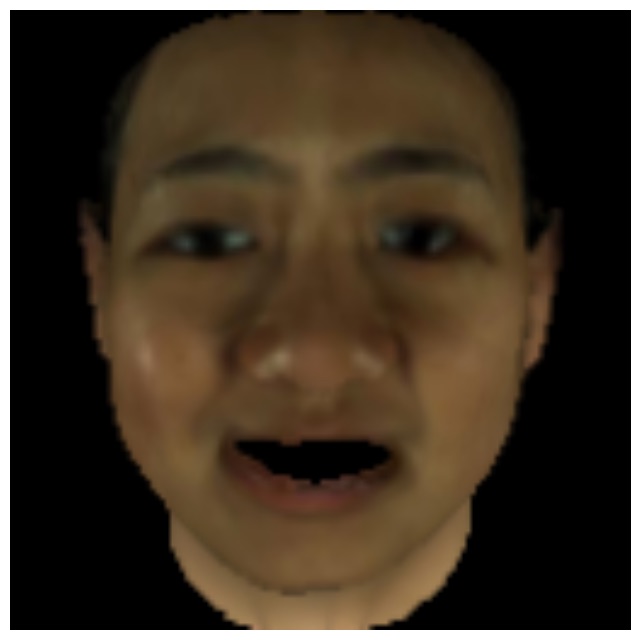}}&
\subfloat{\includegraphics[width=0.1\linewidth, height=0.1\linewidth]{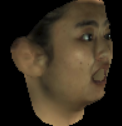}} 
\subfloat{\includegraphics[width=0.1\linewidth, height=0.1\linewidth]{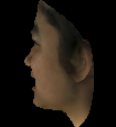}}
\subfloat{\includegraphics[width=0.1\linewidth, height=0.1\linewidth]{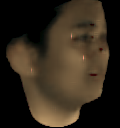}}
\subfloat{\includegraphics[width=0.1\linewidth, height=0.1\linewidth]{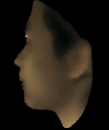}}&
\subfloat{\includegraphics[trim={1cm 0.3cm 1cm 0},clip, width=0.1\linewidth, height=0.1\linewidth]{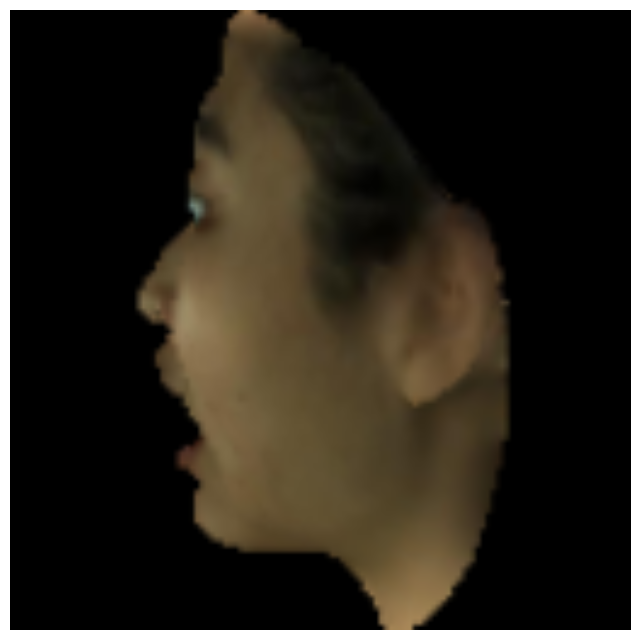}}\\
\end{tabular}
}
\caption{Our network is able to transfer the pose from one face to another by disentangling the pose, expression and identity components of the images. The ground truth has been computed using the ground truth texture with synthetic pose, identity and expression components.}
    \label{synthpose}
\end{figure*}

\subsubsection*{Multilinear Losses}

Directly applying the above losses as constraints to the latent variables does not result in a well-disentangled representation. 
To achieve a better performance, we impose a tensor structure on the image using the following losses:
\begin{equation}
E_{3d} = \| \hat{\bm{x}_{3d}} - \mathcal{B} \times_2 \bm{z}_{exp} \times_3 \bm{z}_{id} \|^2_2 ,
\label{eq:e3d}
\end{equation}
where $\hat{\bm{x}_{3d}}$ is the 3D facial shape of the fitted model.

\begin{equation}
E_{f} = \| \bm{x}_{f} - \mathcal{Q} \times_2 \bm{z}_{l} \times_3 \bm{z}_{exp} \times_4 \bm{z}_{id}) \|^2_2,
\label{eq:ef}
\end{equation}
where $\bm{x}_{f}$ is a semi-frontal face image. During training, $E_{f}$ is only applied on near-frontal face images filtered using $\hat{\bm{x}_p}$.

\begin{equation}
E_{n} = \| \hat{\bm{n}_{f}} - \mathcal{Q} \times_2 \frac{1}{k_l} \bm{1} \times_3 \bm{z}_{exp} \times_4 \bm{z}_{id}) \|^2_2
\label{eq:en}
\end{equation}
where $\hat{\bm{n}_{f}}$ is a near frontal normal map. During training, the loss $E_{n}$ is only applied on near frontal normal maps. 

The model is trained end-to-end by applying gradient descent to batches of images, where Equations~\eqref{eq:e3d}, \eqref{eq:ef} and \eqref{eq:en} are written in the following general form:
\begin{equation} \label{eq:tensor_err}
E = \| \bm{X} - \bm{B}_{(1)} (\bm{Z}^{(1)} \odot \bm{Z}^{(2)} \odot \dots \odot \bm{Z}^{(M)})\|^2_F,
\end{equation}
where $M$ is the number of modes of variations, $\bm{X} \in \mathbb{R}^{k \times n}$ is a data matrix, $\bm{B}_{(1)}$ is the mode-1 matricisation of a tensor $\mathcal{B}$ and $\bm{Z}^{(i)} \in \mathbb{R}^{k_{zi} \times n}$ are the latent variables matrices.

The partial derivative of \eqref{eq:tensor_err} with respect to the latent variable $\bm{Z}^{(i)}$ are computed as follows: 
Let $\hat{\bm{x}} = vec(\bm{X})$ be the vectorised $\bm{X}$, $\hat{\bm{z}}^{(i)} = vec(\bm{Z}^{(i)})$ be the vectorised $\bm{Z}^{(i)}$,  

$\hat{\bm{Z}^{(i-1)}} = \bm{Z}^{(1)} \odot \bm{Z}^{(2)} \odot \dots \odot \bm{Z}^{(i-1)} $
and 
$\hat{\bm{Z}^{(i+1)}} = \bm{Z}^{(i+1)} \odot \dots \odot \bm{Z}^{(M)} $
, then \eqref{eq:tensor_err} is equivalent with:
\begin{equation}
\begin{aligned}
& \| \hat{\bm{x}} - (\bm{I} \otimes \bm{B}_{(1)}) vec(\bm{Z}^{(1)} \odot \bm{Z}^{(2)} \odot \dots \odot \bm{Z}^{(M)} )\|^2_F \\
=& \| \hat{\bm{x}} - (\bm{I} \otimes \bm{B}_{(1)})  (\bm{I} \odot \hat{\bm{Z}^{(i-1)}} ) \otimes \bm{I} \\
& \qquad \cdot \bm{I} \odot (\hat{\bm{Z}^{(i+1)}} (\bm{I} \otimes \bm{\mathbb{1}}) ) \cdot \hat{\bm{z}^{(i)}}\|^2_2
\end{aligned}
\label{eq:tensor_vec}
\end{equation}

Consequently the partial derivative of~\eqref{eq:tensor_err} with respect to $\bm{Z}^{(i)}$ is obtained by matricising the partial derivative of~\eqref{eq:tensor_vec} with respect to $\bm{Z}^{(i)}$, which is easy to compute analytically. The derivation of this can be found in the supplemental material. To efficiently compute the above mentioned operations, Tensorly~\cite{tensorly} has been employed.

\begin{figure*}[t!]
\captionsetup[subfigure]{labelformat=empty, justification=centering,position=top}
{\def\arraystretch{0.5}\tabcolsep=1pt
\begin{tabular}{ cccc } 
\vspace{-9pt}
\subfloat[Original Image]{\includegraphics[trim={0 0.1cm 0 0},clip,width=0.1\linewidth]{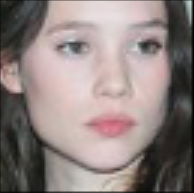}} 
\subfloat[Expression]{\includegraphics[trim={0.1cm 0 0 0},clip, width=0.1\linewidth, height=0.1\linewidth]{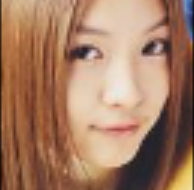}}
\subfloat[Our Recon]{\includegraphics[trim={0.1cm 0 0 0},clip,width=0.1\linewidth, height=0.1\linewidth]{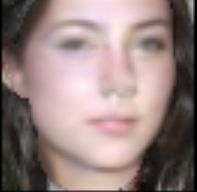}}
\subfloat[Our Exp Edit]{\includegraphics[width=0.1\linewidth, height=0.1\linewidth]{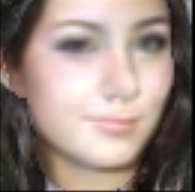}}&
\subfloat[Baseline]{\includegraphics[trim={0 0.3cm 0 0},clip, width=0.1\linewidth, height=0.1\linewidth]{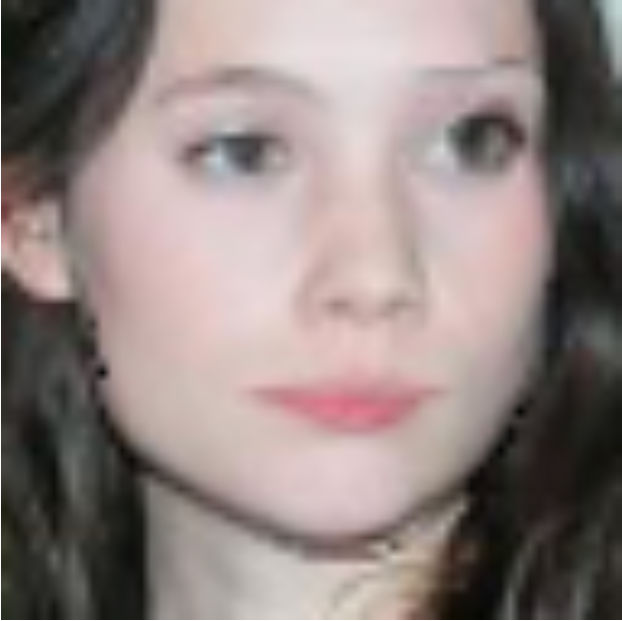}}
&
\subfloat[Original Image]{\includegraphics[width=0.1\linewidth, height=0.1\linewidth]{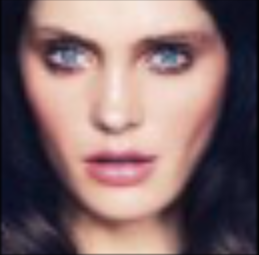}} 
\subfloat[Expression]{\includegraphics[trim={0 0.1cm 0 0.1cm},clip, width=0.1\linewidth, height=0.1\linewidth]{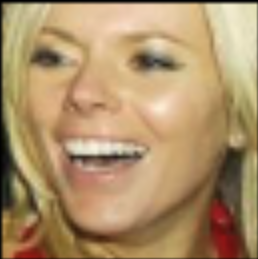}}
\subfloat[Our Recon]{\includegraphics[trim={0 0 0.2cm 0},clip, width=0.1\linewidth, height=0.1\linewidth]{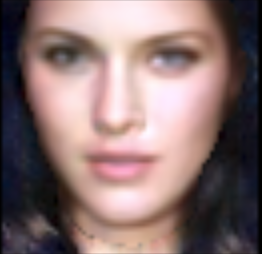}}
\subfloat[Our Exp Edit]{\includegraphics[trim={0 0.1cm 0 0.1cm},clip, width=0.1\linewidth, height=0.1\linewidth]{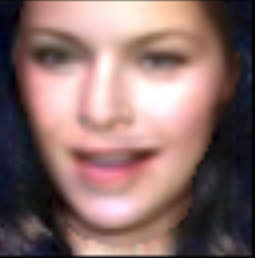}}&
\subfloat[Baseline]{\includegraphics[trim={0 0.1cm 0 0.1cm},clip, width=0.1\linewidth, height=0.1\linewidth]{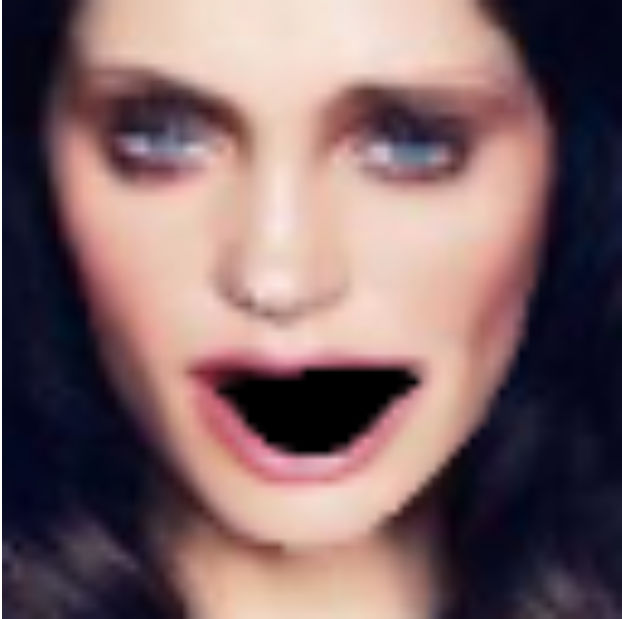}}
\\ \vspace{-9pt}
\subfloat{\includegraphics[width=0.1\linewidth, height=0.1\linewidth]{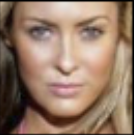}} 
\subfloat{\includegraphics[width=0.1\linewidth, height=0.1\linewidth]{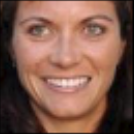}}
\subfloat{\includegraphics[width=0.1\linewidth, height=0.1\linewidth]{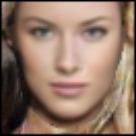}}
\subfloat{\includegraphics[width=0.1\linewidth, height=0.1\linewidth]{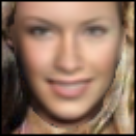}}&
\subfloat{\includegraphics[width=0.1\linewidth, height=0.1\linewidth]{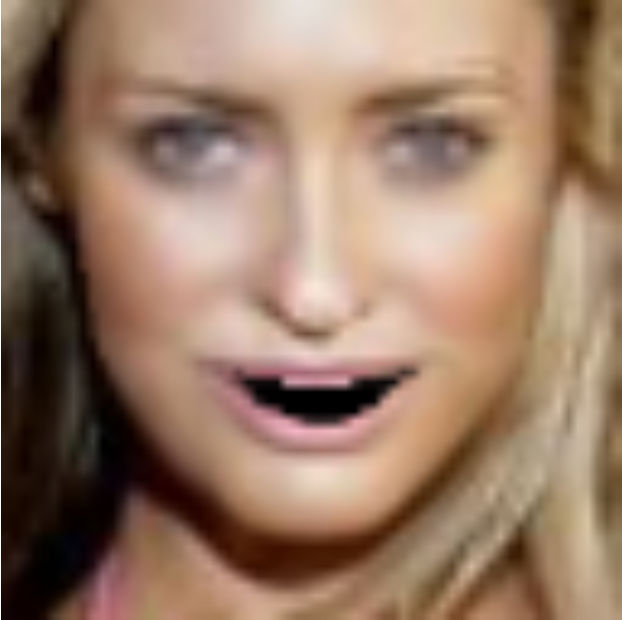}}
&
\subfloat{\includegraphics[width=0.1\linewidth, height=0.1\linewidth]{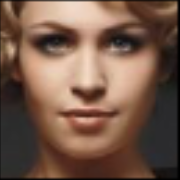}} 
\subfloat{\includegraphics[width=0.1\linewidth, height=0.1\linewidth]{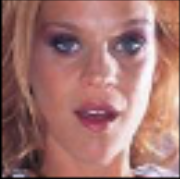}}
\subfloat{\includegraphics[width=0.1\linewidth, height=0.1\linewidth]{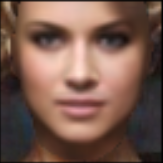}}
\subfloat{\includegraphics[width=0.1\linewidth, height=0.1\linewidth]{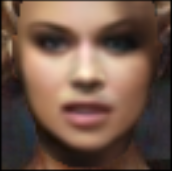}}&
\subfloat{\includegraphics[width=0.1\linewidth, height=0.1\linewidth]{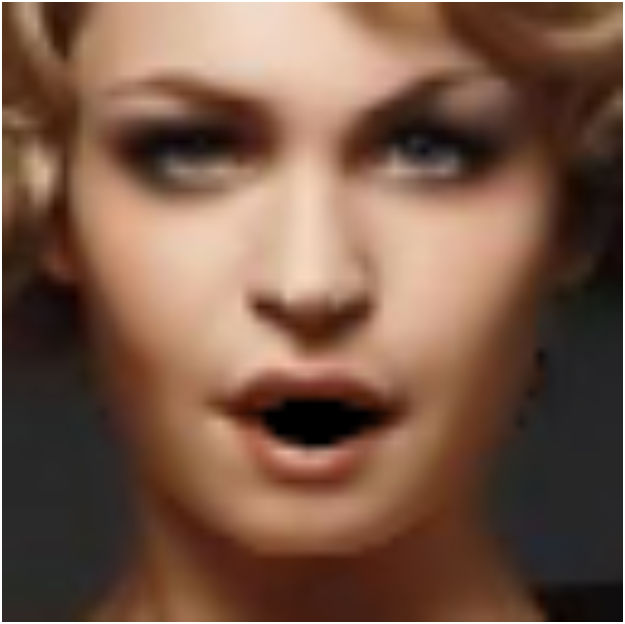}}
\\ \vspace{-9pt}
\subfloat{\includegraphics[trim={0 0.1cm 0.1cm 0.1cm},clip, width=0.1\linewidth, height=0.1\linewidth]{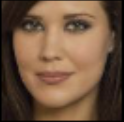}} 
\subfloat{\includegraphics[trim={0 0.1cm 0.1cm 0.1cm},clip, width=0.1\linewidth, height=0.1\linewidth]{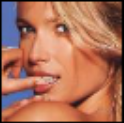}}
\subfloat{\includegraphics[trim={0.1cm 0.1cm 0.1cm 0.1cm},clip,width=0.1\linewidth, height=0.1\linewidth]{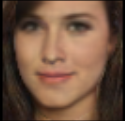}}
\subfloat{\includegraphics[trim={0.1cm 0.15cm 0.1cm 0.2cm},clip, width=0.1\linewidth, height=0.1\linewidth]{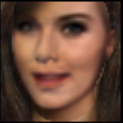}}&
\subfloat{\includegraphics[trim={0 0.45cm 0 0.35cm},clip, width=0.1\linewidth, height=0.1\linewidth]{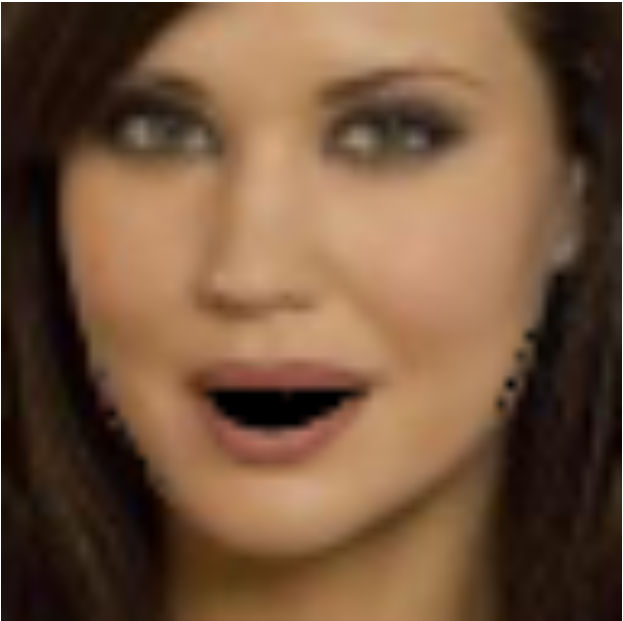}}
&
\subfloat{\includegraphics[width=0.1\linewidth, height=0.1\linewidth]{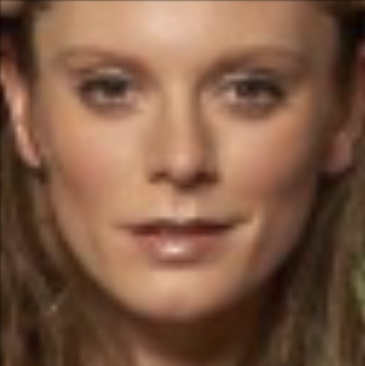}} 
\subfloat{\includegraphics[width=0.1\linewidth, height=0.1\linewidth]{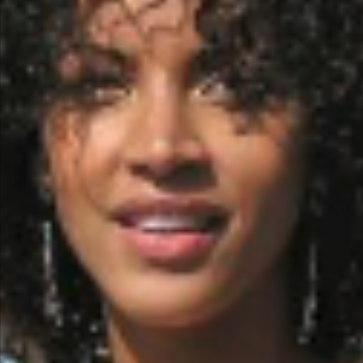}}
\subfloat{\includegraphics[width=0.1\linewidth, height=0.1\linewidth]{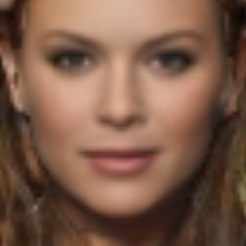}}
\subfloat{\includegraphics[width=0.1\linewidth, height=0.1\linewidth]{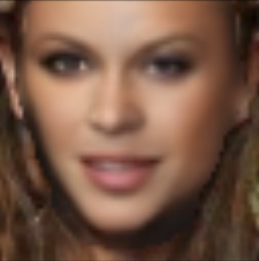}}&
\subfloat{\includegraphics[width=0.1\linewidth, height=0.1\linewidth]{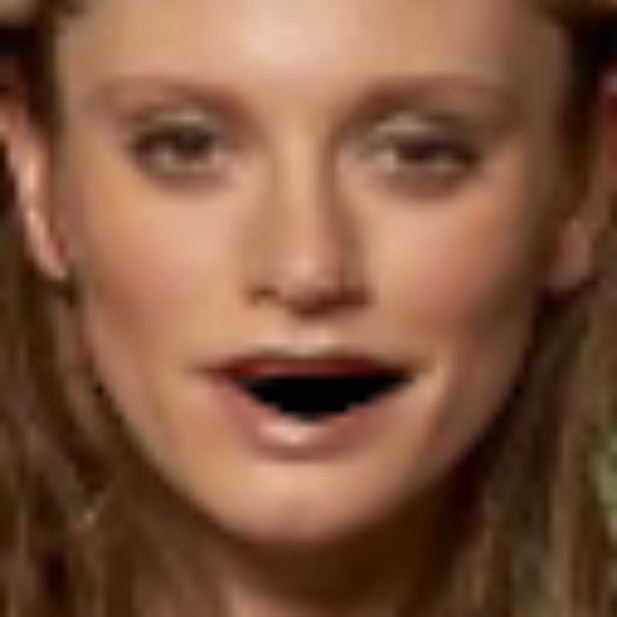}}
\\ \vspace{-9pt}
\subfloat{\includegraphics[width=0.1\linewidth, height=0.1\linewidth]{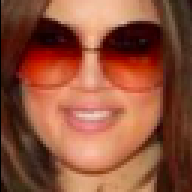}} 
\subfloat{\includegraphics[width=0.1\linewidth, height=0.1\linewidth]{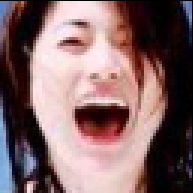}}
\subfloat{\includegraphics[width=0.1\linewidth, height=0.1\linewidth]{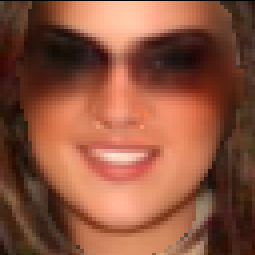}}
\subfloat{\includegraphics[width=0.1\linewidth, height=0.1\linewidth]{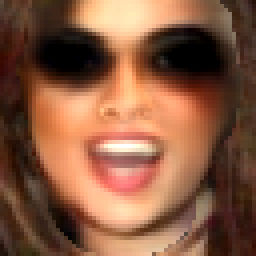}}&
\subfloat{\includegraphics[width=0.1\linewidth, height=0.1\linewidth]{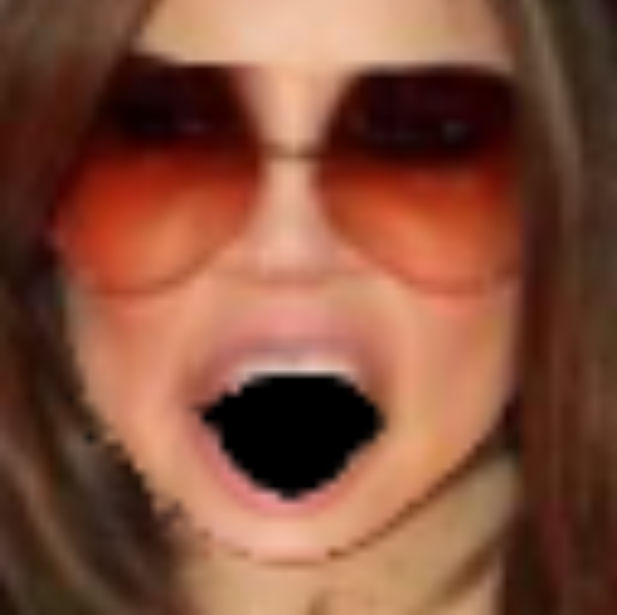}}
&
\subfloat{\includegraphics[width=0.1\linewidth, height=0.1\linewidth]{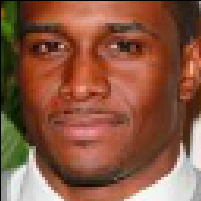}}  
\subfloat{\includegraphics[width=0.1\linewidth, height=0.1\linewidth]{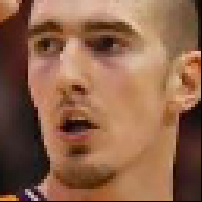}}
\subfloat{\includegraphics[width=0.1\linewidth, height=0.1\linewidth]{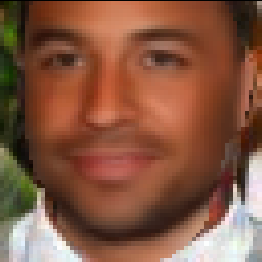}}
\subfloat{\includegraphics[width=0.1\linewidth, height=0.1\linewidth]{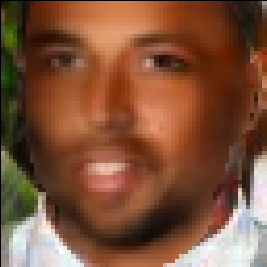}}&
\subfloat{\includegraphics[width=0.1\linewidth, height=0.1\linewidth]{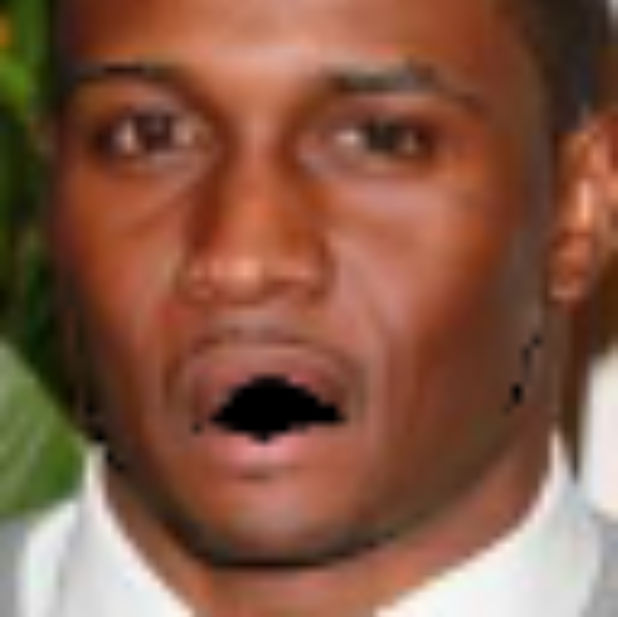}}
\\ \vspace{-9pt}
\subfloat{\includegraphics[width=0.1\linewidth, height=0.1\linewidth]{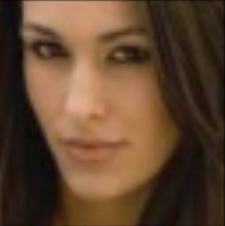}}  
\subfloat{\includegraphics[width=0.1\linewidth, height=0.1\linewidth]{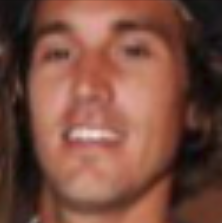}}
\subfloat{\includegraphics[width=0.1\linewidth, height=0.1\linewidth]{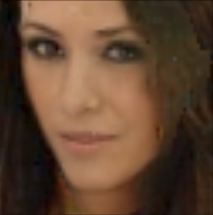}}
\subfloat{\includegraphics[width=0.1\linewidth, height=0.1\linewidth]{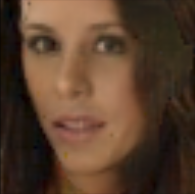}}&
\subfloat{\includegraphics[width=0.1\linewidth, height=0.1\linewidth]{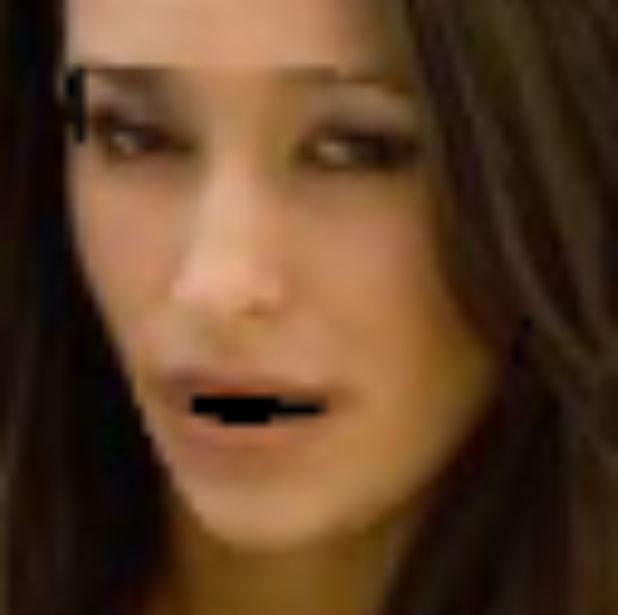}}
&
\subfloat{\includegraphics[width=0.1\linewidth]{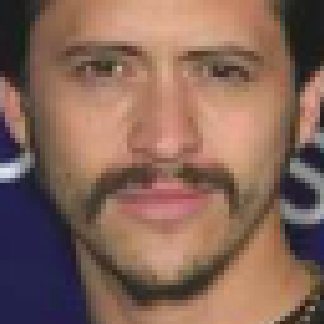}} 
\subfloat{\includegraphics[width=0.1\linewidth]{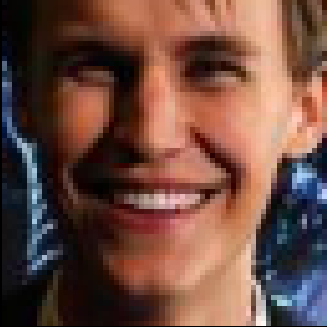}}
\subfloat{\includegraphics[width=0.1\linewidth]{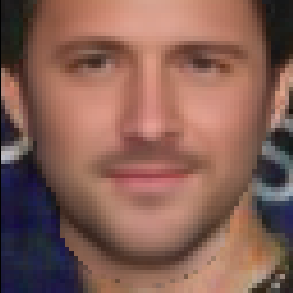}}
\subfloat{\includegraphics[width=0.1\linewidth]{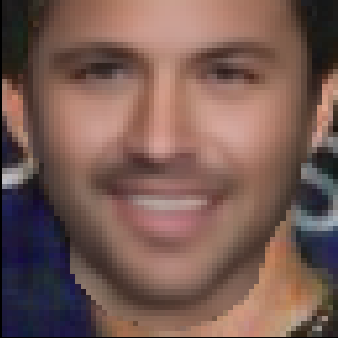}}&
\subfloat{\includegraphics[width=0.1\linewidth]{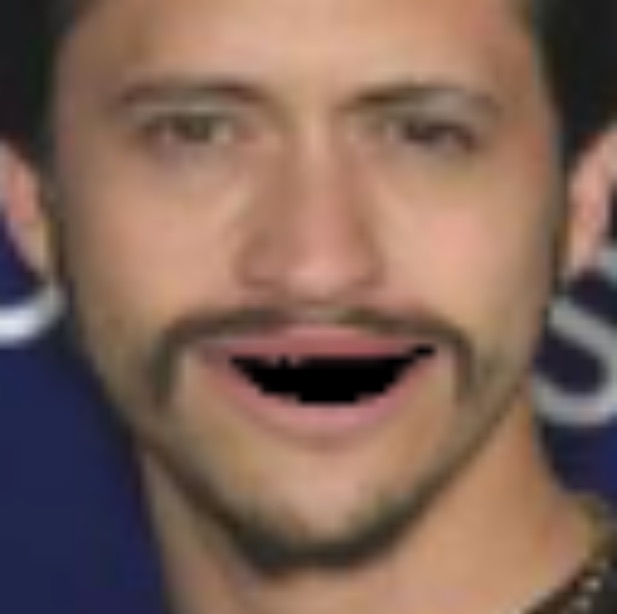}}
\\ \vspace{-9pt}
\subfloat{\includegraphics[width=0.1\linewidth]{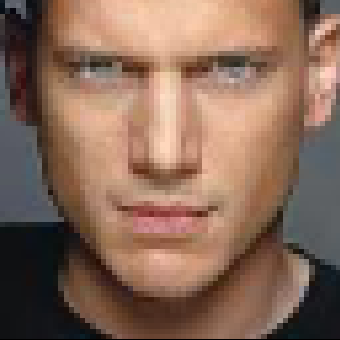}} 
\subfloat{\includegraphics[width=0.1\linewidth]{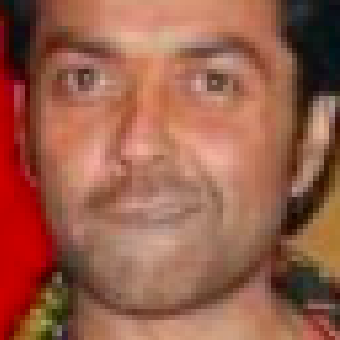}}
\subfloat{\includegraphics[width=0.1\linewidth]{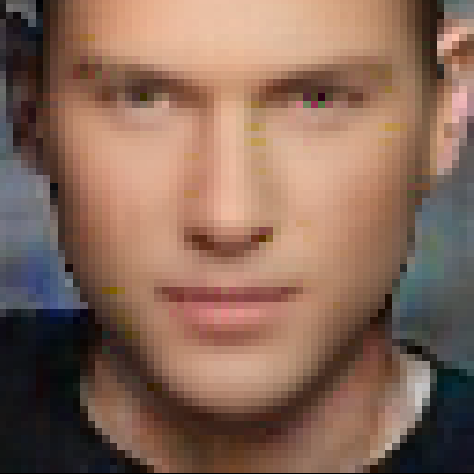}}
\subfloat{\includegraphics[width=0.1\linewidth]{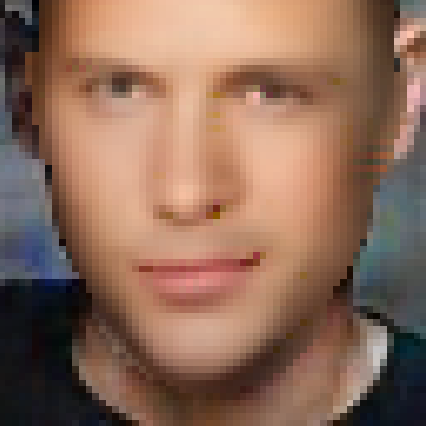}}&
\subfloat{\includegraphics[width=0.1\linewidth]{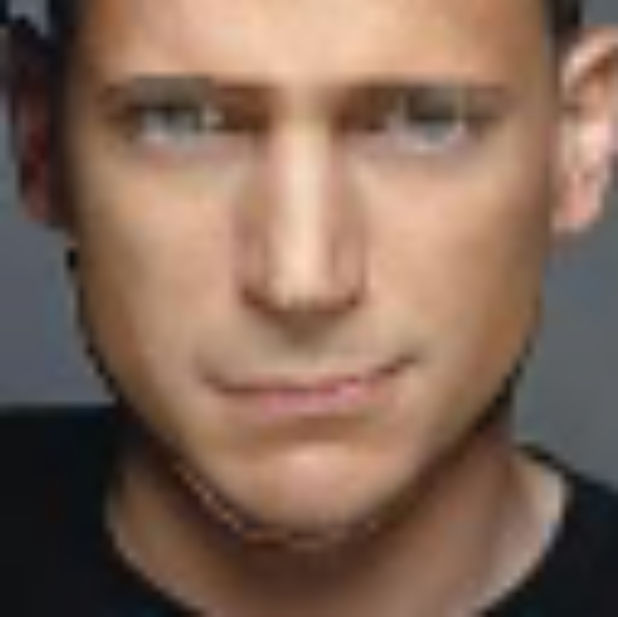}}
&
\subfloat{\includegraphics[trim={0 0.1cm 0.1cm 0},clip, width=0.1\linewidth, height=0.1\linewidth]{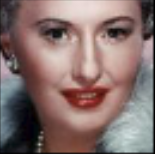}} 
\subfloat{\includegraphics[trim={0.1cm 0 0.1cm 0.1cm},clip, width=0.1\linewidth, height=0.1\linewidth]{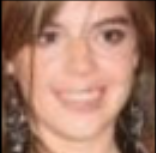}}
\subfloat{\includegraphics[trim={0.1cm 0.1cm 0.1cm 0},clip, width=0.1\linewidth, height=0.1\linewidth]{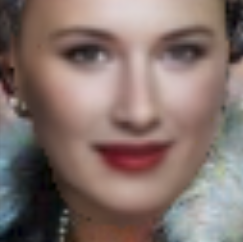}}
\subfloat{\includegraphics[trim={0.1cm 0.1cm 0.1cm 0.1cm},clip, width=0.1\linewidth, height=0.1\linewidth]{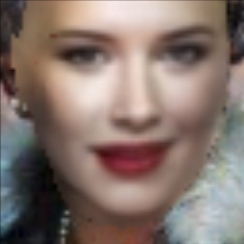}}&
\subfloat{\includegraphics[width=0.1\linewidth, height=0.1\linewidth]{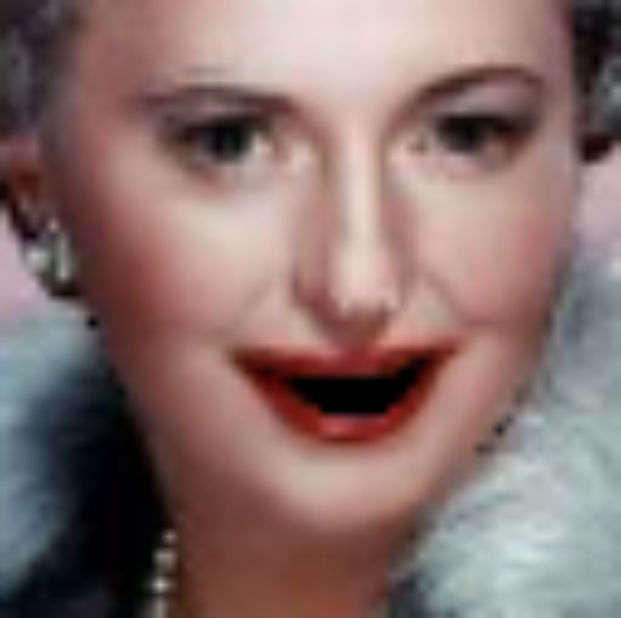}}
\\ \vspace{-9pt}
\subfloat{\includegraphics[width=0.1\linewidth, height=0.1\linewidth]{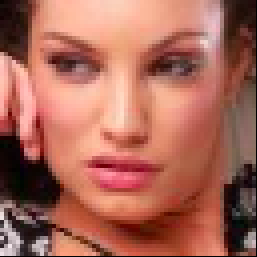}} 
\subfloat{\includegraphics[width=0.1\linewidth, height=0.1\linewidth]{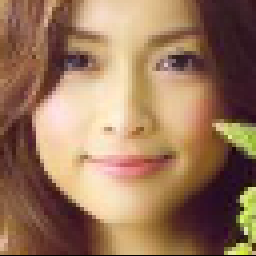}}
\subfloat{\includegraphics[width=0.1\linewidth, height=0.1\linewidth]{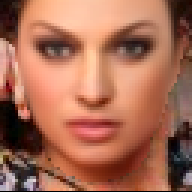}}
\subfloat{\includegraphics[width=0.1\linewidth, height=0.1\linewidth]{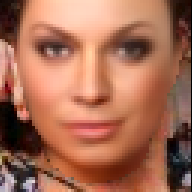}}&
\subfloat{\includegraphics[width=0.1\linewidth, height=0.1\linewidth]{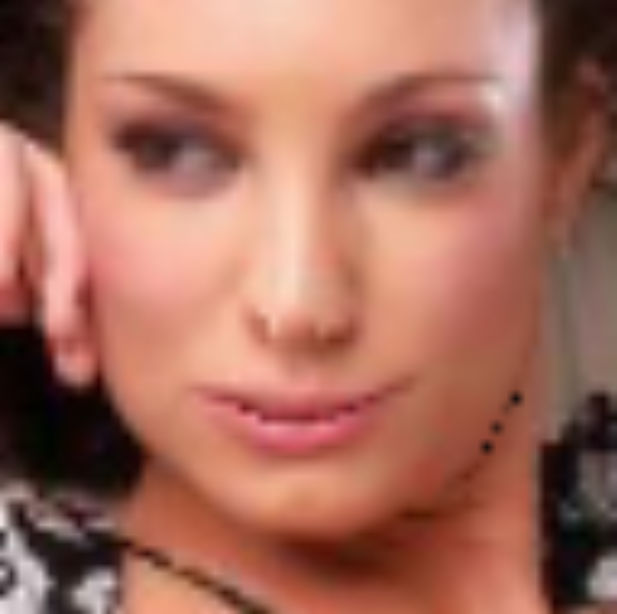}}
&
\subfloat{\includegraphics[width=0.1\linewidth, height=0.1\linewidth]{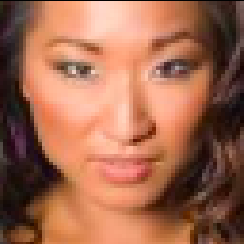}}  
\subfloat{\includegraphics[width=0.1\linewidth, height=0.1\linewidth]{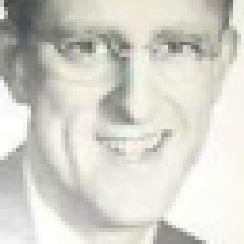}}
\subfloat{\includegraphics[width=0.1\linewidth, height=0.1\linewidth]{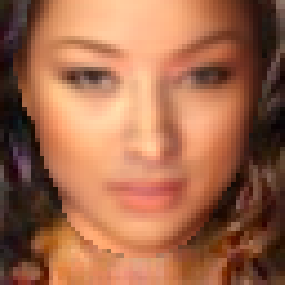}}
\subfloat{\includegraphics[width=0.1\linewidth, height=0.1\linewidth]{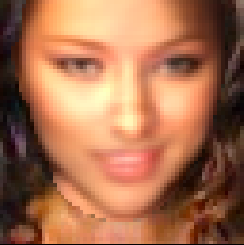}}&
\subfloat{\includegraphics[width=0.1\linewidth, height=0.1\linewidth]{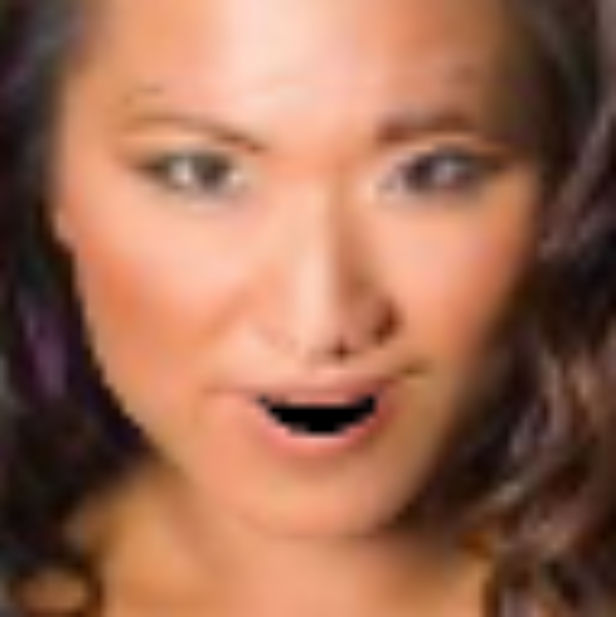}}
\end{tabular}
}
\caption{Our network is able to transfer the expression from one face to another by disentangling the expression components of the images. We compare our expression editing results with a baseline where a 3DMM has been fit to both input images.}
    \label{exp}
\end{figure*}

\begin{figure*}[t!]
\centering
\captionsetup[subfigure]{labelformat=empty, justification=centering,position=top}
{\def\arraystretch{0.5}\tabcolsep=1pt
\begin{tabular}{ cccc } 
\vspace{-9pt}
\subfloat[Original Image]{\includegraphics[trim={0 0.1cm 0 0},clip,width=0.08\linewidth]{121_orig}} 
\subfloat[Expression]{\includegraphics[trim={0.1cm 0 0 0},clip, width=0.08\linewidth, height=0.08\linewidth]{121_exp}}
\subfloat[Our Recon]{\includegraphics[trim={0.1cm 0 0 0},clip,width=0.08\linewidth, height=0.08\linewidth]{121_rec}}
\subfloat[Our Exp Edit]{\includegraphics[width=0.08\linewidth, height=0.08\linewidth]{121_modexp}}
&
\subfloat[B \& W]{\includegraphics[trim={0 0.3cm 0 0},clip, width=0.08\linewidth, height=0.08\linewidth]{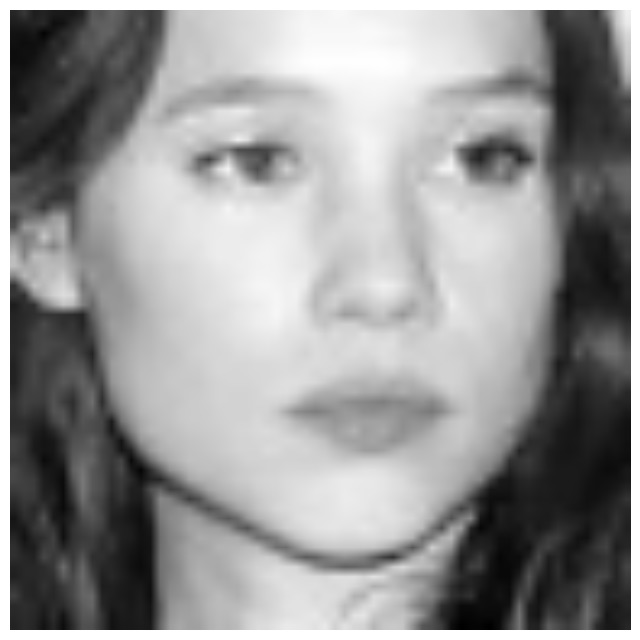}}
\subfloat[\cite{wang2017learning}]{\includegraphics[trim={0 0.3cm 0 0},clip, width=0.08\linewidth, height=0.08\linewidth]{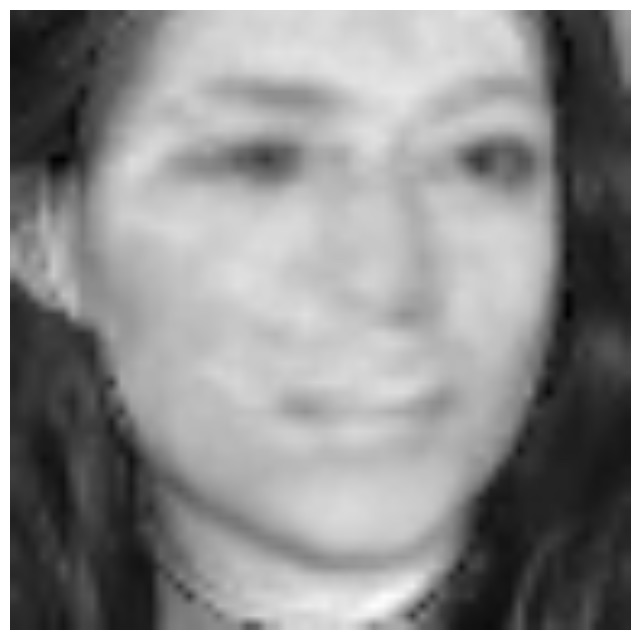}}
&
\subfloat[Original Image]{\includegraphics[width=0.08\linewidth, height=0.08\linewidth]{122_orig}} 
\subfloat[Expression]{\includegraphics[trim={0 0.1cm 0 0.1cm},clip, width=0.08\linewidth, height=0.08\linewidth]{122_exp}}
\subfloat[Our Recon]{\includegraphics[trim={0 0 0.2cm 0},clip, width=0.08\linewidth, height=0.08\linewidth]{122_rec}}
\subfloat[Our Exp Edit]{\includegraphics[trim={0 0.1cm 0 0.1cm},clip, width=0.08\linewidth, height=0.08\linewidth]{122_modexp}}
&
\subfloat[B \& W]{\includegraphics[trim={0 0.1cm 0 0.1cm},clip, width=0.08\linewidth, height=0.08\linewidth]{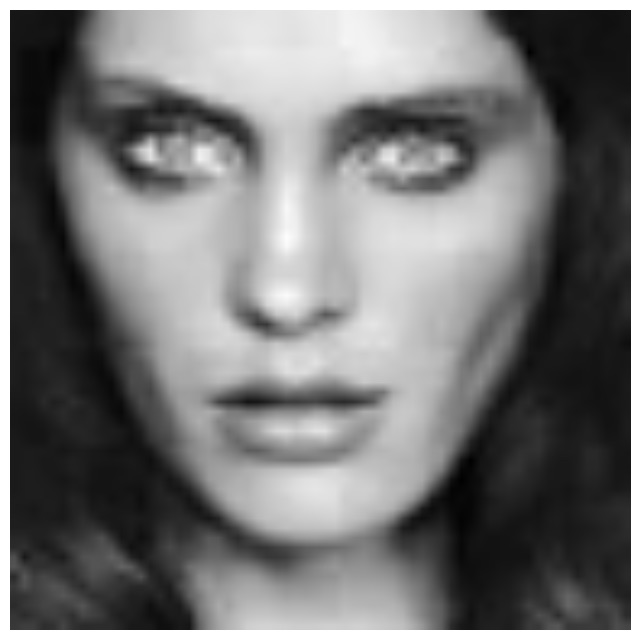}}
\subfloat[\cite{wang2017learning}]{\includegraphics[trim={0 0.1cm 0 0.1cm},clip, width=0.08\linewidth, height=0.08\linewidth]{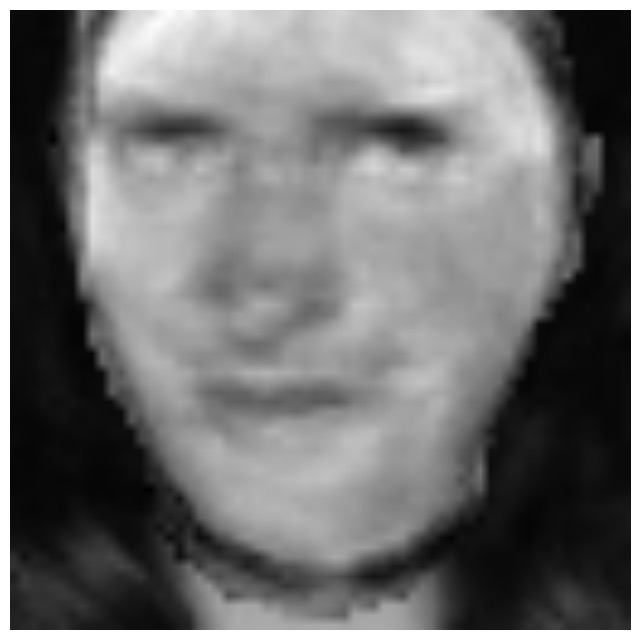}}
\\ \vspace{-9pt}
\subfloat{\includegraphics[width=0.08\linewidth, height=0.08\linewidth]{137_orig}} 
\subfloat{\includegraphics[width=0.08\linewidth, height=0.08\linewidth]{137_exp}}
\subfloat{\includegraphics[width=0.08\linewidth, height=0.08\linewidth]{137_rec}}
\subfloat{\includegraphics[width=0.08\linewidth, height=0.08\linewidth]{137_modexp}}&
\subfloat{\includegraphics[width=0.08\linewidth, height=0.08\linewidth]{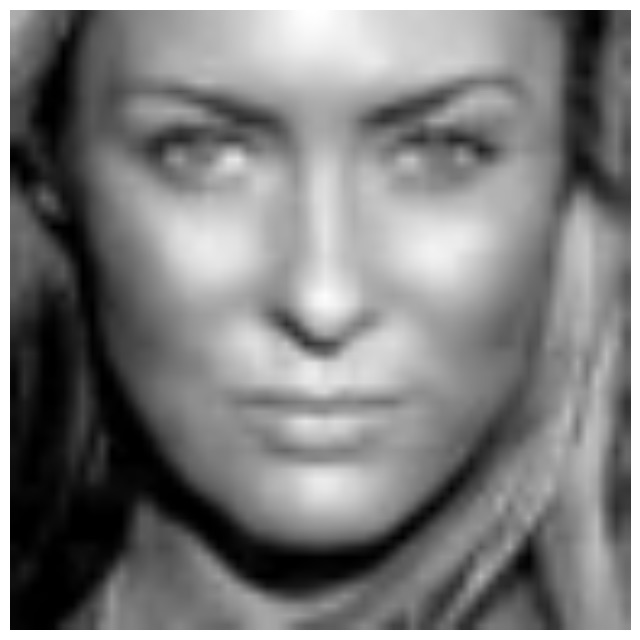}}
\subfloat{\includegraphics[width=0.08\linewidth, height=0.08\linewidth]{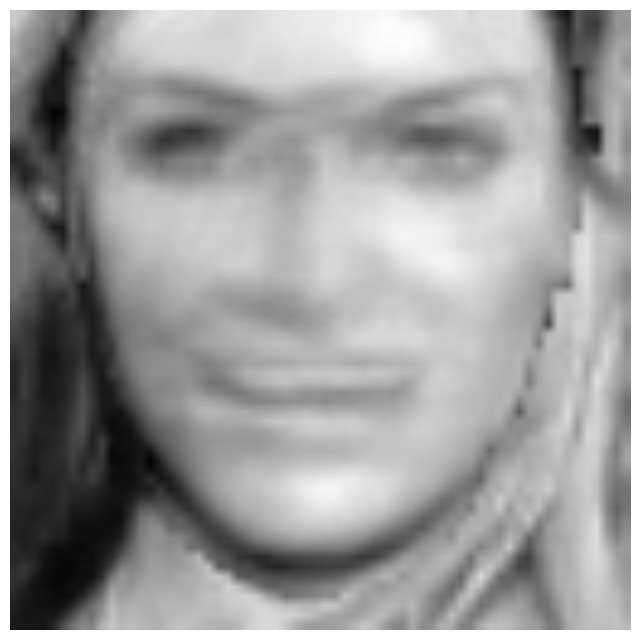}}
&
\subfloat{\includegraphics[width=0.08\linewidth, height=0.08\linewidth]{138_orig}} 
\subfloat{\includegraphics[width=0.08\linewidth, height=0.08\linewidth]{138_exp}}
\subfloat{\includegraphics[width=0.08\linewidth, height=0.08\linewidth]{138_rec}}
\subfloat{\includegraphics[width=0.08\linewidth, height=0.08\linewidth]{138_modexp}}&
\subfloat{\includegraphics[width=0.08\linewidth, height=0.08\linewidth]{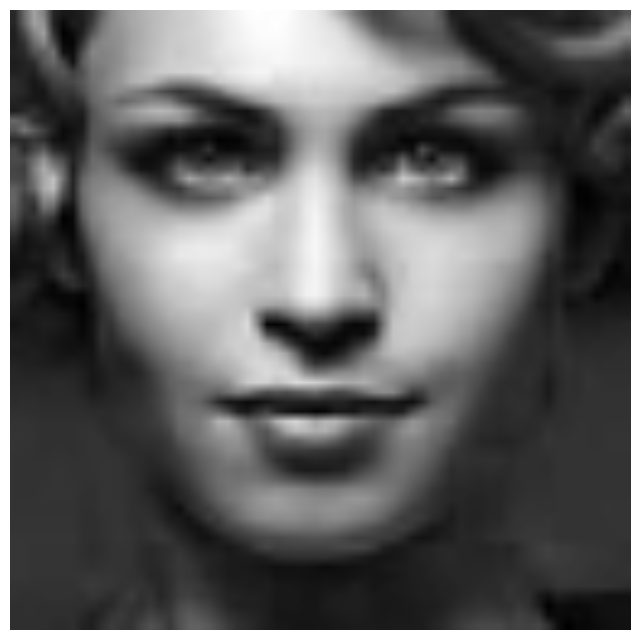}}
\subfloat{\includegraphics[width=0.08\linewidth, height=0.08\linewidth]{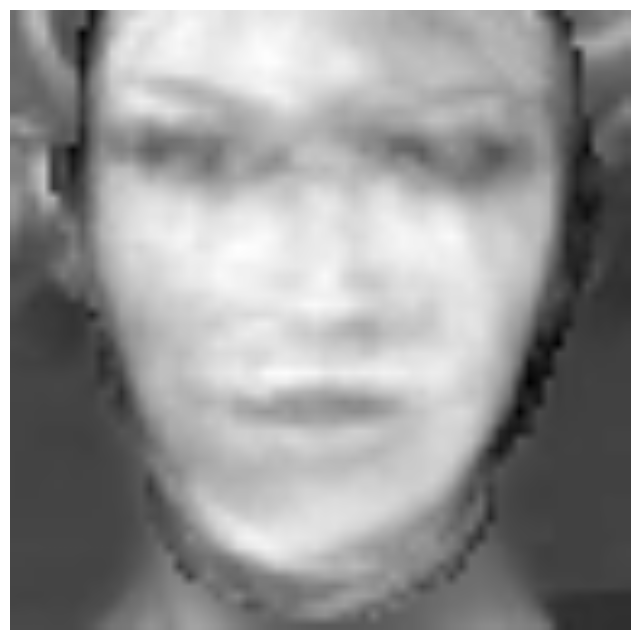}}
\\ \vspace{-9pt}
\subfloat{\includegraphics[trim={0 0.1cm 0.1cm 0.1cm},clip, width=0.08\linewidth, height=0.08\linewidth]{141_orig}} 
\subfloat{\includegraphics[trim={0 0.1cm 0.1cm 0.1cm},clip, width=0.08\linewidth, height=0.08\linewidth]{141_exp}}
\subfloat{\includegraphics[trim={0.1cm 0.1cm 0.1cm 0.1cm},clip,width=0.08\linewidth, height=0.08\linewidth]{141_rec}}
\subfloat{\includegraphics[trim={0.1cm 0.15cm 0.1cm 0.2cm},clip, width=0.08\linewidth, height=0.08\linewidth]{141_modexp}}&
\subfloat{\includegraphics[trim={0 0.45cm 0 0.35cm},clip, width=0.08\linewidth, height=0.08\linewidth]{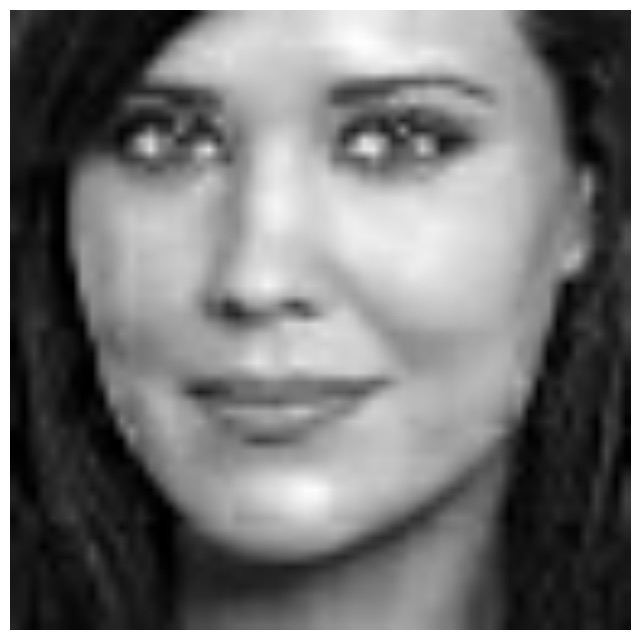}}
\subfloat{\includegraphics[trim={0 0.45cm 0 0.35cm},clip, width=0.08\linewidth, height=0.08\linewidth]{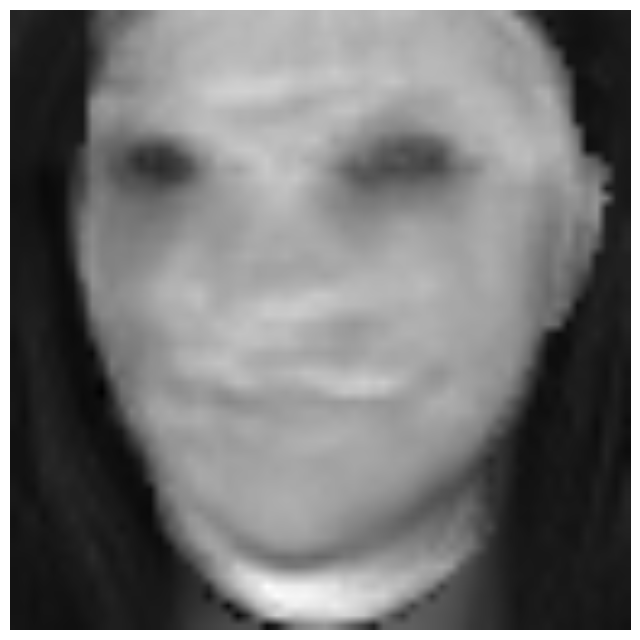}}
&
\subfloat{\includegraphics[width=0.08\linewidth, height=0.08\linewidth]{FF1253_orig}} 
\subfloat{\includegraphics[width=0.08\linewidth, height=0.08\linewidth]{FF1253_exp}}
\subfloat{\includegraphics[width=0.08\linewidth, height=0.08\linewidth]{FF1253_rec}}
\subfloat{\includegraphics[width=0.08\linewidth, height=0.08\linewidth]{FF1253_modexp}}&
\subfloat{\includegraphics[width=0.08\linewidth, height=0.08\linewidth]{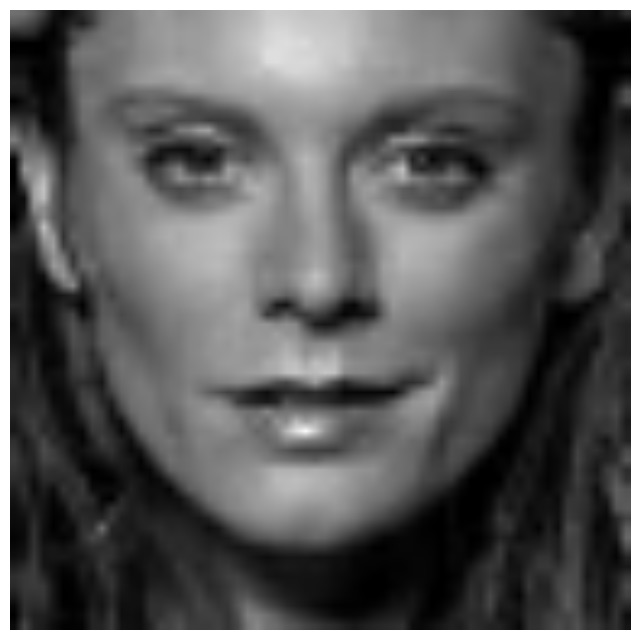}}
\subfloat{\includegraphics[width=0.08\linewidth, height=0.08\linewidth]{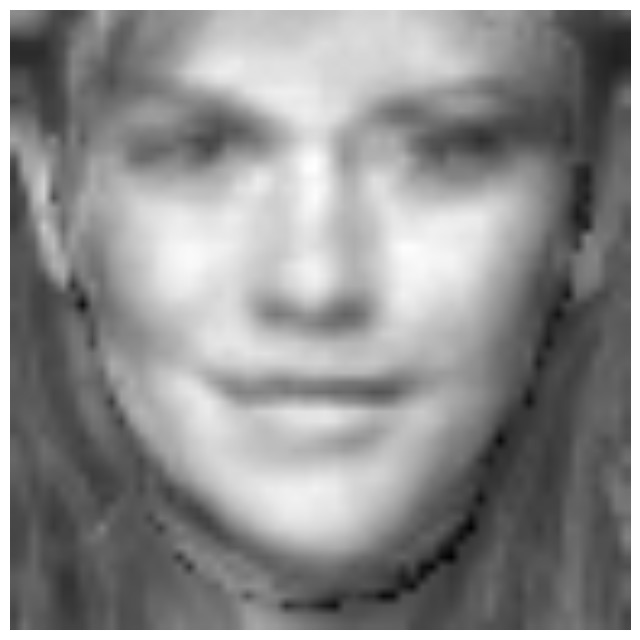}}
\\ \vspace{-9pt}
\subfloat{\includegraphics[width=0.08\linewidth, height=0.08\linewidth]{T7328_orig}} 
\subfloat{\includegraphics[width=0.08\linewidth, height=0.08\linewidth]{T7328_exp}}
\subfloat{\includegraphics[width=0.08\linewidth, height=0.08\linewidth]{T7328_rec}}
\subfloat{\includegraphics[width=0.08\linewidth, height=0.08\linewidth]{T7328_modexp}}&
\subfloat{\includegraphics[width=0.08\linewidth, height=0.08\linewidth]{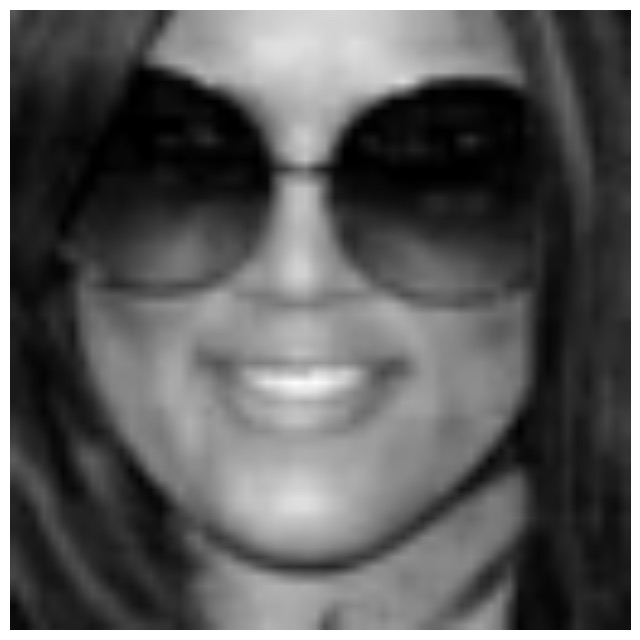}}
\subfloat{\includegraphics[width=0.08\linewidth, height=0.08\linewidth]{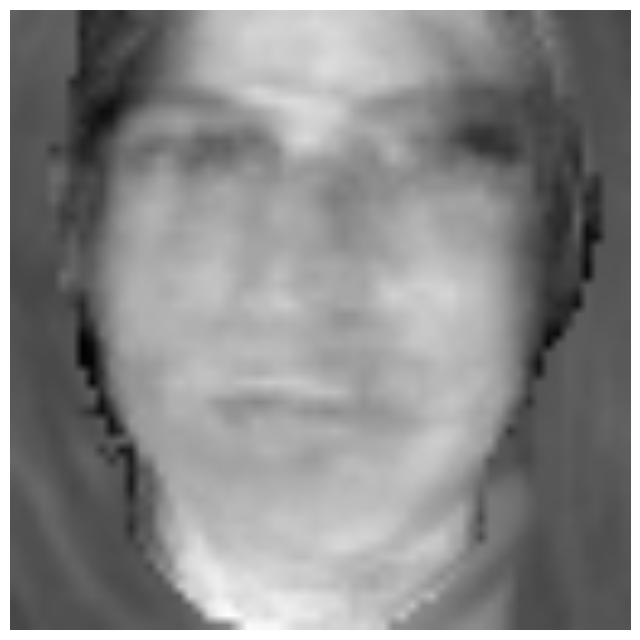}}
&
\subfloat{\includegraphics[width=0.08\linewidth, height=0.08\linewidth]{FF1246_orig}}  
\subfloat{\includegraphics[width=0.08\linewidth, height=0.08\linewidth]{FF1246_exp}}
\subfloat{\includegraphics[width=0.08\linewidth, height=0.08\linewidth]{FF1246_rec}}
\subfloat{\includegraphics[width=0.08\linewidth, height=0.08\linewidth]{FF1246_modexp}}&
\subfloat{\includegraphics[width=0.08\linewidth, height=0.08\linewidth]{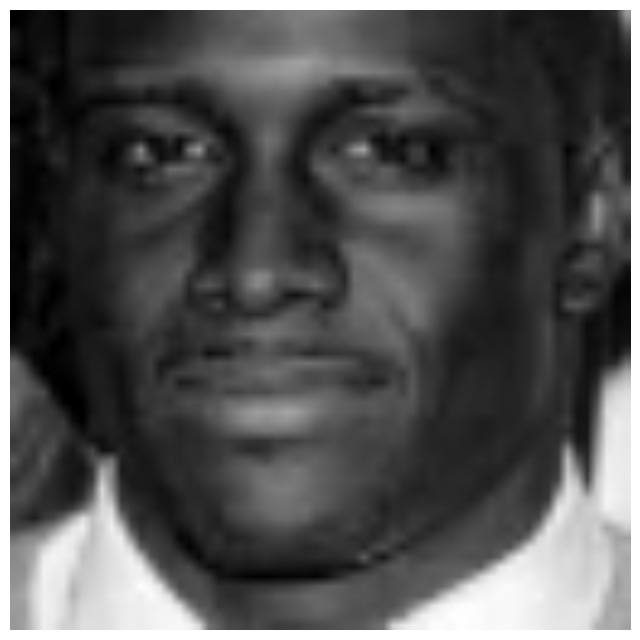}}
\subfloat{\includegraphics[width=0.08\linewidth, height=0.08\linewidth]{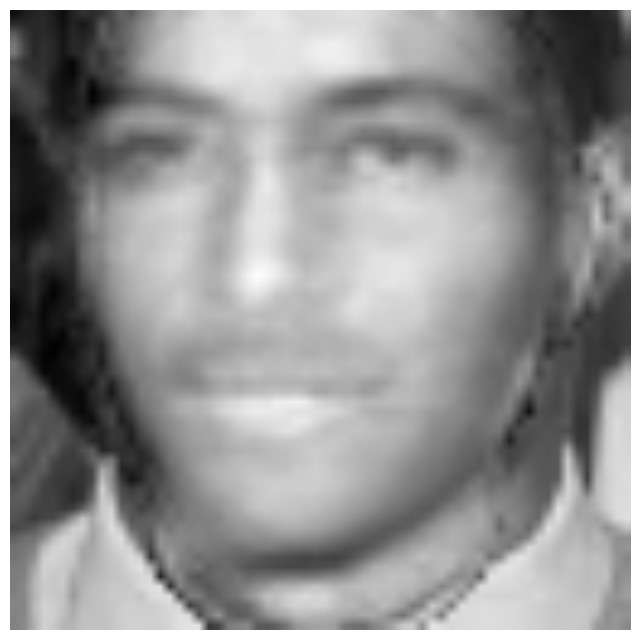}}
\end{tabular}
}
\caption{We compare our expression editing results with \cite{wang2017learning}. As \cite{wang2017learning} is not able to disentangle pose, editing expressions from images of different poses returns noisy results.}
    \label{expcomp}
\end{figure*}

\begin{figure*}[t!]
\captionsetup[subfigure]{labelformat=empty, justification=centering,position=top}
{\def\arraystretch{0.5}\tabcolsep=1pt
\begin{tabular}{ cccc } 
\vspace{-9pt}
\subfloat[Original Image]{\includegraphics[width=0.1\linewidth]{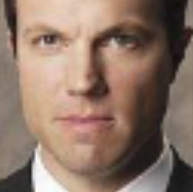}} 
\subfloat[Pose]{\includegraphics[width=0.1\linewidth]{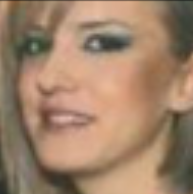}}
\subfloat[Our Recon]{\includegraphics[width=0.1\linewidth]{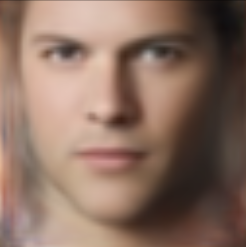}}
\subfloat[Our Pose Edit]{\includegraphics[width=0.1\linewidth]{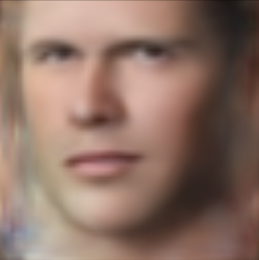}}&
\subfloat[Baseline]{\includegraphics[trim={0.1cm 0.8cm 0cm 1.5cm},clip, width=0.1\linewidth]{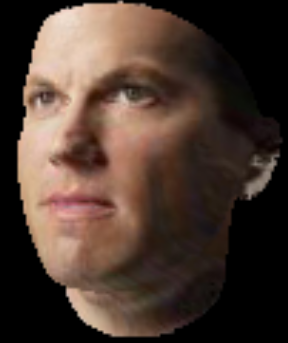}}
&
\subfloat[Original Image]{\includegraphics[width=0.1\linewidth]{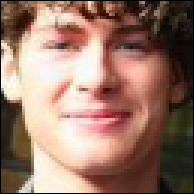}}
\subfloat[Pose]{\includegraphics[width=0.1\linewidth]{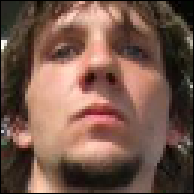}}
\subfloat[Our Recon]{\includegraphics[width=0.1\linewidth]{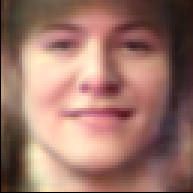}}
\subfloat[Our Pose Edit]{\includegraphics[width=0.1\linewidth]{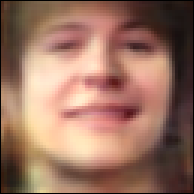}}&
\subfloat[Baseline]{\includegraphics[width=0.1\linewidth]{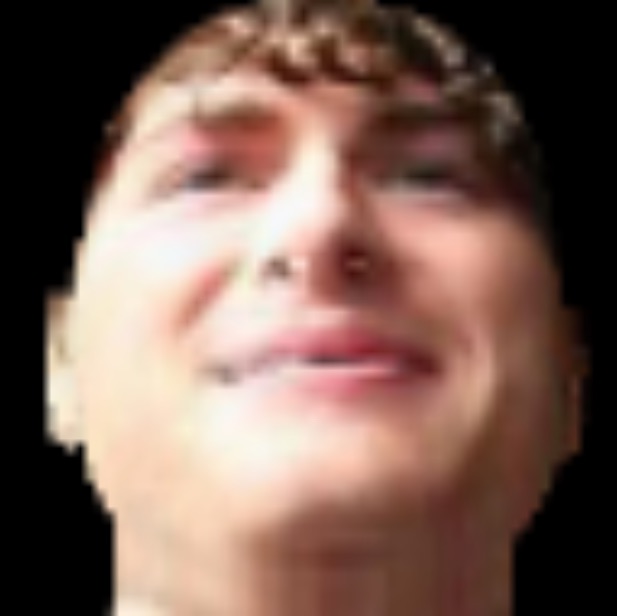}}
\\ \vspace{-9pt}
\subfloat{\includegraphics[width=0.1\linewidth]{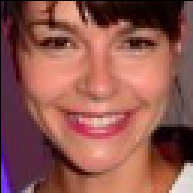}}
\subfloat{\includegraphics[width=0.1\linewidth]{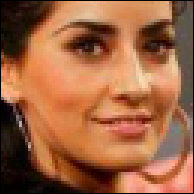}}
\subfloat{\includegraphics[width=0.1\linewidth]{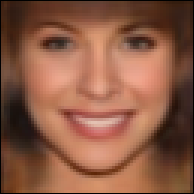}}
\subfloat{\includegraphics[width=0.1\linewidth]{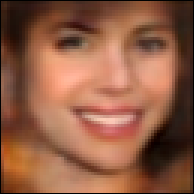}}&\subfloat{\includegraphics[width=0.1\linewidth]{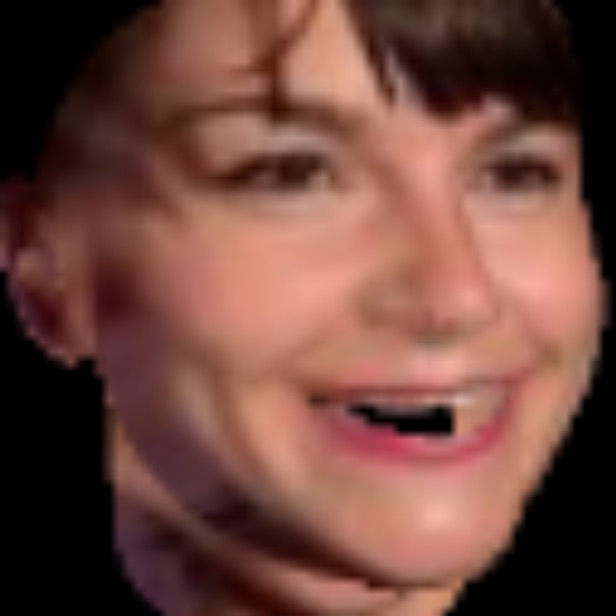}}
&
\subfloat{\includegraphics[width=0.1\linewidth]{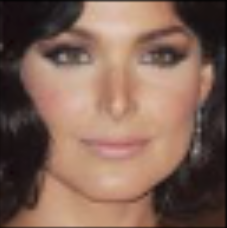}} 
\subfloat{\includegraphics[width=0.1\linewidth]{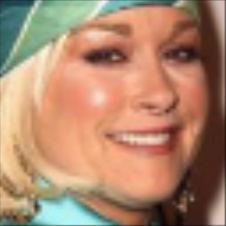}}
\subfloat{\includegraphics[width=0.1\linewidth]{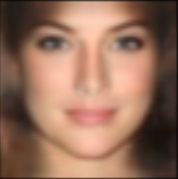}}
\subfloat{\includegraphics[width=0.1\linewidth]{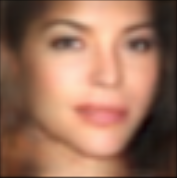}}&
\subfloat{\includegraphics[trim={0.1cm 0.5cm 0.1cm 0.5cm},clip, width=0.1\linewidth]{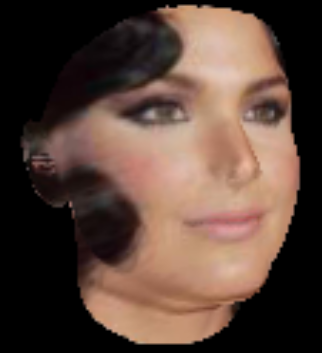}}
\\ \vspace{-9pt}
\subfloat{\includegraphics[width=0.1\linewidth]{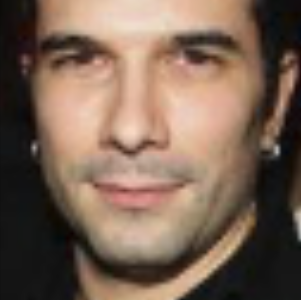}} 
\subfloat{\includegraphics[width=0.1\linewidth]{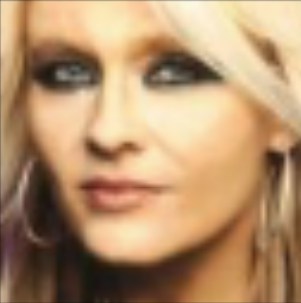}}
\subfloat{\includegraphics[width=0.1\linewidth]{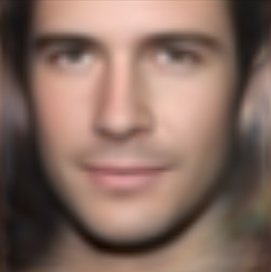}}
\subfloat{\includegraphics[width=0.1\linewidth]{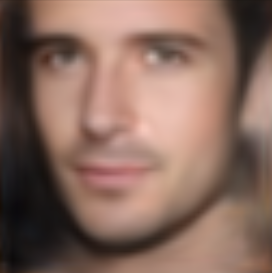}}&
\subfloat{\includegraphics[trim={0.1cm 1cm 0cm 2cm},clip, width=0.1\linewidth]{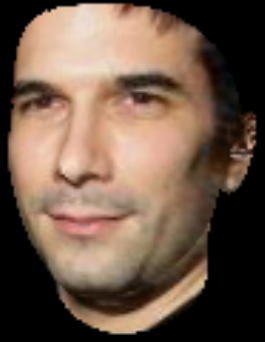}}
&
\subfloat{\includegraphics[width=0.1\linewidth]{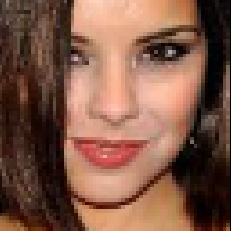}} 
\subfloat{\includegraphics[width=0.1\linewidth]{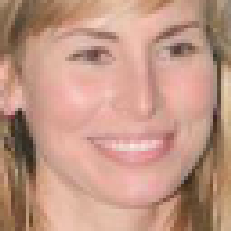}}
\subfloat{\includegraphics[width=0.1\linewidth]{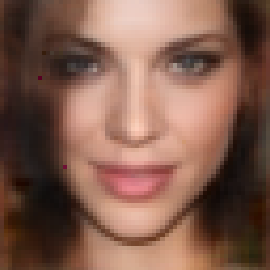}}
\subfloat{\includegraphics[width=0.1\linewidth]{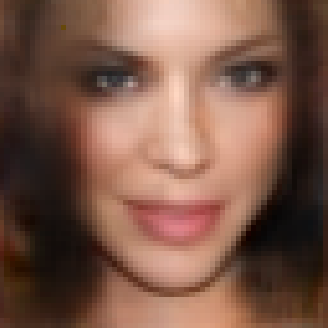}}&
\subfloat{\includegraphics[trim={0cm 0.4cm 0cm 1.5cm},clip, width=0.1\linewidth]{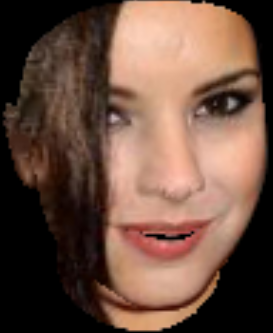}}
\end{tabular}
}
\caption{Our network is able to transfer the pose of one face to another by disentangling the pose components of the images. We compare our pose editing results with a baseline where a 3DMM has been fit to both input images. }
    \label{pose}
\end{figure*}

\begin{figure}[t!]
\captionsetup[subfigure]{labelformat=empty, justification=centering,position=top}
{\def\arraystretch{0.5}\tabcolsep=1pt
\begin{tabular}{ cc } 
\vspace{-9pt}
&
\subfloat[Source]{\includegraphics[trim={0.88cm 1cm 0cm 0cm},clip, width=0.17\linewidth]{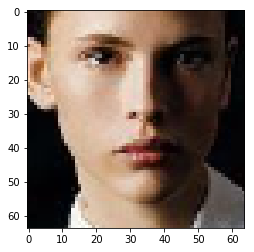}}
\subfloat[$\hat{\bm{s}}^{source}$]{\includegraphics[trim={0.88cm 1cm 0cm 0cm},clip,width=0.17\linewidth]{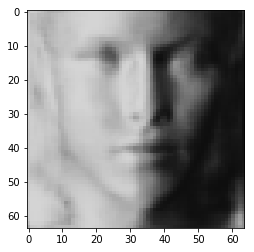}}
\subfloat[\cite{shu2017neural} $\hat{\bm{s}}^{source}$]{\includegraphics[width=0.17\linewidth]{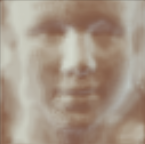}}
\\
Ours
&
\subfloat[Target]{\includegraphics[trim={0.88cm 1cm 0cm 0cm},clip,width=0.17\linewidth]{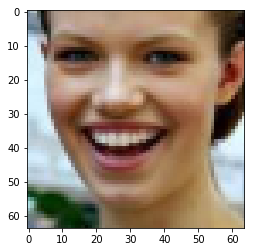}}
\subfloat[Reconstruction]{\includegraphics[trim={0.88cm 1cm 0cm 0cm},clip,width=0.17\linewidth]{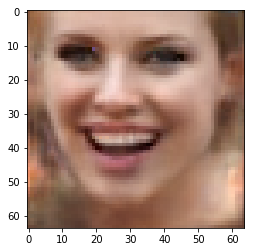}}
\subfloat[$\hat{\bm{s}}^{target}$]{\includegraphics[trim={0.88cm 1cm 0cm 0cm},clip,width=0.17\linewidth]{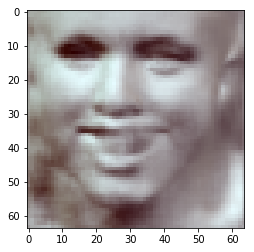}}
\subfloat[$\bm{s}^{transfer}$]{\includegraphics[trim={0.88cm 1cm 0cm 0cm},clip,width=0.17\linewidth]{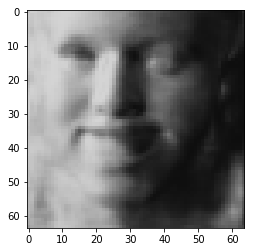}}
\subfloat[Result]{\includegraphics[trim={0.88cm 1cm 0cm 0cm},clip,width=0.17\linewidth]{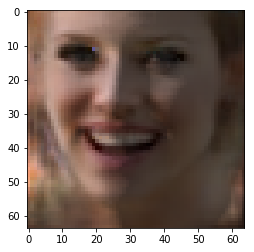}}
\\
\cite{shu2017neural}
&
\subfloat{\includegraphics[trim={0.88cm 1cm 0cm 0cm},clip,height=0.17\linewidth]{1_o}}
\subfloat{\includegraphics[height=0.17\linewidth,width=0.17\linewidth]{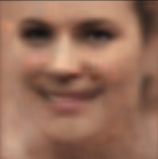}}
\subfloat{\includegraphics[height=0.17\linewidth,width=0.17\linewidth]{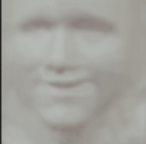}}
\subfloat{\includegraphics[height=0.17\linewidth,width=0.17\linewidth]{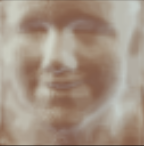}}
\subfloat{\includegraphics[height=0.17\linewidth,width=0.17\linewidth]{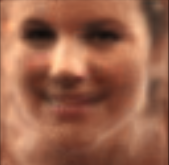}}
\\
Ours
&
\subfloat{\includegraphics[trim={0.88cm 1cm 0cm 0cm},clip,width=0.17\linewidth]{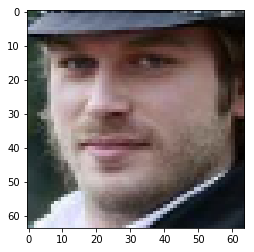}}
\subfloat{\includegraphics[trim={0.88cm 1cm 0cm 0cm},clip,width=0.17\linewidth]{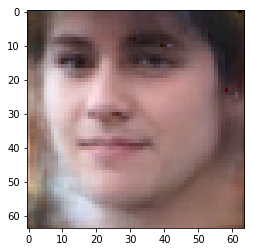}}
\subfloat{\includegraphics[trim={0.88cm 1cm 0cm 0cm},clip,width=0.17\linewidth]{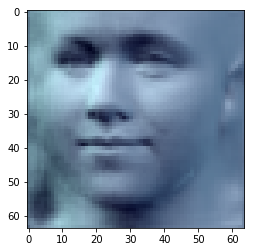}}
\subfloat{\includegraphics[trim={0.88cm 1cm 0cm 0cm},clip,width=0.17\linewidth]{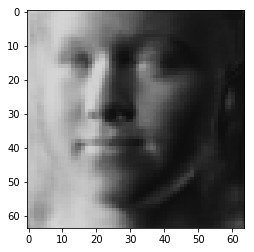}}
\subfloat{\includegraphics[trim={0.88cm 1cm 0cm 0cm},clip,width=0.17\linewidth]{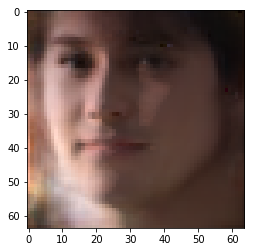}}
\\
\cite{shu2017neural}
&
\subfloat{\includegraphics[trim={0.88cm 1cm 0cm 0cm},clip,height=0.17\linewidth]{8_o}}
\subfloat{\includegraphics[height=0.17\linewidth,width=0.17\linewidth]{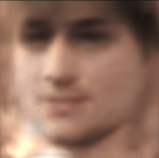}}
\subfloat{\includegraphics[height=0.17\linewidth,width=0.17\linewidth]{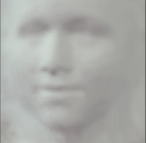}}
\subfloat{\includegraphics[height=0.17\linewidth,width=0.17\linewidth]{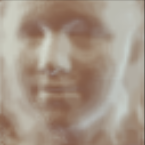}}
\subfloat{\includegraphics[height=0.17\linewidth,width=0.17\linewidth]{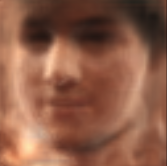}}
\end{tabular}
}
\caption{Using the illumination and normals estimated by our network, we are able to relight target faces using illumination from the source image. The source $\hat{\bm{s}}^{source}$ and target shading $\hat{\bm{s}}^{target}$ are displayed to visualise against the new transferred shading $\bm{s}^{transfer}$. We compare against~\cite{shu2017neural}.}
    \label{relight}
\end{figure}

\begin{figure}[!thb]
\captionsetup[subfigure]{labelformat=empty, justification=centering,position=top}
{\def\arraystretch{0.5}\tabcolsep=1pt
\begin{tabular}{ cc } 
\subfloat[Input]{\includegraphics[width=0.166\linewidth]{121_orig}}
\subfloat[Reconstruction]{\includegraphics[trim={4cm 2cm 4cm 2cm},clip,width=0.166\linewidth]{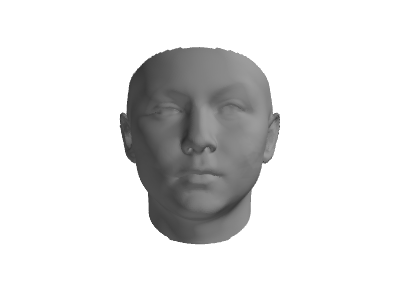}}
\subfloat[Ground Truth]{\includegraphics[trim={4cm 2cm 4cm 2cm},clip,width=0.166\linewidth]{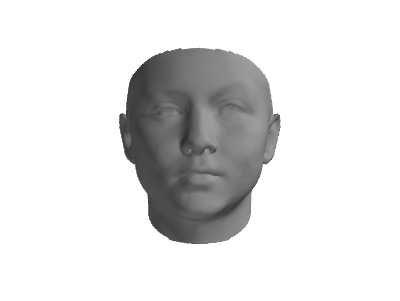}} &
\subfloat[Input]{\includegraphics[width=0.166\linewidth]{122_orig}}
\subfloat[Reconstruction]{\includegraphics[trim={4cm 2cm 4cm 2cm},clip,width=0.166\linewidth]{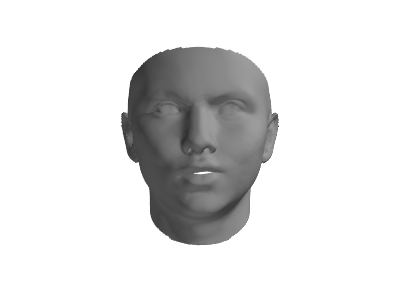}}
\subfloat[Ground Truth]{\includegraphics[trim={4cm 2cm 4cm 2cm},clip,width=0.166\linewidth]{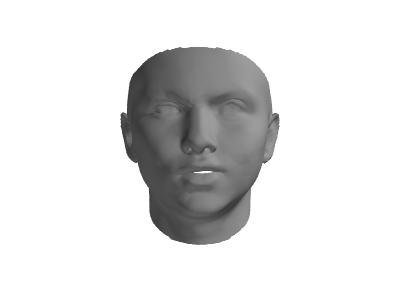}} \\
\subfloat{\includegraphics[width=0.166\linewidth]{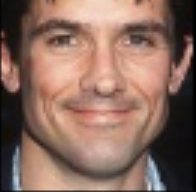}}
\subfloat{\includegraphics[trim={4cm 2cm 4cm 2cm},clip,width=0.166\linewidth]{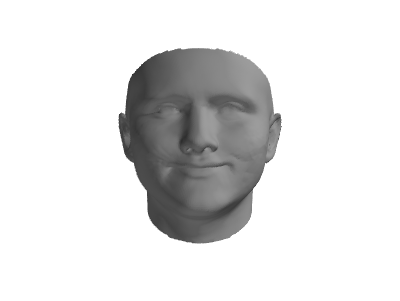}}
\subfloat{\includegraphics[trim={4cm 2cm 4cm 2cm},clip,width=0.166\linewidth]{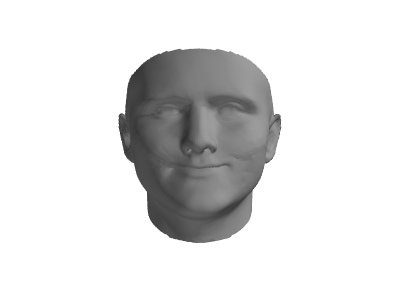}} 
& 

\subfloat{\includegraphics[width=0.166\linewidth]{137_orig}}
\subfloat{\includegraphics[trim={4cm 2cm 4cm 2cm},clip,width=0.166\linewidth]{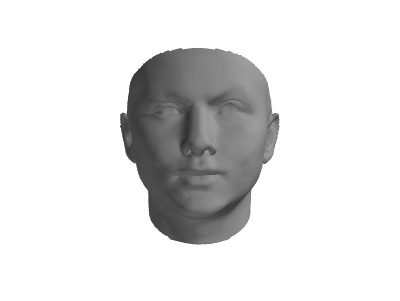}}
\subfloat{\includegraphics[trim={4cm 2cm 4cm 2cm},clip,width=0.166\linewidth]{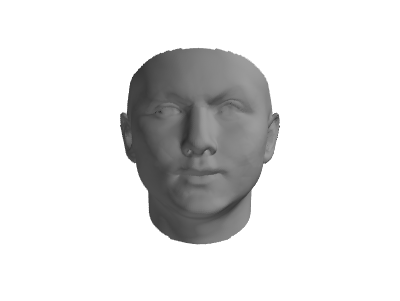}} 
\\

\subfloat{\includegraphics[width=0.166\linewidth]{141_orig}}
\subfloat{\includegraphics[trim={4cm 2cm 4cm 2cm},clip,width=0.166\linewidth]{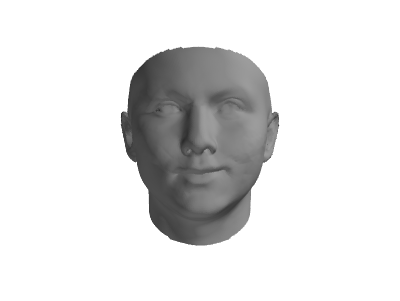}}
\subfloat{\includegraphics[trim={4cm 2cm 4cm 2cm},clip,width=0.166\linewidth]{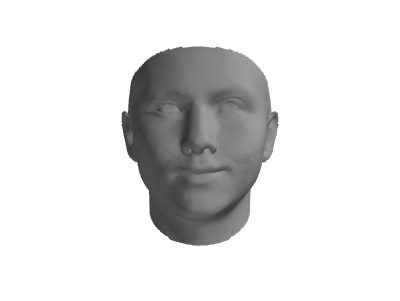}} 
&

\subfloat{\includegraphics[width=0.166\linewidth]{T7153_orig}}
\subfloat{\includegraphics[trim={4cm 2cm 4cm 2cm},clip,width=0.166\linewidth]{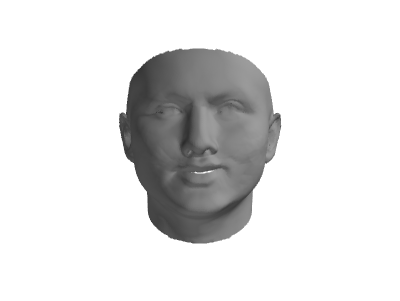}}
\subfloat{\includegraphics[trim={4cm 2cm 4cm 2cm},clip,width=0.166\linewidth]{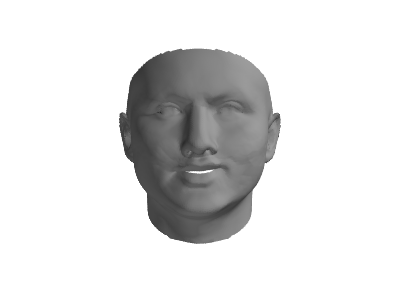}} \\

\subfloat{\includegraphics[width=0.166\linewidth]{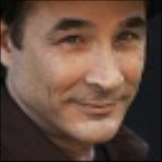}}
\subfloat{\includegraphics[trim={4cm 2cm 4cm 2cm},clip,width=0.166\linewidth]{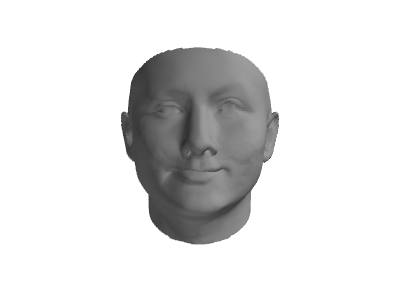}}
\subfloat{\includegraphics[trim={4cm 2cm 4cm 2cm},clip,width=0.166\linewidth]{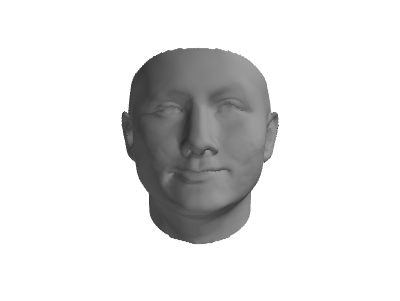}} &

\subfloat{\includegraphics[width=0.166\linewidth]{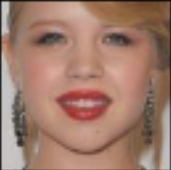}}
\subfloat{\includegraphics[trim={4cm 2cm 4cm 2cm},clip,width=0.166\linewidth]{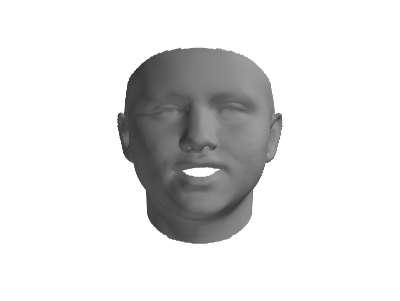}}
\subfloat{\includegraphics[trim={4cm 2cm 4cm 2cm},clip,width=0.166\linewidth]{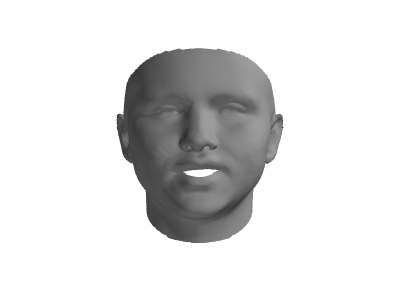}} 

\end{tabular}
}
\caption{Given a single image, we infer meaningful expression and identity components to reconstruct a 3D mesh of the face. We compare the reconstruction against the ground truth provided by 3DMM fitting.}
    \label{3d}
    \vspace{-10pt}
\end{figure}

\section{Proof of Concept Experiments}

We develop a lighter version of our proposed network, a proof-of-concept network (visualised in Figure~\ref{simplenetwork}), to show that our network is able to learn and disentangle pose, expression and identity.

In order to showcase the ability of the network, we leverage our newly proposed 4DFAB database \cite{cheng4dfab}, where subjects were invited to attend four sessions at different times in a span of five years. In each experiment session, the subject was asked to articulate 6 different facial expressions (\emph{anger, disgust, fear, happiness, sadness, surprise}), and we manually select the most expressive mesh (i.e. the apex frame) for this experiment. In total, 1795 facial meshes from 364 recording sessions (with 170 unique identities) are used. We keep 148 identities for training and leave 22 identities for testing. Note that there are no overlapping of identities between both sets. Within the training set, we synthetically augment each facial mesh by generating new facial meshes with 20 randomly selected expressions. Our training set contains in total 35900 meshes. The test set contains 387 meshes. For each mesh, we have the ground truth facial texture as well as expression and identity components of the 3DMM model. 

\subsection{Disentangling Expression and Identity}

We create frontal images of the facial meshes. Hence there is no illumination or pose variation in this training dataset. We train a lighter version of our network by removing the illumination and pose streams, a proof-of-concept network, visualised in Figure~\ref{simplenetwork}, on this synthetic dataset.

\subsubsection{Expression Editing}
We show the disentanglement between expression and identity by transferring the expression of one person to another. 

For this experiment, we work with unseen data (a hold-out set consisting of 22 unseen identities) and no labels. We first encode both input images $\bm{x}^i$ and $\bm{x}^j$:
\begin{equation}
\begin{aligned}
E(\bm{x}^i) &= \bm{z}_{exp}^i, \bm{z}_{id}^i,\\
E(\bm{x}^j) &= \bm{z}_{exp}^j, \bm{z}_{id}^j,
\end{aligned}
\end{equation}
where $E(\cdot)$ is our encoder and $\bm{z}_{exp}$ and $\bm{z}_{id}$ are the latent representations of expression and identity respectively. 

Assuming we want $\bm{x}^i$ to emulate the expression of $\bm{x}^j$, we decode on:
\begin{equation}
\begin{aligned}
D(\bm{z}_{exp}^j, \bm{z}_{id}^i) &= \bm{x}^{ji}, 
\end{aligned}
\end{equation}
where $D(\cdot)$ is our decoder. The resulting $\bm{x}^{ji}$ becomes our edited image where $\bm{x}^{i}$ has the expression of $\bm{x}^{j}$. Figure~\ref{synthexp} shows how the network is able to separate expression and identity. The edited images clearly maintain the identity while expression changes.

\subsubsection{3D Reconstruction and Facial Texture}
The latent variables $\bm{z}_{exp}$ and $\bm{z}_{id}$ that our network learns are extremely meaningful. Not only can they be used to reconstruct the image in 2D, but also they can be mapped into the expression ($\bm{x}_{exp}$) and identity ($\bm{x}_{id}$) components of a 3DMM model. This mapping is learnt inside the network. By replacing the expression and identity components of a mean face shape with $\hat{\bm{x}_{exp}}$ and $\hat{\bm{x}_{id}}$, we are able to reconstruct the 3D mesh of a face given a single input image. We compare these reconstructed meshes against the ground truth 3DMM used to create the input image in Figure~\ref{synth3d}. 

At the same time, the network is able to learn a mapping from $\bm{z}_{id}$ to facial texture. Therefore, we can predict the facial texture given a single input image. We compare the reconstructed facial texture with the ground truth facial texture in Figure~\ref{synthtex}.

\subsection{Disentangling Pose, Expression and Identity}
Our synthetic training set contains in total 35900 meshes. For each mesh, we have the ground truth facial texture as well as expression and identity components of the 3DMM, from which we create a corresponding image with one of 7 given poses. As there is no illumination variation in this training set, we train a proof-of-concept network by removing the illumination stream, visualised in Figure~\ref{simplenetwork}b, on this synthetic dataset. 

\subsubsection{Pose Editing}
We show the disentanglement between pose, expression and identity by transferring the pose of one person to another. 
Figure~\ref{synthpose} shows how the network is able to separate pose from expression and identity. This experiment highlights the ability of our proposed network to learn large pose variations even from profile to frontal faces.

\section{Experiments in-the-wild}

We train our network on in-the-wild data and perform several experiments on unseen data to show that our network is indeed able to disentangle illumination, pose, expression and identity. 

We edit expression or pose by swapping the latent expression/pose component learnt by the encoder $E$ (Eq.~\eqref{eq:encode}) with the latent expression/pose component predicted from another image. We feed the decoder $D$ (Eq.~\eqref{eq:decode}) with the modified latent component to retrieve our edited image.

\subsection{Expression and Pose Editing in-the-wild}

Given two in-the-wild images of faces, we are able to transfer the expression or pose of one person to another. Transferring the expression from two different facial images without fitting a 3D model is a very challenging problem. Generally, it is considered in the context of the same person under an elaborate blending framework \cite{yang2011expression} or by transferring certain classes of expressions \cite{sagonas2017robust}. 

For this experiment, we work with completely unseen data (a hold-out set of CelebA) and no labels. We first encode both input images $\bm{x}^i$ and $\bm{x}^j$:
\begin{equation}
\begin{aligned}
E(\bm{x}^i) &= \bm{z}_{exp}^i, \bm{z}_{id}^i, \bm{z}_{p}^i\\
E(\bm{x}^j) &= \bm{z}_{exp}^j, \bm{z}_{id}^j, \bm{z}_{p}^j,
\end{aligned}
\end{equation}
where $E(\cdot)$ is our encoder and $\bm{z}_{exp}$, $\bm{z}_{id}$, $\bm{z}_{p}$ are the latent representations of expression, identity and pose respectively. 

Assuming we want $\bm{x}^i$ to take on the expression or pose of $\bm{x}^j$, we then decode on:
\begin{equation}
\begin{aligned}
D(\bm{z}_{exp}^j, \bm{z}_{id}^i, \bm{z}_{p}^i) &= \bm{x}^{jii} \\
D(\bm{z}_{exp}^i, \bm{z}_{id}^i, \bm{z}_{p}^j)  &= \bm{x}^{iij},
\end{aligned}
\end{equation}
where $D(\cdot)$ is our decoder. 

The resulting $\bm{x}^{jii}$ then becomes our result image where $\bm{x}^{i}$ has the expression of $\bm{x}^{j}$. $\bm{x}^{jii}$ is the edited image where $\bm{x}^{i}$ changed to the pose of $\bm{x}^{j}$.

As there is currently no prior work for this expression editing experiment without fitting an AAM~\cite{cootes2001active} or 3DMM, we used the image synthesised by the 3DMM fitted models as a baseline, which indeed performs quite well. Compared with our method, other very closely related works~\cite{wang2017learning,shu2017neural} are not able to disentangle illumination, pose, expression and identity. In particular, \cite{shu2017neural} disentangles illumination of an image while \cite{wang2017learning} disentangles illumination, expression and identity from ``frontalised" images. Hence they are not able to disentangle pose. None of these methods can be applied to the expression/pose editing experiments on a dataset that contains pose variations such as CelebA. If \cite{wang2017learning} is applied directly on our test images, it would not be able to perform expression editing well, as shown by Figure~\ref{expcomp}.

For the 3DMM baseline, we fit a shape model to both images and extract the expression components of the model. We then generate a new face shape using the expression components of one face and the identity components of another face in the same 3DMM setting. This technique has much higher overhead than our proposed method as it requires time-consuming 3DMM fitting of the images. Our expression editing results and the baseline results are shown in Figure~\ref{exp}. Though the baseline is very strong, it does not change the texture of the face which can produce unnatural looking faces shown with original expression. Also, the baseline method can not fill up the inner mouth area. Our editing results show more natural looking faces. 

For pose editing, the background is unknown once the pose has changed, thus, for this experiment, we mainly focus on the face region. Figure~\ref{pose} shows our pose editing results. For the baseline method, we fit a 3DMM to both images and estimate the rotation matrix. We then synthesise $\bm{x}_{i}$ with the rotation of $\bm{x}_{j}$. This technique has high overhead as it requires expensive 3DMM fitting of the images. 

\begin{figure}[t!]
\centering
\captionsetup[subfigure]{labelformat=empty, justification=centering,position=top}
\subfloat{\includegraphics[width=1\linewidth]{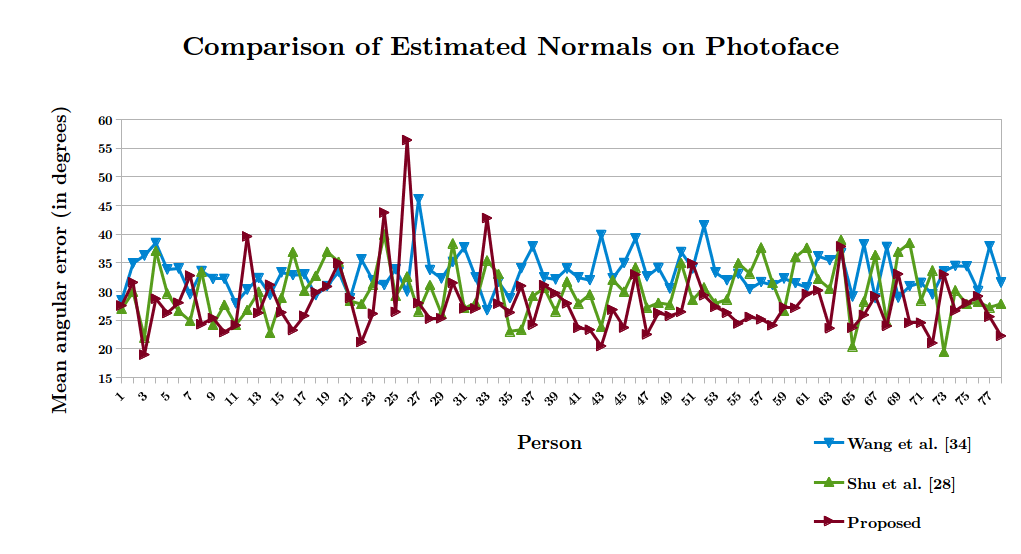}}
\caption{Comparison of the estimated normals obtained using the proposed model vs the ones obtained by \cite{wang2017learning} and \cite{shu2017neural}.}
\label{normalcomp}
\end{figure}

\begin{figure*}[t!]
\centering
\captionsetup[subfigure]{labelformat=empty, justification=centering,position=top}
\subfloat[$\bm{Z}_{exp}$]{\includegraphics[trim={0cm 1cm 4cm 1cm},clip, height=0.21\linewidth]{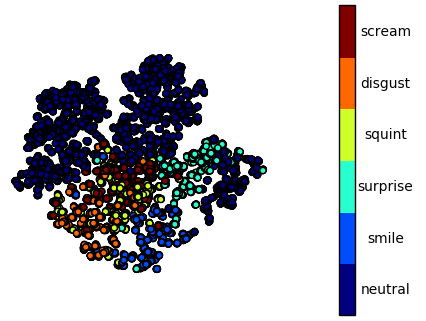}}
\subfloat[$\bm{Z}_0$]{\includegraphics[height=0.21\linewidth]{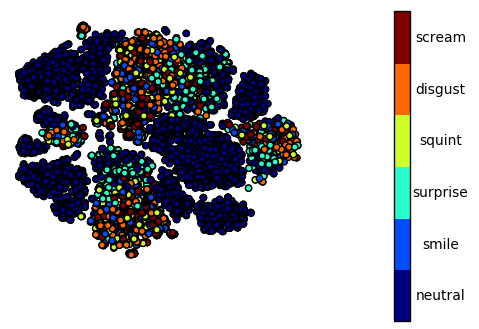}}
\caption{Visualisation of our $\bm{Z}_{exp}$ and baseline $\bm{Z}_{0}$ using t-SNE. Our latent $\bm{Z}_{exp}$ clusters better with regards to expression than the latent space $\bm{Z}_{0}$ of an auto-encoder.}
\label{zexp}
\vspace{-10pt}
\end{figure*}

\begin{figure*}[t!]
\centering
\captionsetup[subfigure]{labelformat=empty, justification=centering,position=top}
\subfloat[$\bm{Z}_{p}$]{\includegraphics[trim={0cm 0cm 3cm 0cm},clip, height=0.21\linewidth]{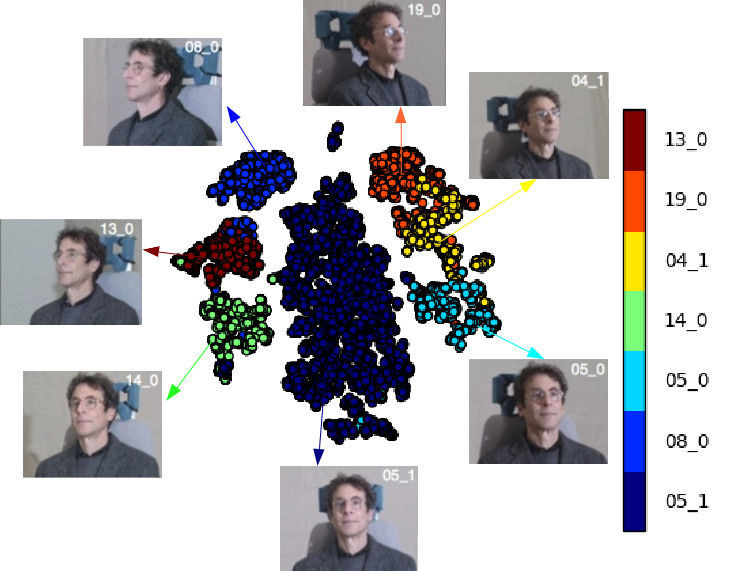}}
\subfloat[$\bm{Z}_0$]{\includegraphics[height=0.21\linewidth]{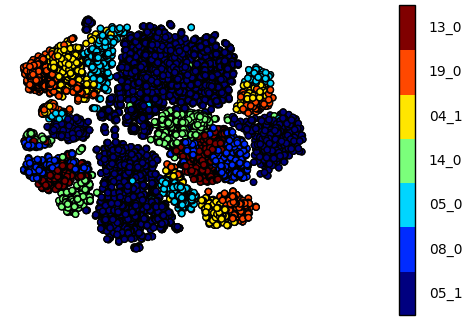}}
\caption{Visualisation of our $\bm{Z}_{p}$ and baseline $\bm{Z}_{0}$ using t-SNE. It is evident that the proposed disentangled $\bm{Z}_{p}$ clusters better with regards to pose than the latent space $\bm{Z}_{0}$ of an auto-encoder.}
\label{zpose}
\vspace{-10pt}
\end{figure*}

\subsection{Illumination Editing}

We transfer illumination by estimating the normals $\hat{\bm{n}}$, albedo $\hat{\bm{a}}$ and illumination components $\hat{\bm{l}}$ of the source ($\bm{x}^{source}$) and target ($\bm{x}^{target}$) images. Then we use $\hat{\bm{n}}^{target}$ and $\hat{\bm{l}}^{source}$ to compute the transferred shading $\bm{s}^{transfer}$ and multiply the new shading by $\hat{\bm{a}}^{target}$ to create the relighted image result $\bm{x}^{transfer}$. In Figure~\ref{relight} we show the performance of our method and compare against~\cite{shu2017neural} on illumination transfer. We observe that our method outperforms~\cite{shu2017neural} as we obtain more realistic looking results. We include further comparison images with ~\cite{shu2017neural} in the supplemental material.

\subsection{3D Reconstruction}

The latent variables $\bm{z}_{exp}$ and $\bm{z}_{id}$ that our network learns are extremely meaningful. Not only can they be used to reconstruct the image in 2D, they can be mapped into the expression ($\bm{x}_{exp}$) and identity ($\bm{x}_{id}$) components of a 3DMM. This mapping is learnt inside the network. By replacing the expression and identity components of a mean face shape with $\hat{\bm{x}_{exp}}$ and $\hat{\bm{x}_{id}}$, we are able to reconstruct the 3D mesh of a face given a single in-the-wild 2D image. We compare these reconstructed meshes against the fitted 3DMM to the input image. 

The results of the experiment are visualised in Figure~\ref{3d}. We observe that the reconstruction is very close to the ground truth. Both techniques though do not capture well the identity of the person in the input image due to a known weakness in 3DMM. 

\subsection{Normal Estimation}

\begin{table}
    \centering
    \begin{tabular}{ || l | r | r | r ||  }
      \hline			
     \textbf{Method} & \textbf{Mean$\pm$Std against~\cite{Woodham1980}} & \textbf{\textless 35$\degree$} & \textbf{\textless 40$\degree$} \\
      \hline
      \cite{wang2017learning} & 33.37\degree $\pm$ \hspace{1ex}3.29\degree & 75.3\% & 96.3\%\\
      \cite{shu2017neural} & 30.09\degree $\pm$ \hspace{1ex}4.66\degree & 84.6\% & 98.1\%\\
      \textbf{Proposed} & \textbf{28.67\degree $\pm$ \hspace{1ex}5.79\degree} & \textbf{89.1\%} & \textbf{96.3\%}\\
      \hline  
    \end{tabular}
    \caption{Angular error for the various surface normal estimation methods on the Photoface~\cite{Zafeiriou2013} dataset}
    \label{table:normal_table}
\end{table}

We evaluate our method on the surface normal estimation task on the Photoface~\cite{Zafeiriou2013} dataset which has information about illumination. Assuming the normals found using calibrated Photometric Stereo~\cite{Woodham1980} as ``ground truth'', we calculate the angular error between our estimated normals and the ``ground truth''.
Figure~\ref{normalcomp} and Table~\ref{table:normal_table} quantitatively evaluates our proposed method against prior works \cite{wang2017learning,shu2017neural} in the normal estimation task. We observe that our proposed method performs on par or outperforms previous methods.

\subsection{Quantitative Evaluation of the Latent Space}

We want to test whether our latent space corresponds well to the variation that it is supposed to learn. For our quantitative experiment, we used Multi-PIE~\cite{Gross2010} as our test dataset. This dataset contains labelled variations in identity, expressions and pose. Disentanglement of variations in Multi-PIE is particularly challenging as its images are captured under laboratory conditions which is quite different from that of our training images. As a matter of fact, the expressions contained in Multi-PIE do not correspond to the 7 basic expressions and can be easily confused.

We encoded 10368 images of the Multi-PIE dataset with 54 identities, 6 expressions and 7 poses and trained a linear SVM classifier using 90\% of the identity labels and the latent variables $\bm{z}_{id}$. We then test on the remaining 10\% $\bm{z}_{id}$ to check whether they are discriminative for identity classification. We use 10-fold cross-validation to evaluate the accuracy of the learnt classifier. We repeat this experiment for expression with $\bm{z}_{exp}$ and pose with $\bm{z}_{p}$ respectively. Our results in Table~\ref{acc} show that our latent representation is indeed discriminative. This experiment showcases the discriminative power of our latent representation on a previously unseen dataset. In order to quantitatively compare with~\cite{wang2017learning}, we run another experiment on only frontal images of the dataset with 54 identities, 6 expressions and 16 illuminations. The results in Table~\ref{acc_comp} shows how our proposed model outperforms~\cite{wang2017learning} in these classification tasks. Our latent representation has stronger discriminative power than the one learnt by~\cite{wang2017learning}.

\begin{table}[!thb]
\centering
\begin{tabular}{ |c|c|c|c| } 
 \hline
  & \bm{$z_{identity}$} & \bm{$z_{expression}$} & \bm{$z_{pose}$} \\ \hline
Accuracy & 83.85\% & 86.07\% & 95.73\%  \\ 
 \hline
\end{tabular}
\caption{Classification accuracy results: we try to classify 54 identities using $\bm{z}_{id}$, 6 expressions using $\bm{z}_{exp}$ and 7 poses using $\bm{z}_{p}$.}
\label{acc}
\vspace{-10pt}
\end{table}

\begin{table}[!thb]
\centering
\begin{tabular}{ |c|c|c| } 
 \multicolumn{1}{c}{}&\multicolumn{2}{c}{Identity} 
 \\  \hline
  & \bm{$z_{identity}$} & \bm{$C$} \cite{wang2017learning} 
\\ \hline
 Accuracy & \textbf{99.33\%} & 19.18 \% 
\\ 
 \hline
\end{tabular}
\begin{tabular}{ |c|c|c| } 
 \multicolumn{1}{c}{} & \multicolumn{2}{c}{Expression} 
 \\  \hline
  & \bm{$z_{expression}$} & \bm{$E$} \cite{wang2017learning} 
\\ \hline
 Accuracy 
 & \textbf{78.92\%} & 35.49 \\ 
 \hline
\end{tabular}
\begin{tabular}{ |c|c|c| } 
 \multicolumn{1}{c}{}& \multicolumn{2}{c}{Illumination}
 \\  \hline
 &\bm{$z_{illumination}$}   & \bm{$L$} \cite{wang2017learning}\\ \hline
 Accuracy 
 & \textbf{64.11\%} & 48.85\%\\ 
 \hline
\end{tabular}
\caption{Classification accuracy results in comparison with~\cite{wang2017learning}: As~\cite{wang2017learning} works on frontal images, we only consider frontal images in this experiment. We try to classify 54 identities using $\bm{z}_{id}$ vs. $\bm{C}$, 6 expressions using $\bm{z}_{exp}$ vs. $\bm{E}$ and 16 illumination using $\bm{z}_{ill}$ vs. $\bm{L}$.}
\label{acc_comp}
\end{table}

We visualise, using t-SNE \cite{maaten2008visualizing}, the latent $\bm{Z}_{exp}$ and  $\bm{Z}_{p}$ encoded from Multi-PIE according to their expression and pose label and compare against the latent representation $\bm{Z}_0$ learnt by an in-house large-scale adversarial auto-encoder of similar architecture trained with 2 million faces \cite{makhzani2015adversarial}. Figures~\ref{zexp} and~\ref{zpose} show that even though our encoder has not seen any images of Multi-PIE, it manages to create informative latent representations that cluster well expression and pose (contrary to the representation learned by the tested auto-encoder).


\section{Conclusion}
We proposed the first, to the best of our knowledge, attempt to jointly disentangle modes of variation that correspond to expression, identity, illumination and pose using no explicit labels regarding these attributes. More specifically, we proposed the first, as far as we know, approach that combines a powerful Deep Convolutional Neural Network (DCNN) architecture with unsupervised tensor decompositions. We demonstrate the power of our methodology in expression and pose transfer, as well as discovering powerful features for pose and expression classification. 

{\small
\bibliographystyle{ieee}
\bibliography{egbib}
}

\clearpage
\appendix
\renewcommand\thesection{\Alph{section}}
\section{Network Details}

The convolutional encoder stack (Fig.~\ref{network}) is composed of three convolutions with $96 \ast 5 \times 5$, $48 \ast 5 \times 5$ and $24 \ast 5 \times 5$ filter sets. Each convolution is followed by max-pooling and a thresholding nonlinearity. We pad the filter responses so that the final output of the convolutional stack is a set of filter responses with size $24 \ast 8 \times 8$ for an input image $3 \ast 64 \times 64$. The pooling indices of the max-pooling are preserved for the unpooling layers in the decoder stack.

The decoder stacks for the mask and background are strictly symmetric to the encoder stack and have skip connections to the input encoder
stack at the corresponding unpooling layers. These skip connections
between the encoder and the decoder allow for the details
of the background to be preserved. 

The other decoder stacks use upsampling and are also strictly symmetric to the encoder stack. 


\begin{figure*}[t!]
\vspace{-10pt}
\captionsetup[subfigure]{labelformat=empty, justification=centering,position=top}
{\def\arraystretch{0.5}\tabcolsep=1pt
\begin{tabular}{ cccc } 
\vspace{-9pt}
\subfloat[Original Image]{\includegraphics[width=0.1\linewidth]{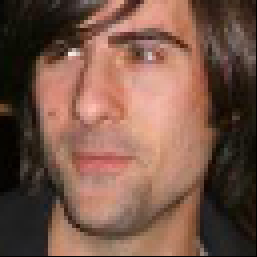}} 
\subfloat[Expression]{\includegraphics[width=0.1\linewidth]{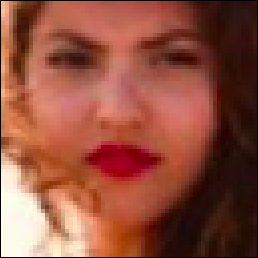}}
\subfloat[Recon]{\includegraphics[width=0.1\linewidth]{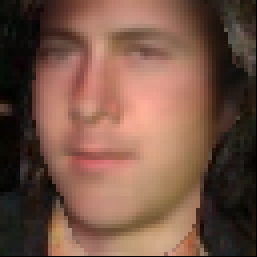}}
\subfloat[Our Exp Edit]{\includegraphics[width=0.1\linewidth]{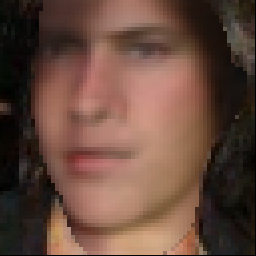}}&
\subfloat[Baseline]{\includegraphics[width=0.1\linewidth]{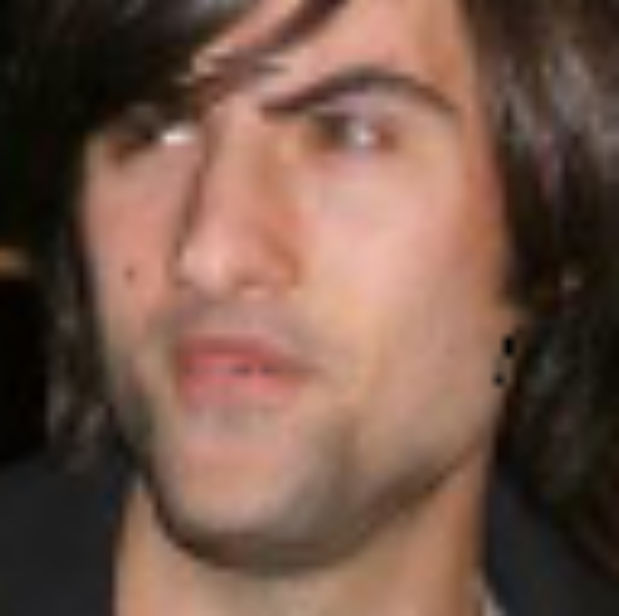}}
&
\subfloat[Original Image]{\includegraphics[width=0.1\linewidth, height=0.1\linewidth]{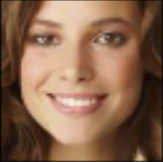}} 
\subfloat[Expression]{\includegraphics[width=0.1\linewidth, height=0.1\linewidth]{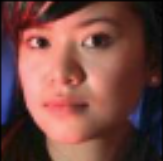}}
\subfloat[Recon]{\includegraphics[width=0.1\linewidth, height=0.1\linewidth]{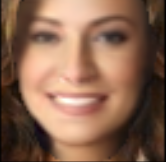}}
\subfloat[Our Exp Edit]{\includegraphics[width=0.1\linewidth, height=0.1\linewidth]{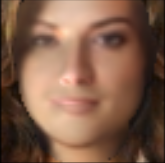}}&
\subfloat[Baseline]{\includegraphics[width=0.1\linewidth, height=0.1\linewidth]{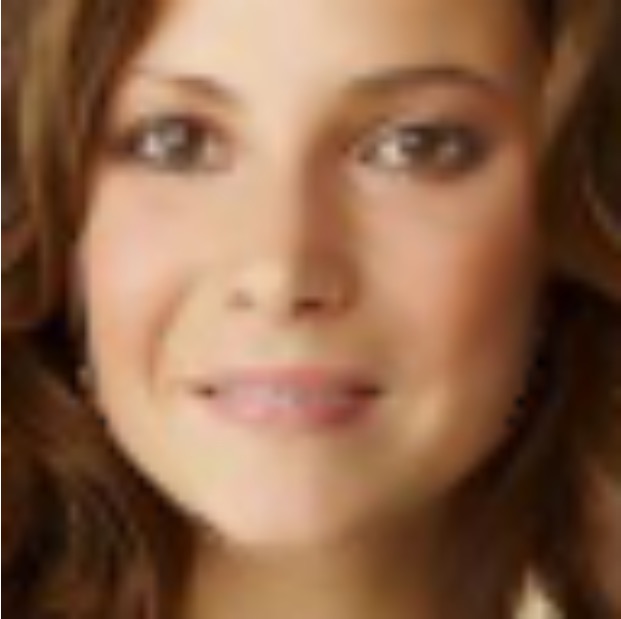}}
\\ \vspace{-9pt}
\subfloat{\includegraphics[width=0.1\linewidth, height=0.1\linewidth]{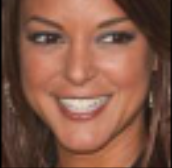}} 
\subfloat{\includegraphics[width=0.1\linewidth, height=0.1\linewidth]{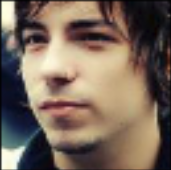}}
\subfloat{\includegraphics[width=0.1\linewidth, height=0.1\linewidth]{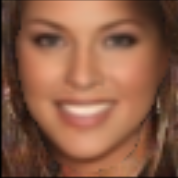}}
\subfloat{\includegraphics[width=0.1\linewidth, height=0.1\linewidth]{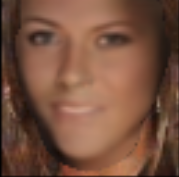}}&
\subfloat{\includegraphics[width=0.1\linewidth, height=0.1\linewidth]{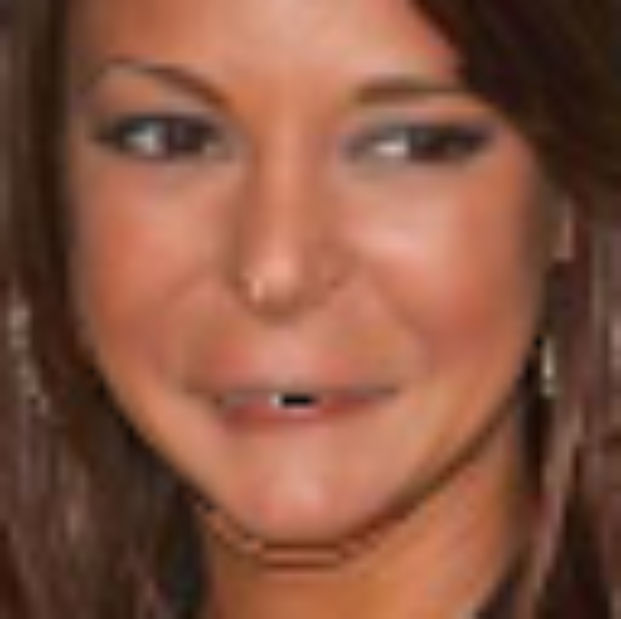}}
&
\subfloat{\includegraphics[width=0.1\linewidth, height=0.1\linewidth]{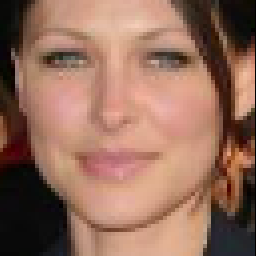}} 
\subfloat{\includegraphics[width=0.1\linewidth, height=0.1\linewidth]{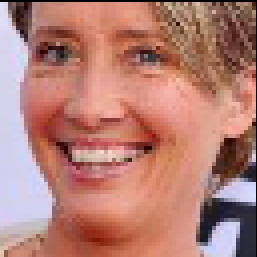}}
\subfloat{\includegraphics[width=0.1\linewidth, height=0.1\linewidth]{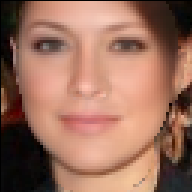}}
\subfloat{\includegraphics[width=0.1\linewidth, height=0.1\linewidth]{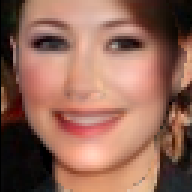}}&
\subfloat{\includegraphics[width=0.1\linewidth, height=0.1\linewidth]{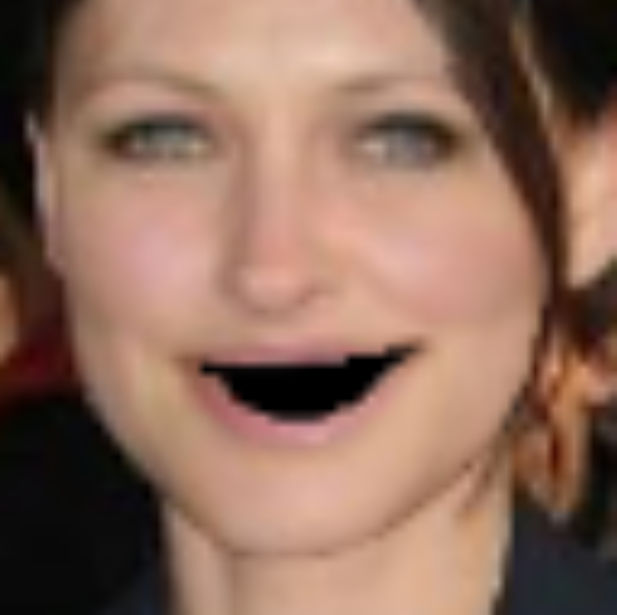}}
\\ \vspace{-9pt}
\subfloat{\includegraphics[width=0.1\linewidth]{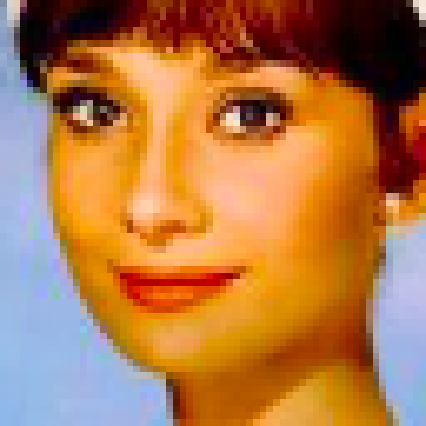}} 
\subfloat{\includegraphics[width=0.1\linewidth]{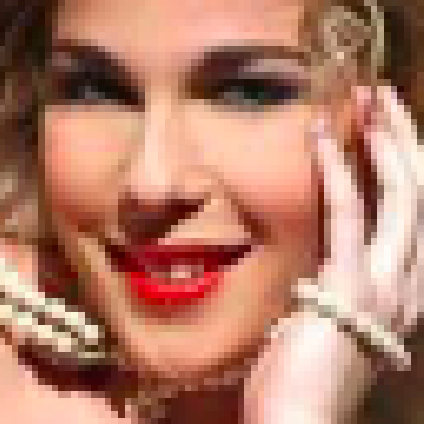}}
\subfloat{\includegraphics[width=0.1\linewidth]{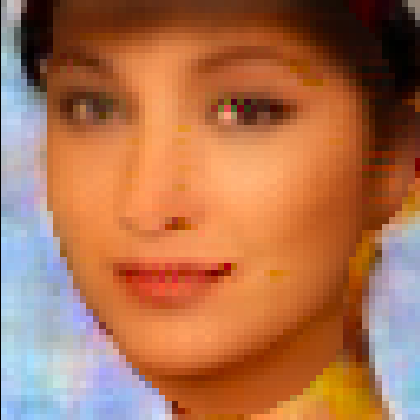}}
\subfloat{\includegraphics[width=0.1\linewidth]{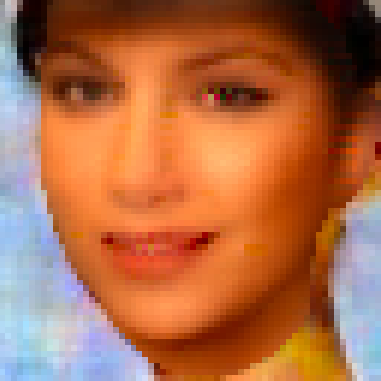}}&
\subfloat{\includegraphics[width=0.1\linewidth]{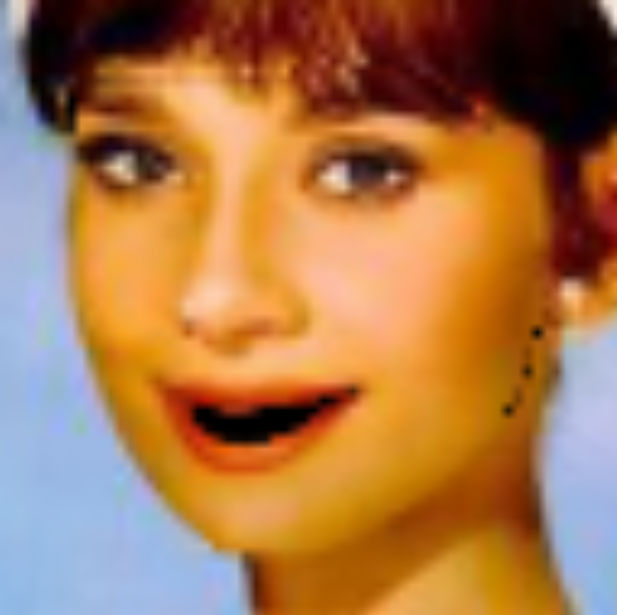}}
&
\subfloat{\includegraphics[width=0.1\linewidth]{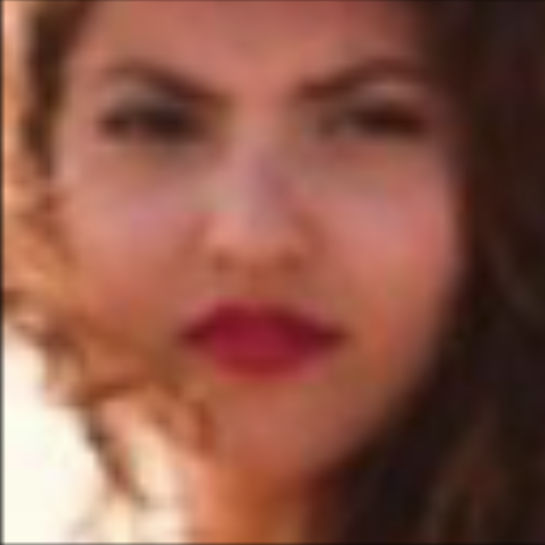}} 
\subfloat{\includegraphics[width=0.1\linewidth]{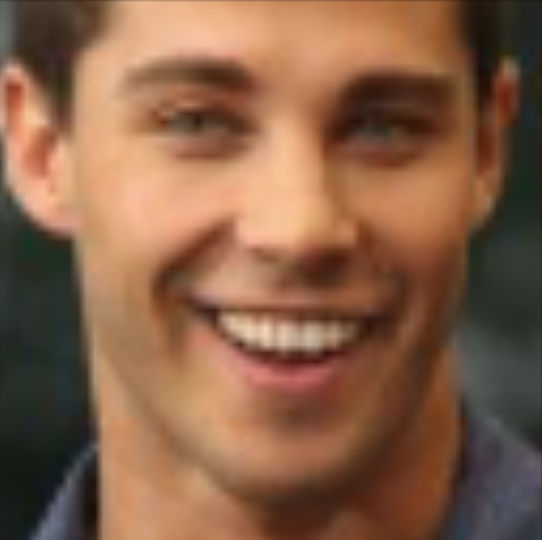}}
\subfloat{\includegraphics[width=0.1\linewidth]{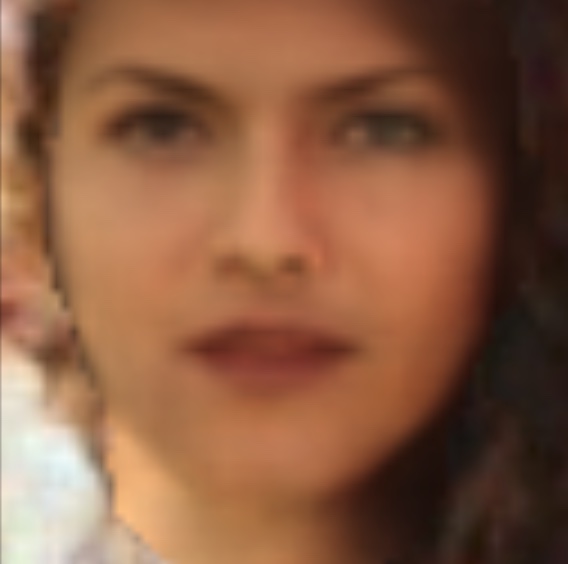}}
\subfloat{\includegraphics[width=0.1\linewidth]{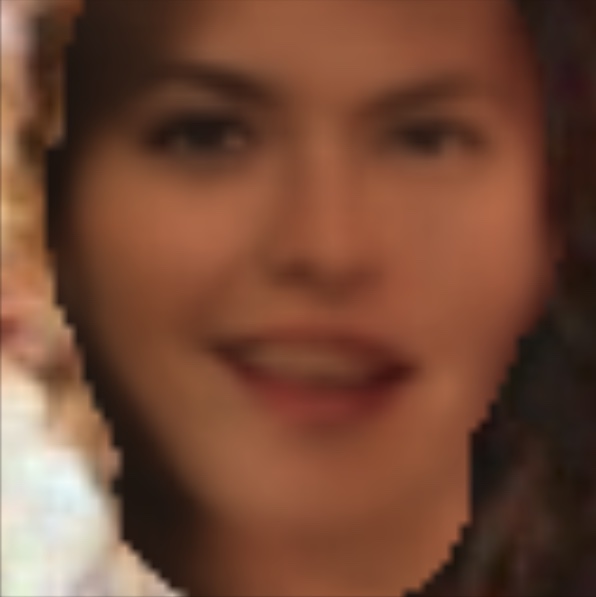}}&
\subfloat{\includegraphics[width=0.1\linewidth]{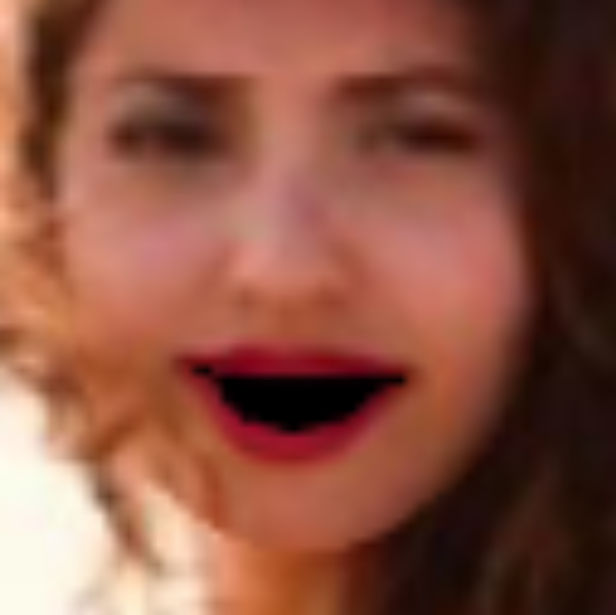}}
\\ \vspace{-9pt}
\subfloat{\includegraphics[width=0.1\linewidth, height=0.1\linewidth]{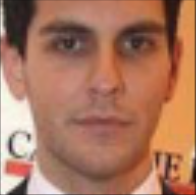}} 
\subfloat{\includegraphics[width=0.1\linewidth, height=0.1\linewidth]{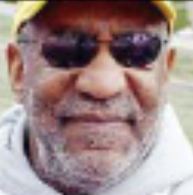}}
\subfloat{\includegraphics[width=0.1\linewidth, height=0.1\linewidth]{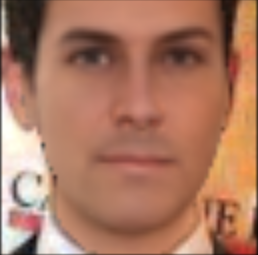}}
\subfloat{\includegraphics[width=0.1\linewidth, height=0.1\linewidth]{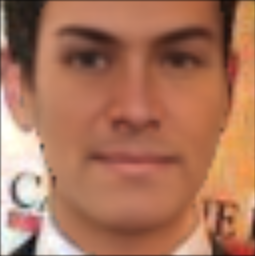}}&
\subfloat{\includegraphics[width=0.1\linewidth, height=0.1\linewidth]{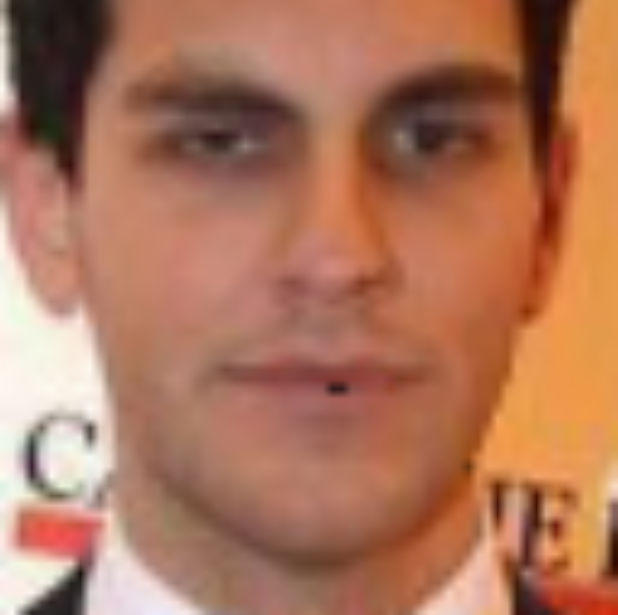}}
&
\subfloat{\includegraphics[width=0.1\linewidth]{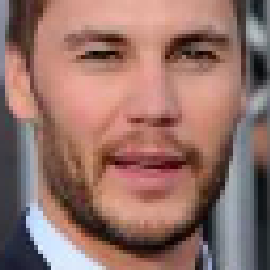}} 
\subfloat{\includegraphics[width=0.1\linewidth]{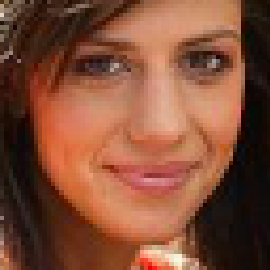}}
\subfloat{\includegraphics[width=0.1\linewidth]{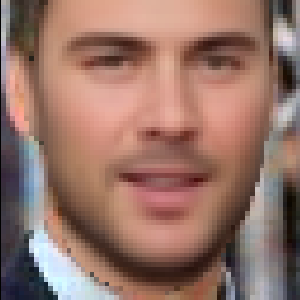}}
\subfloat{\includegraphics[width=0.1\linewidth]{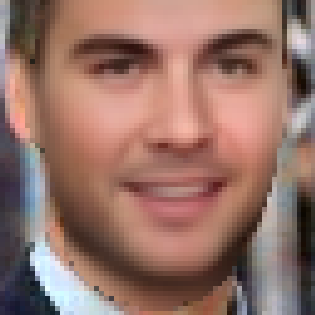}}&
\subfloat{\includegraphics[width=0.1\linewidth]{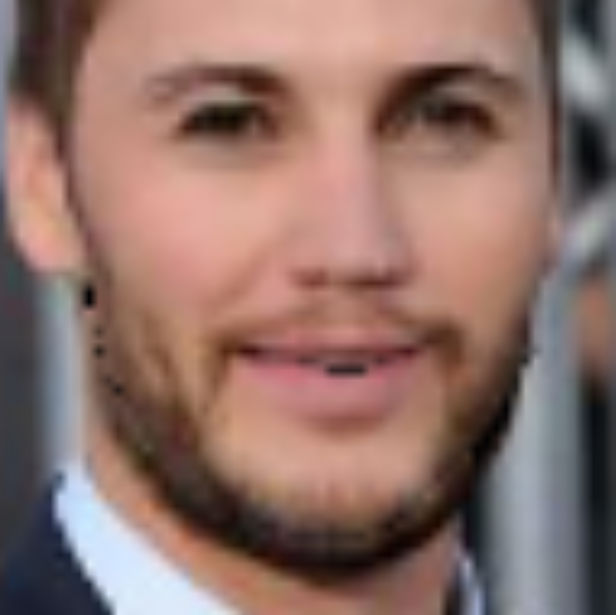}}
\\ \vspace{-9pt}
\subfloat{\includegraphics[width=0.1\linewidth]{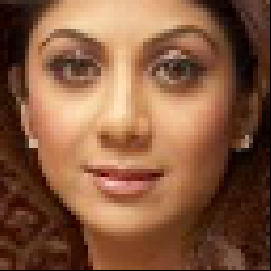}} 
\subfloat{\includegraphics[width=0.1\linewidth]{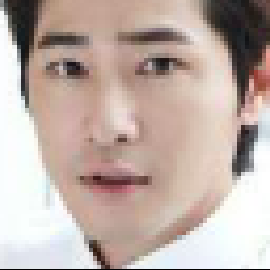}}
\subfloat{\includegraphics[width=0.1\linewidth]{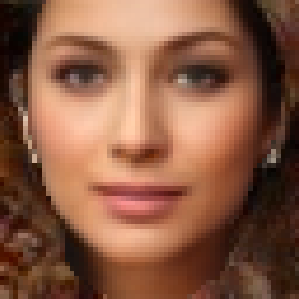}}
\subfloat{\includegraphics[width=0.1\linewidth]{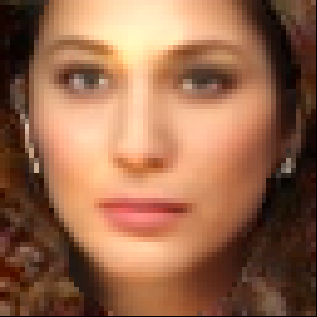}}&
\subfloat{\includegraphics[width=0.1\linewidth]{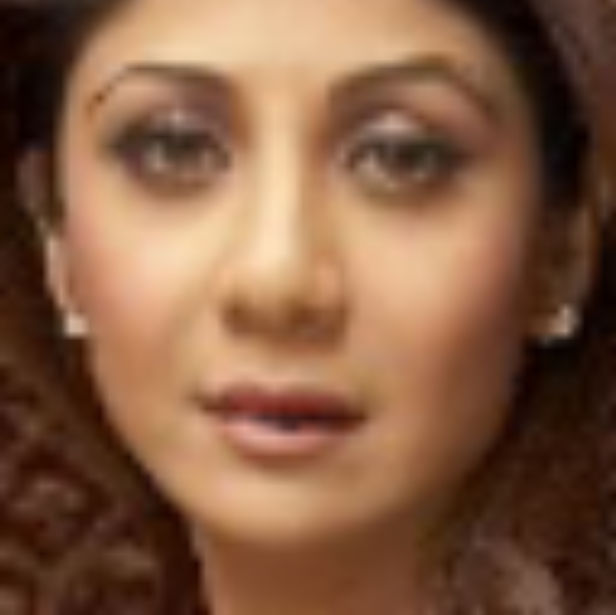}}
&
\subfloat{\includegraphics[width=0.1\linewidth, height=0.1\linewidth]{T5185_orig2}} 
\subfloat{\includegraphics[width=0.1\linewidth, height=0.1\linewidth]{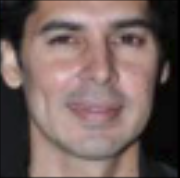}}
\subfloat{\includegraphics[width=0.1\linewidth, height=0.1\linewidth]{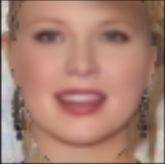}}
\subfloat{\includegraphics[width=0.1\linewidth, height=0.1\linewidth]{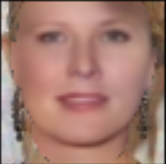}}&
\subfloat{\includegraphics[width=0.1\linewidth, height=0.1\linewidth]{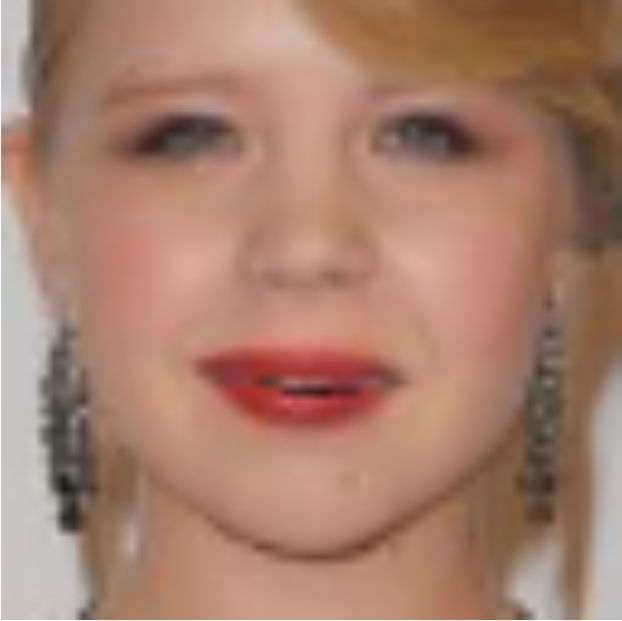}}
\\ \vspace{-9pt}
\subfloat{\includegraphics[width=0.1\linewidth]{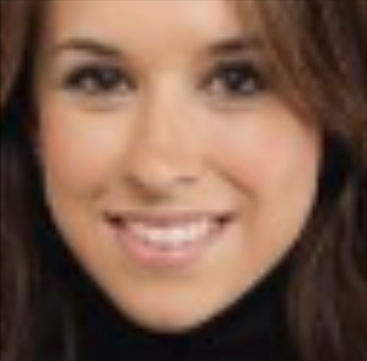}} 
\subfloat{\includegraphics[width=0.1\linewidth]{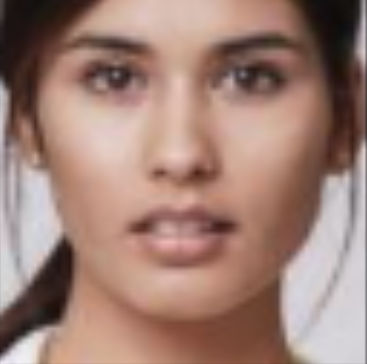}}
\subfloat{\includegraphics[width=0.1\linewidth]{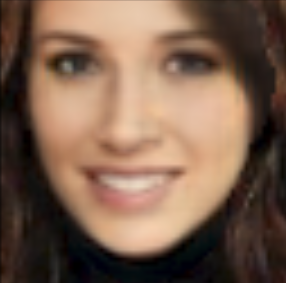}}
\subfloat{\includegraphics[width=0.1\linewidth]{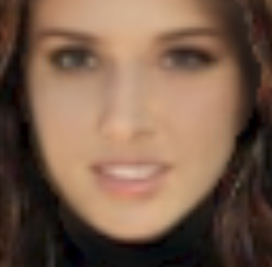}}&
\subfloat{\includegraphics[width=0.1\linewidth]{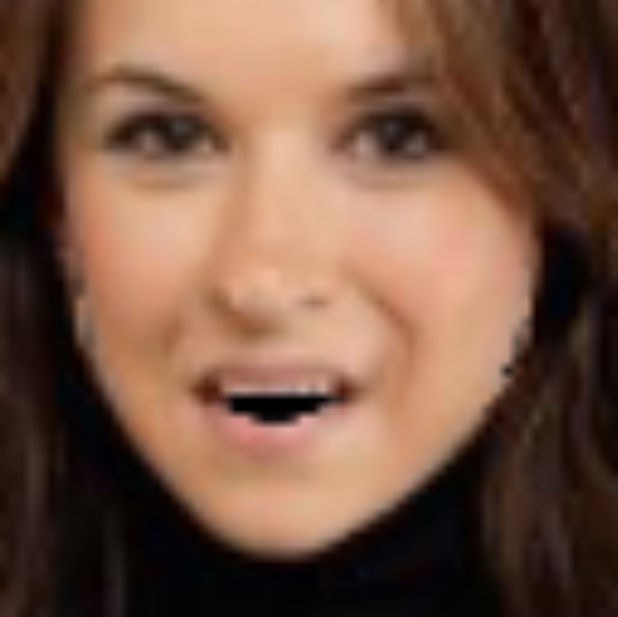}}
&
\subfloat{\includegraphics[width=0.1\linewidth, height=0.1\linewidth]{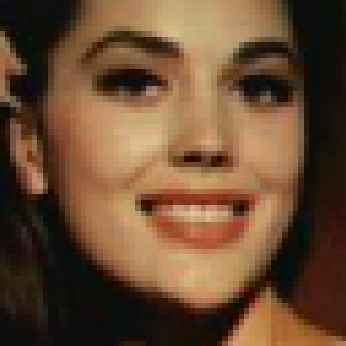}} 
\subfloat{\includegraphics[width=0.1\linewidth, height=0.1\linewidth]{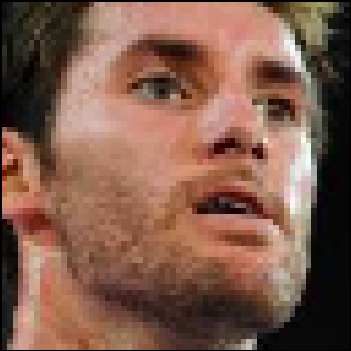}}
\subfloat{\includegraphics[width=0.1\linewidth, height=0.1\linewidth]{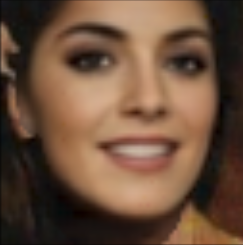}}
\subfloat{\includegraphics[width=0.1\linewidth, height=0.1\linewidth]{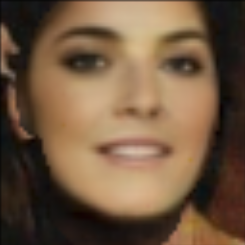}}&
\subfloat{\includegraphics[width=0.1\linewidth, height=0.1\linewidth]{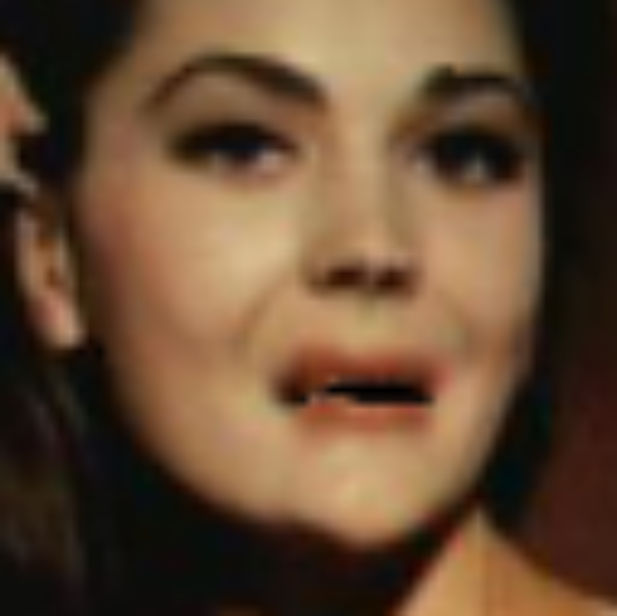}}
\\ \vspace{-9pt}
\subfloat{\includegraphics[width=0.1\linewidth, height=0.1\linewidth]{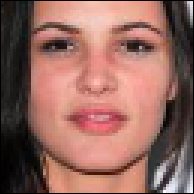}} 
\subfloat{\includegraphics[width=0.1\linewidth, height=0.1\linewidth]{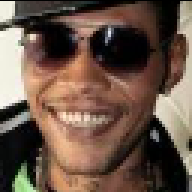}}
\subfloat{\includegraphics[width=0.1\linewidth, height=0.1\linewidth]{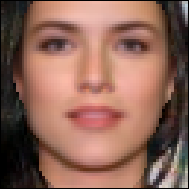}}
\subfloat{\includegraphics[width=0.1\linewidth, height=0.1\linewidth]{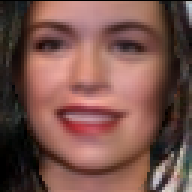}}&
\subfloat{\includegraphics[width=0.1\linewidth, height=0.1\linewidth]{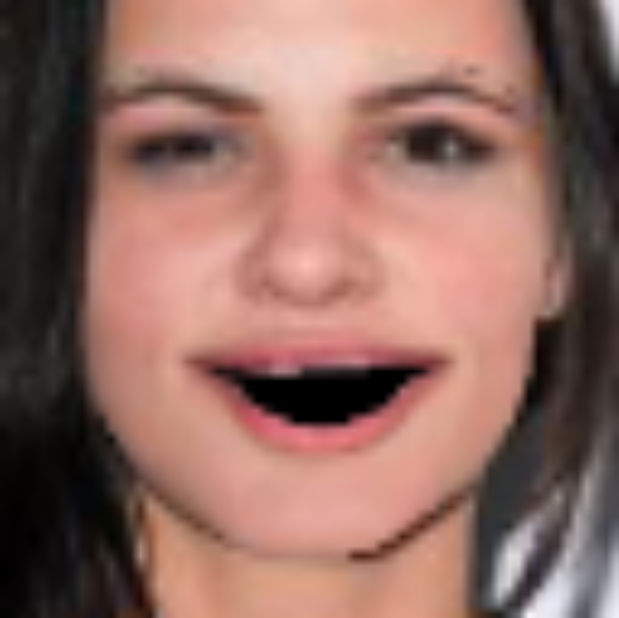}}
&
\subfloat{\includegraphics[width=0.1\linewidth, height=0.1\linewidth]{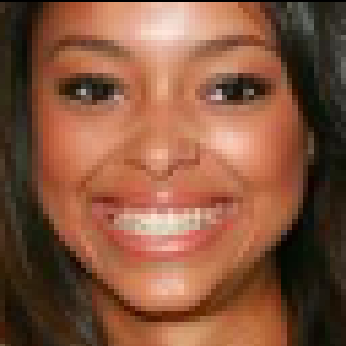}} 
\subfloat{\includegraphics[width=0.1\linewidth, height=0.1\linewidth]{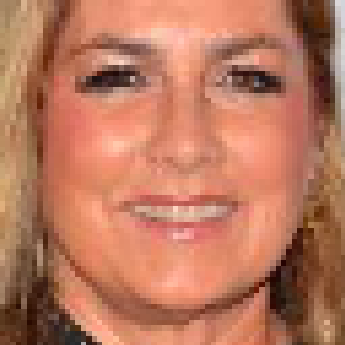}}
\subfloat{\includegraphics[width=0.1\linewidth, height=0.1\linewidth]{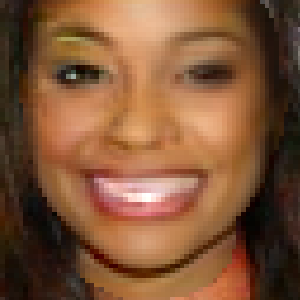}}
\subfloat{\includegraphics[width=0.1\linewidth, height=0.1\linewidth]{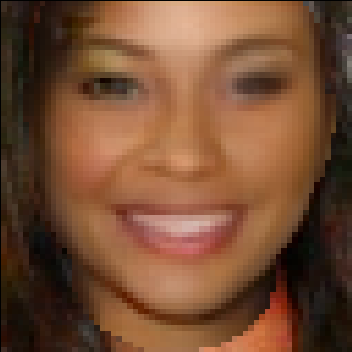}}&
\subfloat{\includegraphics[width=0.1\linewidth, height=0.1\linewidth]{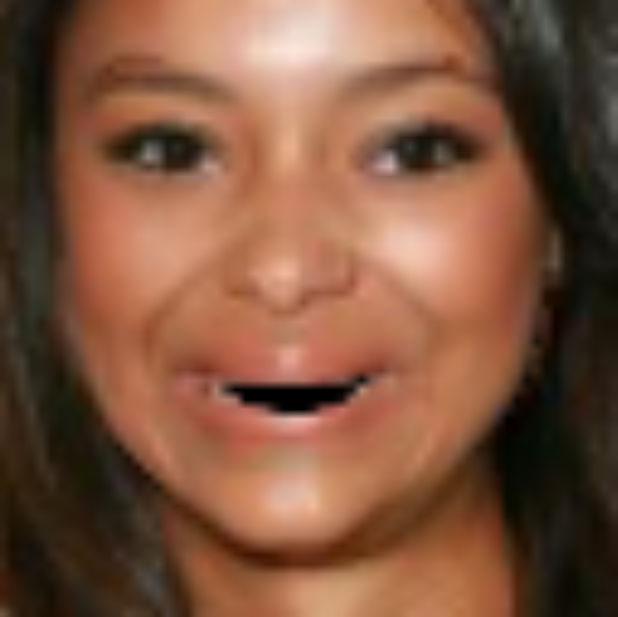}}
\\ \vspace{-9pt}
\subfloat{\includegraphics[width=0.1\linewidth]{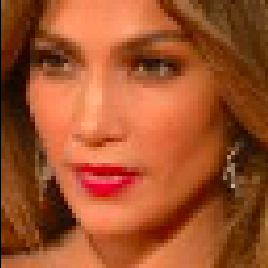}} 
\subfloat{\includegraphics[width=0.1\linewidth]{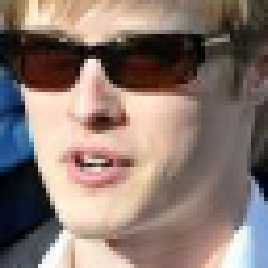}}
\subfloat{\includegraphics[width=0.1\linewidth]{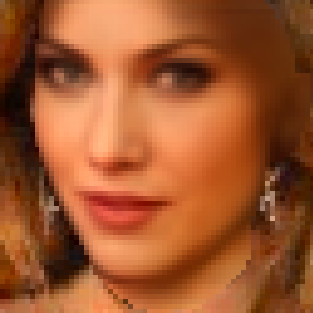}}
\subfloat{\includegraphics[width=0.1\linewidth]{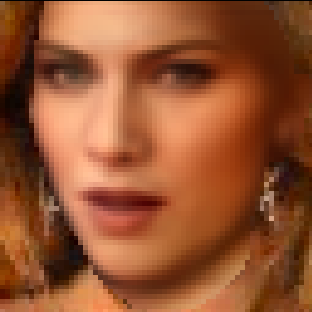}}&
\subfloat{\includegraphics[width=0.1\linewidth]{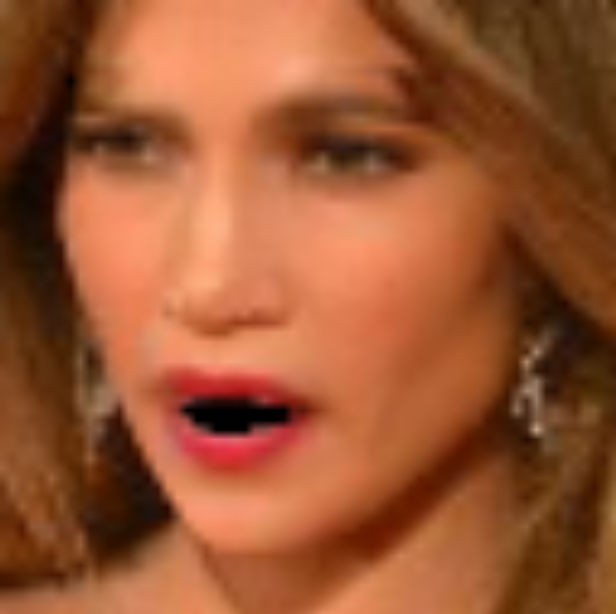}}
&
\subfloat{\includegraphics[width=0.1\linewidth]{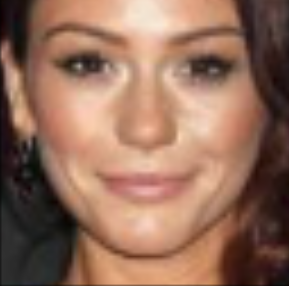}} 
\subfloat{\includegraphics[width=0.1\linewidth]{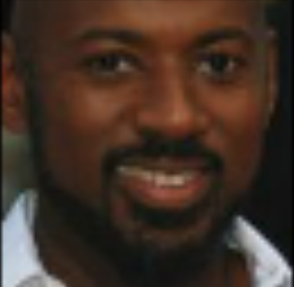}}
\subfloat{\includegraphics[width=0.1\linewidth]{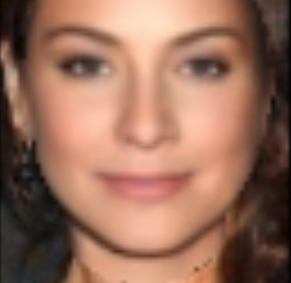}}
\subfloat{\includegraphics[width=0.1\linewidth]{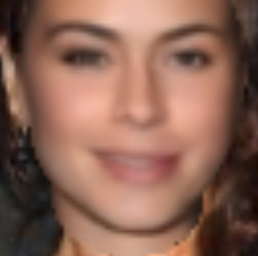}}&
\subfloat{\includegraphics[width=0.1\linewidth]{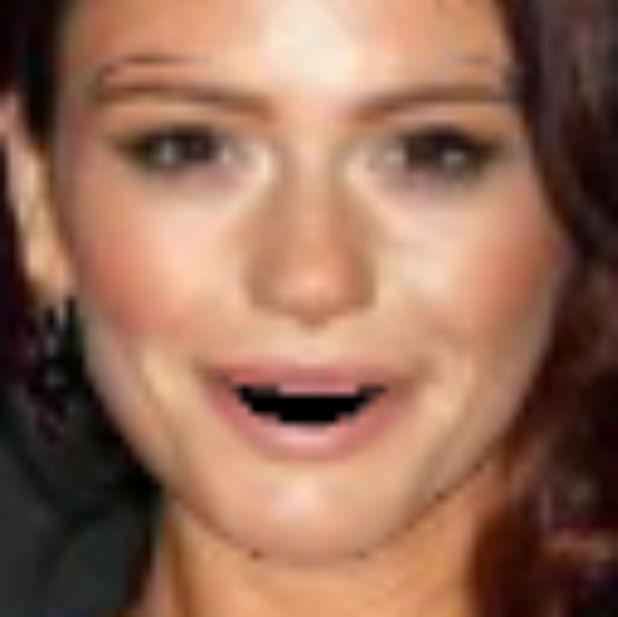}}
\end{tabular}
}
\caption{Expression Editing}
\label{exp2}
\vspace{-18pt}
\end{figure*}

\begin{figure*}[t!]
\captionsetup[subfigure]{labelformat=empty, justification=centering,position=top}
{\def\arraystretch{0.5}\tabcolsep=1pt
\begin{tabular}{ cccc } 
\vspace{-9pt}
\subfloat[Original Image]{\includegraphics[width=0.1\linewidth]{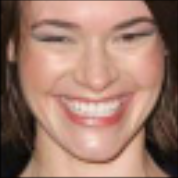}} 
\subfloat[Pose]{\includegraphics[width=0.1\linewidth]{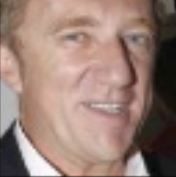}}
\subfloat[Recon]{\includegraphics[width=0.1\linewidth]{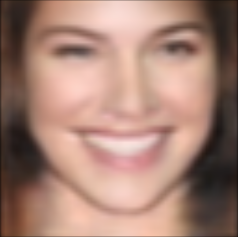}}
\subfloat[Our Pose Edit]{\includegraphics[width=0.1\linewidth]{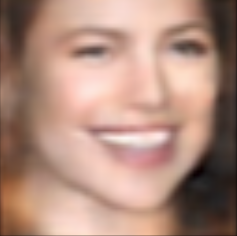}}
&
\subfloat[Baseline]{\includegraphics[trim={0.1cm 0.6cm 0cm 1cm},clip, width=0.1\linewidth]{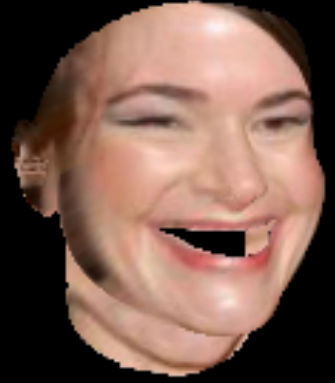}}
&
\subfloat[Original Image]{\includegraphics[width=0.1\linewidth]{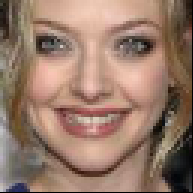}}
\subfloat[Pose]{\includegraphics[width=0.1\linewidth]{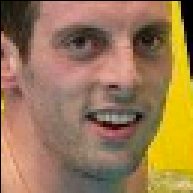}}
\subfloat[Recon]{\includegraphics[width=0.1\linewidth]{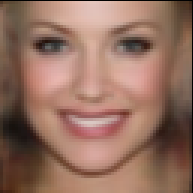}}
\subfloat[Our Pose Edit]{\includegraphics[width=0.1\linewidth]{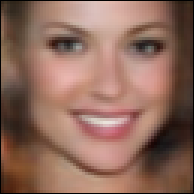}}
&
\subfloat[Baseline]{\includegraphics[width=0.1\linewidth]{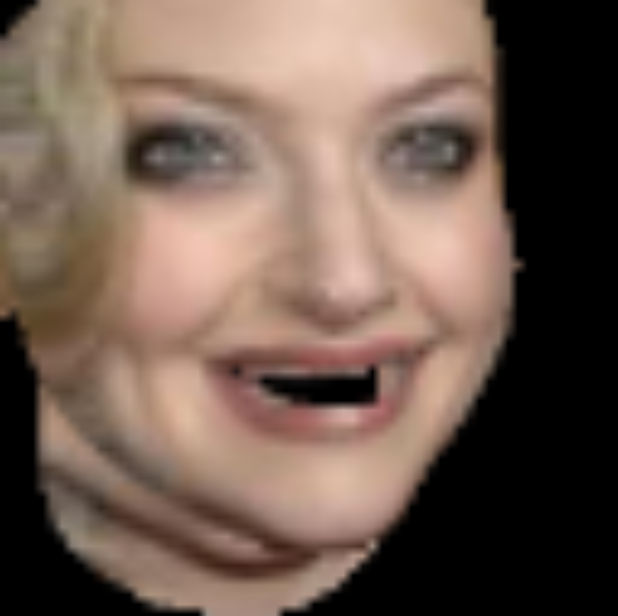}}
\\ \vspace{-9pt}
\subfloat{\includegraphics[width=0.1\linewidth]{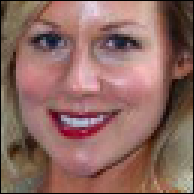}}
\subfloat{\includegraphics[width=0.1\linewidth]{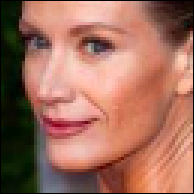}}
\subfloat{\includegraphics[width=0.1\linewidth]{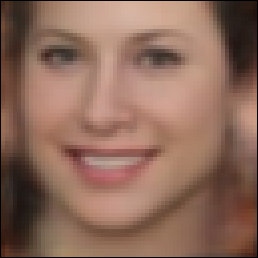}}
\subfloat{\includegraphics[width=0.1\linewidth]{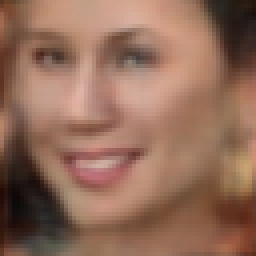}}
&
\subfloat{\includegraphics[trim={0.5cm 1cm 0cm 0cm},clip, width=0.1\linewidth]{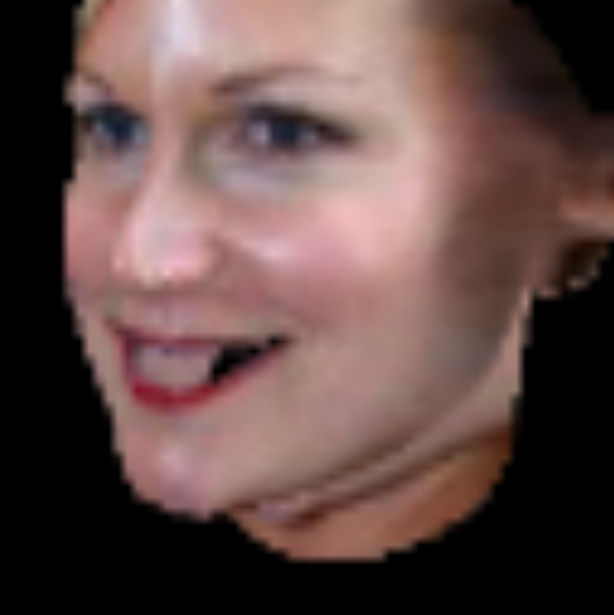}}
&
\subfloat{\includegraphics[width=0.1\linewidth]{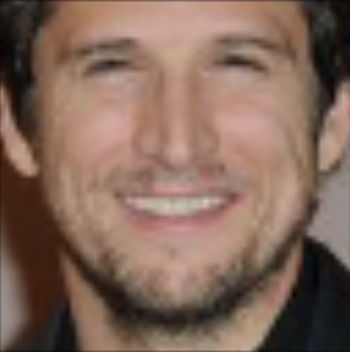}} 
\subfloat{\includegraphics[width=0.1\linewidth]{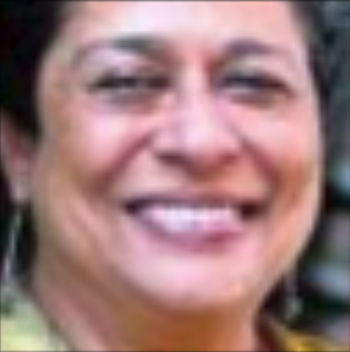}}
\subfloat{\includegraphics[width=0.1\linewidth]{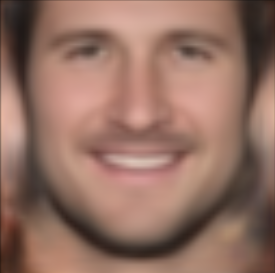}}
\subfloat{\includegraphics[width=0.1\linewidth]{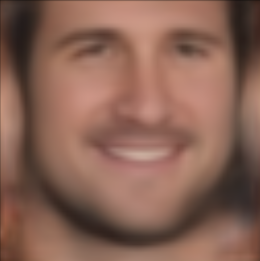}}
&
\subfloat{\includegraphics[trim={0.1cm 1cm 0cm 2cm},clip, width=0.1\linewidth]{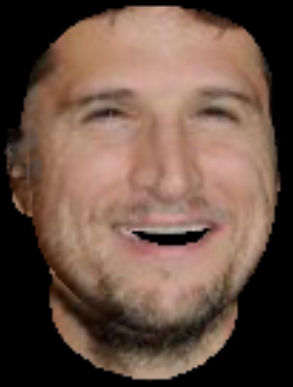}}
\\ \vspace{-9pt}
\subfloat{\includegraphics[width=0.1\linewidth]{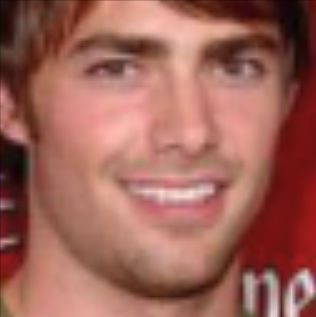}} 
\subfloat{\includegraphics[width=0.1\linewidth]{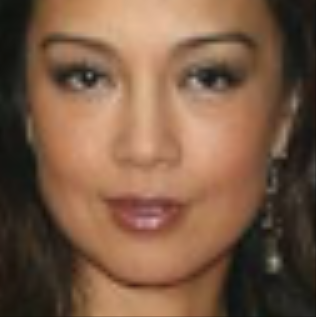}}
\subfloat{\includegraphics[width=0.1\linewidth]{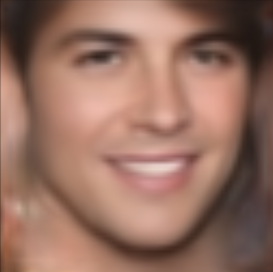}}
\subfloat{\includegraphics[width=0.1\linewidth]{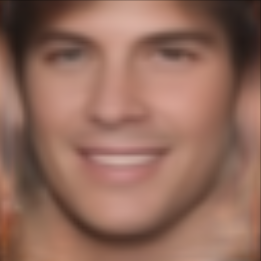}}
&
\subfloat{\includegraphics[trim={0.1cm 0.5cm 0cm 1.5cm},clip, width=0.1\linewidth]{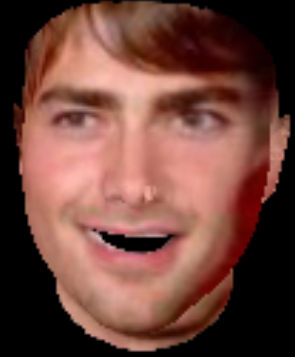}}
&
\subfloat{\includegraphics[width=0.1\linewidth]{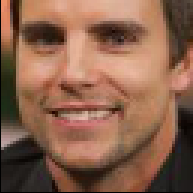}}
\subfloat{\includegraphics[width=0.1\linewidth]{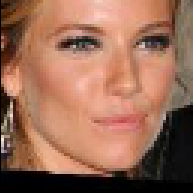}}
\subfloat{\includegraphics[width=0.1\linewidth]{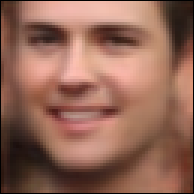}}
\subfloat{\includegraphics[width=0.1\linewidth]{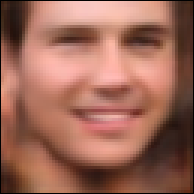}}
&
\subfloat{\includegraphics[width=0.1\linewidth]{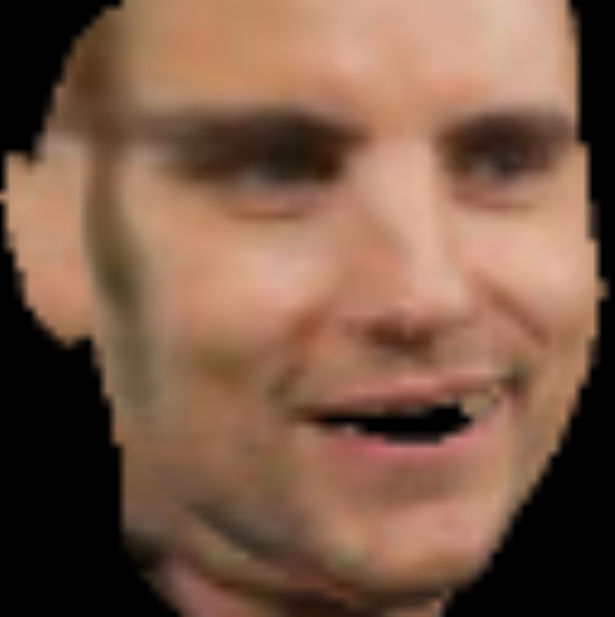}}
\end{tabular}
}
\caption{Pose Editing}
\label{pose2}
\end{figure*}
\section{Derivation Details}

The model is trained end-to-end by applying gradient descent to batches of images, where~\eqref{eq:e3d}, \eqref{eq:ef} and \eqref{eq:en} are written in the following general form:
\begin{equation} \tag{\ref{eq:tensor_err}} 
E = \| \bm{X} - \bm{B}_{(1)} (\bm{Z}^{(1)} \odot \bm{Z}^{(2)} \odot \dots \odot \bm{Z}^{(M)})\|^2_F,
\end{equation}
where $\bm{X} \in \mathbb{R}^{k \times n}$ is a data matrix, $\bm{B}_{(1)}$ is the mode-1 matricisation of a tensor $\mathcal{B}$ and $\bm{Z}^{(i)} \in \mathbb{R}^{k_{zi} \times n}$ are the latent variables matrices.

The partial derivative of \eqref{eq:tensor_err} with respect to the latent variable $\bm{Z}^{(i)}$ are computed as follows: 
Let $\hat{\bm{x}} = vec(\bm{X})$ be a vectorisation of $\bm{X}$ 
, then \eqref{eq:tensor_err} is equivalent with:

\begin{equation} 
\begin{aligned}
 &\| \bm{X} - \bm{B}_{(1)} (\bm{Z}^{(1)} \odot \bm{Z}^{(2)} \odot \dots \odot \bm{Z}^{(M)})\|^2_F \\
 = &\| vec(\bm{X} - \bm{B}_{(1)} (\bm{Z}^{(1)} \odot \bm{Z}^{(2)} \odot \dots \odot \bm{Z}^{(M)}))\|^2_2 \\
  = &\| \hat{\bm{x}}  - vec(\bm{B}_{(1)} (\bm{Z}^{(1)} \odot \bm{Z}^{(2)} \odot \dots \odot \bm{Z}^{(M)}))\|^2_2,
\end{aligned}
\end{equation}
as both the Frobenius norm and the $L_2$ norm are the sum of all elements squared.

\begin{equation} 
\begin{aligned}
&\| \hat{\bm{x}}  - vec(\bm{B}_{(1)} (\bm{Z}^{(1)} \odot \bm{Z}^{(2)} \odot \dots \odot \bm{Z}^{(M)}))\|^2_2 \\
= & \| \hat{\bm{x}} - (\bm{I} \otimes \bm{B}_{(1)}) vec(\bm{Z}^{(1)} \odot \bm{Z}^{(2)} \odot \dots \odot \bm{Z}^{(M)} )\|^2_2, 
\end{aligned}
\end{equation}
as the property $vec(\bm{BZ})  = (\bm{I} \otimes \bm{B}) vec(\bm{Z})$ holds~\cite{neudecker1969some}.

Using $vec(\bm{Z}^{(1)} \odot \bm{Z}^{(2)} ) = (\bm{I} \odot \bm{Z}^{(1)}) \otimes \bm{I} \cdot vec(\bm{Z}^{(2)}) $\cite{roemer2012advanced} and let $\hat{\bm{Z}^{(i-1)}} = \bm{Z}^{(1)} \odot \bm{Z}^{(2)} \odot \dots \odot \bm{Z}^{(i-1)} $ and $\hat{\bm{Z}^{(i)}} = \bm{Z}^{(i)} \odot \dots \odot \bm{Z}^{(M)} $ the following holds:
\begin{equation}
\begin{aligned}
& \| \hat{\bm{x}} - (\bm{I} \otimes \bm{B}_{(1)}) vec(\bm{Z}^{(1)} \odot \bm{Z}^{(2)} \odot \dots \odot \bm{Z}^{(M)} )\|^2_2 \\
=& \| \hat{\bm{x}} - (\bm{I} \otimes \bm{B}_{(1)})  (\bm{I} \odot \hat{\bm{Z}^{(i-1)}} ) \otimes \bm{I}  \cdot vec(\hat{\bm{Z}^{(i)}} )\|^2_2 \\
\end{aligned}
\end{equation}

Using $vec(\bm{Z}^{(1)} \odot \bm{Z}^{(2)} ) = \bm{I} \odot (\bm{Z}^{(2)} (\bm{I} \otimes \bm{\mathbb{1}}) ) \cdot vec(\bm{Z}^{(1)}) $\cite{roemer2012advanced} and let $\hat{\bm{Z}^{(i+1)}} = \bm{Z}^{(i+1)} \odot \dots \odot \bm{Z}^{(M)} $:
\begin{equation}
\begin{aligned}
& \| \hat{\bm{x}} - (\bm{I} \otimes \bm{B}_{(1)})  (\bm{I} \odot \hat{\bm{Z}^{(i-1)}} ) \otimes \bm{I}  \cdot vec(\hat{\bm{Z}^{(i)}} )\|^2_2 \\
= & \| \hat{\bm{x}} - (\bm{I} \otimes \bm{B}_{(1)})  (\bm{I} \odot \hat{\bm{Z}^{(i-1)}} ) \otimes \bm{I} \\
& \cdot \bm{I} \odot (\hat{\bm{Z}^{(i+1)}} (\bm{I} \otimes \bm{\mathbb{1}}) ) \cdot vec(\bm{Z}^{(i)})\|^2_2 \\
\end{aligned}
\end{equation}

Let $\hat{\bm{z}}^{(i)} = vec(\bm{Z}^{(i)})$ be a vectorisation of $\bm{Z}^{(i)}$, this becomes:
\begin{equation}
\begin{aligned}
& \| \hat{\bm{x}} - (\bm{I} \otimes \bm{B}_{(1)})  (\bm{I} \odot \hat{\bm{Z}^{(i-1)}} ) \otimes \bm{I} \\
& \cdot \bm{I} \odot (\hat{\bm{Z}^{(i+1)}} (\bm{I} \otimes \bm{\mathbb{1}}) ) \cdot \hat{\bm{z}^{(i)}}\|^2_2 \\
\end{aligned}
\tag{\ref{eq:tensor_vec}} 
\end{equation}

We then compute the partial derivative of~\eqref{eq:tensor_vec} with respect to $\hat{\bm{z}^{(i)}}$:
\begin{equation}
\frac{\partial \| \hat{\bm{x}} - \bm{A}  \hat{\bm{z}^{(i)}} \|^2_2}{\partial \hat{\bm{z}^{(i)}}} = 2 \bm{A}^T (\bm{A} \cdot \hat{\bm{z}^{(i)}} - \hat{\bm{x}} ),
\label{eq:tensor_vec2}
\end{equation}
where $\bm{A} = (\bm{I} \otimes \bm{B}_{(1)})  (\bm{I} \odot \hat{\bm{Z}^{(i-1)}} ) \otimes \bm{I} \cdot \bm{I} \odot (\hat{\bm{Z}^{(i+1)}} (\bm{I} \otimes \bm{\mathbb{1}}) ) $.

The partial derivative of~\eqref{eq:tensor_err} with respect to $\bm{Z}^{(i)}$ is obtained by matricising~\eqref{eq:tensor_vec2}.

\section{More expression and pose transfer images}
Figures~\ref{exp2} and ~\ref{pose2} show additional expression and pose editing results.

\begin{figure*}[t!]
\captionsetup[subfigure]{labelformat=empty, justification=centering,position=top}
{\def\arraystretch{0.5}\tabcolsep=1pt
\begin{tabular}{ c } 
\vspace{-9pt}

\subfloat{\includegraphics[width=1\linewidth]{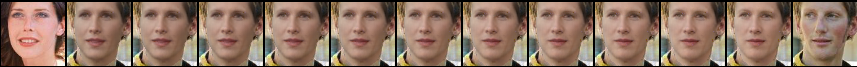}}\\ \vspace{-8pt}
\subfloat{\includegraphics[width=1\linewidth]{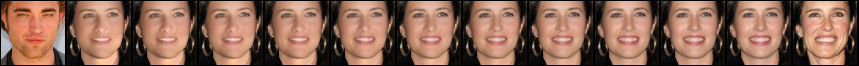}}\\ \vspace{-8pt}
\subfloat{\includegraphics[width=1\linewidth]{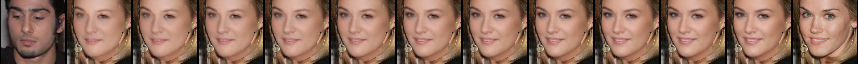}}\\ \vspace{-8pt}
\subfloat{\includegraphics[width=1\linewidth]{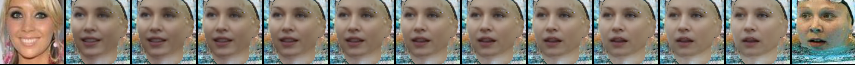}}\\ \vspace{-8pt}
\end{tabular}
}
\caption{Expression Interpolation}
\label{expinter}
\end{figure*}

\begin{figure*}[t!]
\captionsetup[subfigure]{labelformat=empty, justification=centering,position=top}
{\def\arraystretch{0.5}\tabcolsep=1pt
\begin{tabular}{ c } 
\vspace{-9pt}

\subfloat{\includegraphics[width=1\linewidth]{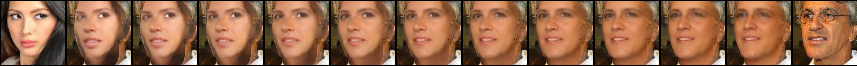}} \\ \vspace{-8pt}
\subfloat{\includegraphics[width=1\linewidth]{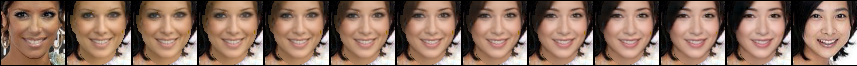}}\\ \vspace{-8pt}
\subfloat{\includegraphics[width=1\linewidth]{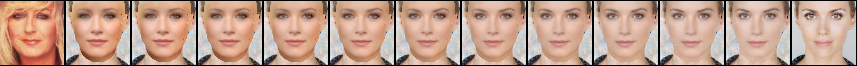}}\\ \vspace{-8pt}
\subfloat{\includegraphics[width=1\linewidth]{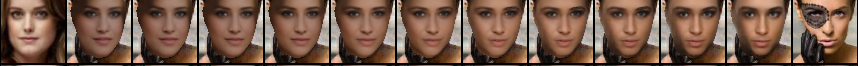}}\\ \vspace{-8pt}
\end{tabular}
}
\caption{Identity Interpolation}
\label{idinter}
\end{figure*}

\begin{figure*}[t!]
\captionsetup[subfigure]{labelformat=empty, justification=centering,position=top}
{\def\arraystretch{0.5}\tabcolsep=1pt
\begin{tabular}{ ccc } 
\vspace{-9pt}
&
\subfloat[Source]{\includegraphics[trim={0.88cm 1cm 0cm 0cm},clip, width=0.1\linewidth]{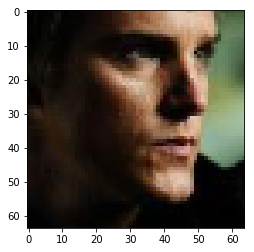}}
\subfloat[$\hat{\bm{s}}^{source}$]{\includegraphics[trim={0.88cm 1cm 0cm 0cm},clip,width=0.1\linewidth]{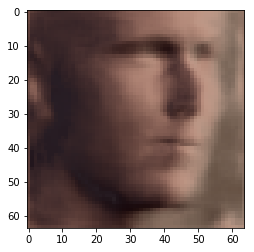}}
&
\subfloat[\cite{shu2017neural} $\hat{\bm{s}}^{source}$]{\includegraphics[trim={0cm 0cm 1.4cm 1.5cm},clip, width=0.1\linewidth]{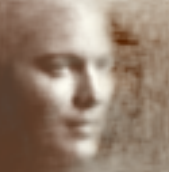}}
\\ \vspace{-9pt}
Ours
&
\subfloat[Target]{\includegraphics[trim={0.88cm 1cm 0cm 0cm},clip,width=0.1\linewidth]{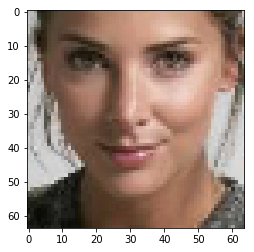}}
\subfloat[$\hat{\bm{s}}^{target}$]{\includegraphics[trim={0.88cm 1cm 0cm 0cm},clip,width=0.1\linewidth]{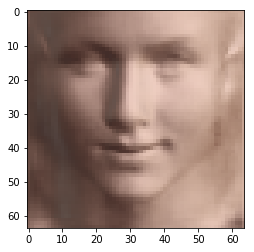}}
\subfloat[$\bm{s}^{transfer}$]{\includegraphics[trim={0.88cm 1cm 0cm 0cm},clip,width=0.1\linewidth]{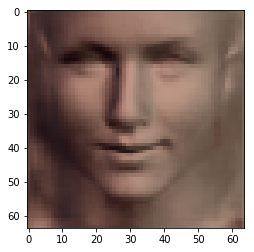}}
\subfloat[Result]{\includegraphics[trim={0.88cm 1cm 0cm 0cm},clip,width=0.1\linewidth]{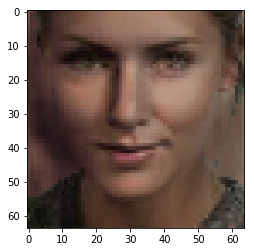}}
&
\subfloat[Target]{\includegraphics[trim={0.88cm 1cm 0cm 0cm},clip,width=0.1\linewidth]{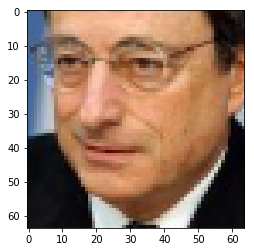}}
\subfloat[$\hat{\bm{s}}^{target}$]{\includegraphics[trim={0.88cm 1cm 0cm 0cm},clip,width=0.1\linewidth]{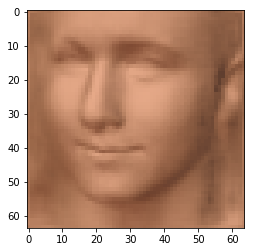}}
\subfloat[$\bm{s}^{transfer}$]{\includegraphics[trim={0.88cm 1cm 0cm 0cm},clip,width=0.1\linewidth]{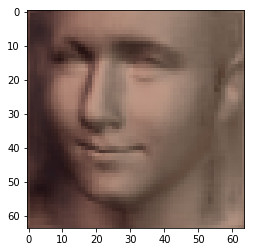}}
\subfloat[Result]{\includegraphics[trim={0.88cm 1cm 0cm 0cm},clip,width=0.1\linewidth]{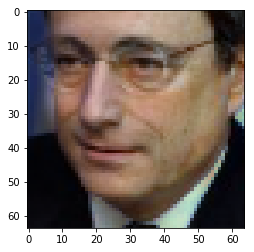}}
\\ \vspace{-9pt}
\cite{shu2017neural}
&
\subfloat{\includegraphics[trim={0.88cm 1cm 0cm 0cm},clip,height=0.1\linewidth]{test3}}
\subfloat{\includegraphics[trim={0.8cm 0cm 0.3cm 1.2cm},clip, height=0.1\linewidth,width=0.1\linewidth]{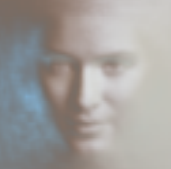}}
\subfloat{\includegraphics[trim={0.8cm 0cm 0.3cm 1.2cm},clip, height=0.1\linewidth,width=0.1\linewidth]{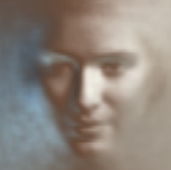}}
\subfloat{\includegraphics[trim={0.8cm 0cm 0.3cm 1.2cm},clip, height=0.1\linewidth,width=0.1\linewidth]{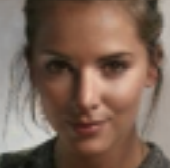}}
&
\subfloat{\includegraphics[trim={0.88cm 1cm 0cm 0cm},clip,height=0.1\linewidth]{test4}}
\subfloat{\includegraphics[trim={1.2cm 0cm 0cm 1cm},clip, height=0.1\linewidth,width=0.1\linewidth]{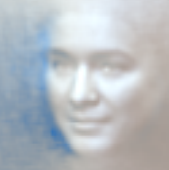}}
\subfloat{\includegraphics[trim={1.2cm 0cm 0cm 1cm},clip, height=0.1\linewidth,width=0.1\linewidth]{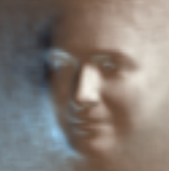}}
\subfloat{\includegraphics[trim={1.2cm 0cm 0cm 1cm},clip, height=0.1\linewidth,width=0.1\linewidth]{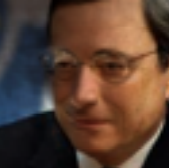}}
\\ \vspace{-9pt}
Ours
&
\subfloat{\includegraphics[trim={0.88cm 1cm 0cm 0cm},clip,width=0.1\linewidth]{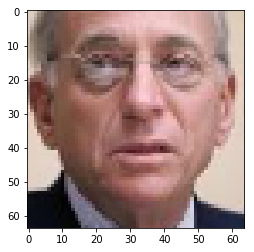}}
\subfloat{\includegraphics[trim={0.88cm 1cm 0cm 0cm},clip,width=0.1\linewidth]{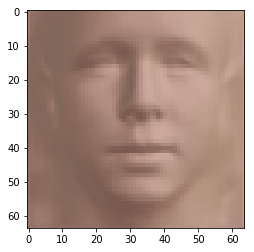}}
\subfloat{\includegraphics[trim={0.88cm 1cm 0cm 0cm},clip,width=0.1\linewidth]{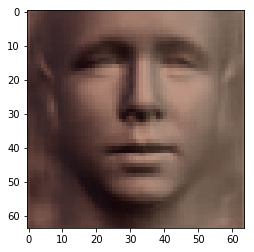}}
\subfloat{\includegraphics[trim={0.88cm 1cm 0cm 0cm},clip,width=0.1\linewidth]{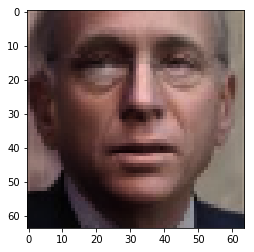}}
&
\subfloat{\includegraphics[trim={0.88cm 1cm 0cm 0cm},clip,width=0.1\linewidth]{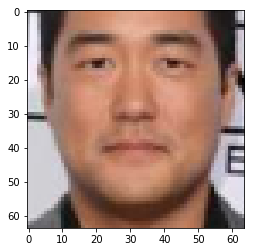}}
\subfloat{\includegraphics[trim={0.88cm 1cm 0cm 0cm},clip,width=0.1\linewidth]{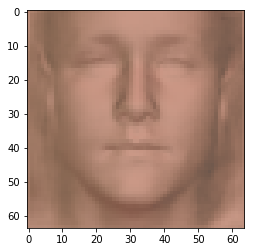}}
\subfloat{\includegraphics[trim={0.88cm 1cm 0cm 0cm},clip,width=0.1\linewidth]{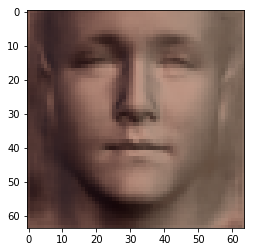}}
\subfloat{\includegraphics[trim={0.88cm 1cm 0cm 0cm},clip,width=0.1\linewidth]{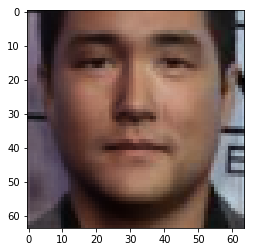}}
\\ \vspace{-9pt}
\cite{shu2017neural}
&
\subfloat{\includegraphics[trim={0.88cm 1cm 0cm 0cm},clip,height=0.1\linewidth]{test12}}
\subfloat{\includegraphics[trim={0.4cm 0cm 0.5cm 1cm},clip, height=0.1\linewidth,width=0.1\linewidth]{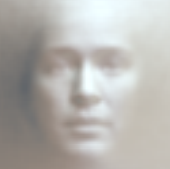}}
\subfloat{\includegraphics[trim={0.4cm 0cm 0.5cm 1cm},clip, height=0.1\linewidth,width=0.1\linewidth]{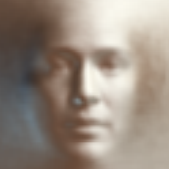}}
\subfloat{\includegraphics[trim={0.4cm 0cm 0.5cm 1cm},clip, height=0.1\linewidth,width=0.1\linewidth]{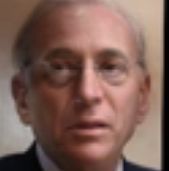}}
&
\subfloat{\includegraphics[trim={0.88cm 1cm 0cm 0cm},clip,height=0.1\linewidth]{test20}}
\subfloat{\includegraphics[trim={0.5cm 0cm 0.4cm 1cm},clip, height=0.1\linewidth,width=0.1\linewidth]{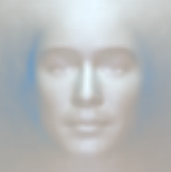}}
\subfloat{\includegraphics[trim={0.5cm 0cm 0.4cm 1cm},clip, height=0.1\linewidth,width=0.1\linewidth]{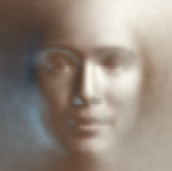}}
\subfloat{\includegraphics[trim={0.5cm 0cm 0.4cm 1cm},clip, height=0.1\linewidth,width=0.1\linewidth]{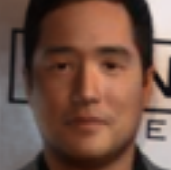}}
\\ \vspace{-9pt}
Ours
&
\subfloat{\includegraphics[trim={0.88cm 1cm 0cm 0cm},clip,width=0.1\linewidth]{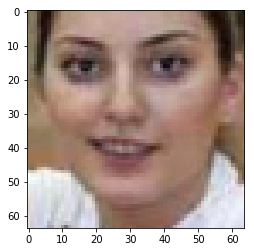}}
\subfloat{\includegraphics[trim={0.88cm 1cm 0cm 0cm},clip,width=0.1\linewidth]{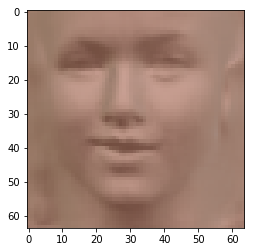}}
\subfloat{\includegraphics[trim={0.88cm 1cm 0cm 0cm},clip,width=0.1\linewidth]{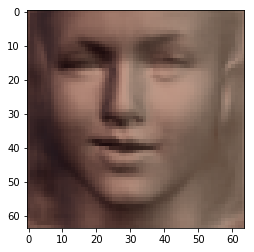}}
\subfloat{\includegraphics[trim={0.88cm 1cm 0cm 0cm},clip,width=0.1\linewidth]{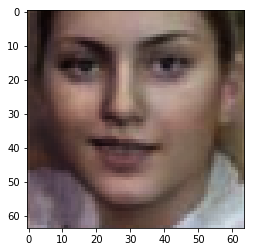}}
&
\subfloat{\includegraphics[trim={0.88cm 1cm 0cm 0cm},clip,width=0.1\linewidth]{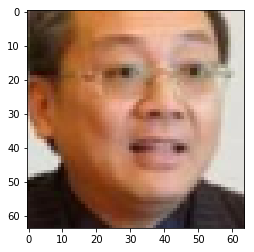}}
\subfloat{\includegraphics[trim={0.88cm 1cm 0cm 0cm},clip,width=0.1\linewidth]{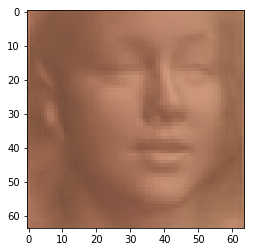}}
\subfloat{\includegraphics[trim={0.88cm 1cm 0cm 0cm},clip,width=0.1\linewidth]{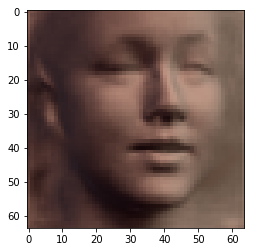}}
\subfloat{\includegraphics[trim={0.88cm 1cm 0cm 0cm},clip,width=0.1\linewidth]{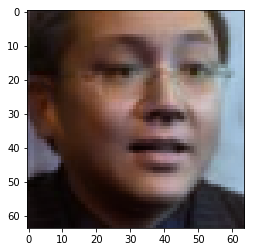}}
\\ \vspace{-9pt}
\cite{shu2017neural}
&
\subfloat{\includegraphics[trim={0.88cm 1cm 0cm 0cm},clip,height=0.1\linewidth]{test24}}
\subfloat{\includegraphics[trim={0.7cm 0cm 0.2cm 1cm},clip, height=0.1\linewidth,width=0.1\linewidth]{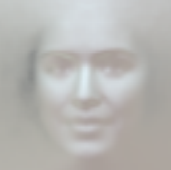}}
\subfloat{\includegraphics[trim={0.7cm 0cm 0.2cm 1cm},clip, height=0.1\linewidth,width=0.1\linewidth]{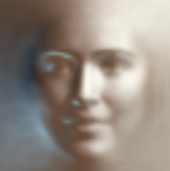}}
\subfloat{\includegraphics[trim={0.7cm 0cm 0.2cm 1cm},clip, height=0.1\linewidth,width=0.1\linewidth]{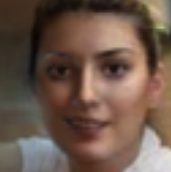}}
&
\subfloat{\includegraphics[trim={0.88cm 1cm 0cm 0cm},clip,height=0.1\linewidth]{test26}}
\subfloat{\includegraphics[trim={0cm 0cm 1cm 1cm},clip, height=0.1\linewidth,width=0.1\linewidth]{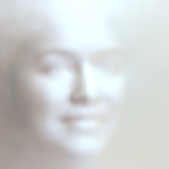}}
\subfloat{\includegraphics[trim={0cm 0cm 1cm 1cm},clip, height=0.1\linewidth,width=0.1\linewidth]{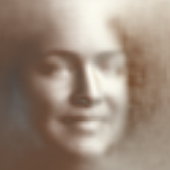}}
\subfloat{\includegraphics[trim={0cm 0cm 1cm 1cm},clip, height=0.1\linewidth,width=0.1\linewidth]{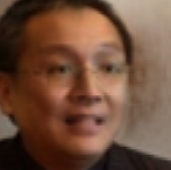}}
\\ \vspace{-9pt}
Ours
&
\subfloat{\includegraphics[trim={0.88cm 1cm 0cm 0cm},clip,width=0.1\linewidth]{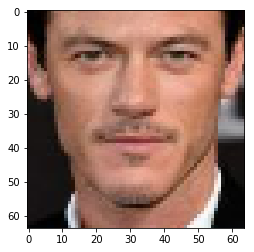}}
\subfloat{\includegraphics[trim={0.88cm 1cm 0cm 0cm},clip,width=0.1\linewidth]{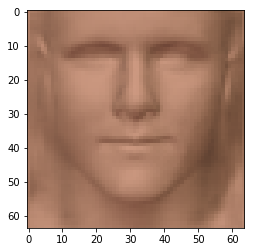}}
\subfloat{\includegraphics[trim={0.88cm 1cm 0cm 0cm},clip,width=0.1\linewidth]{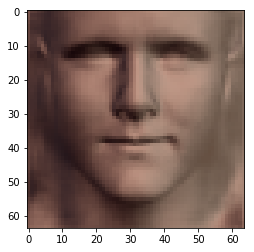}}
\subfloat{\includegraphics[trim={0.88cm 1cm 0cm 0cm},clip,width=0.1\linewidth]{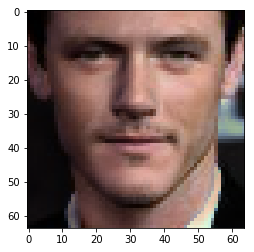}}
&
\subfloat{\includegraphics[trim={0.88cm 1cm 0cm 0cm},clip,width=0.1\linewidth]{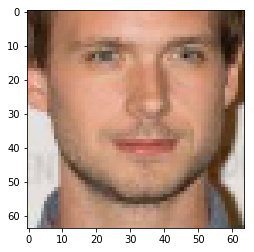}}
\subfloat{\includegraphics[trim={0.88cm 1cm 0cm 0cm},clip,width=0.1\linewidth]{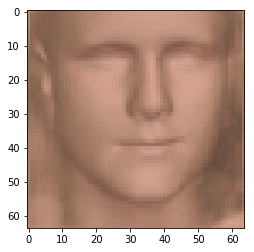}}
\subfloat{\includegraphics[trim={0.88cm 1cm 0cm 0cm},clip,width=0.1\linewidth]{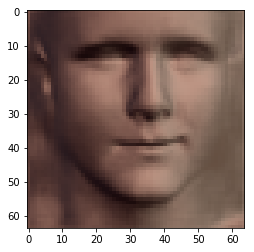}}
\subfloat{\includegraphics[trim={0.88cm 1cm 0cm 0cm},clip,width=0.1\linewidth]{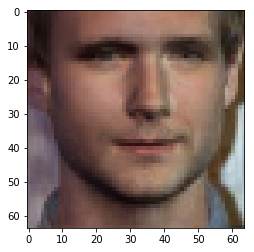}}
\\ \vspace{-9pt}
\cite{shu2017neural}
&
\subfloat{\includegraphics[trim={0.88cm 1cm 0cm 0cm},clip,height=0.1\linewidth]{test28}}
\subfloat{\includegraphics[trim={0.5cm 0cm 0.5cm 1cm},clip, height=0.1\linewidth,width=0.1\linewidth]{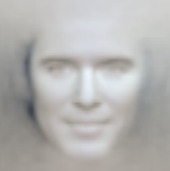}}
\subfloat{\includegraphics[trim={0.5cm 0cm 0.5cm 1cm},clip, height=0.1\linewidth,width=0.1\linewidth]{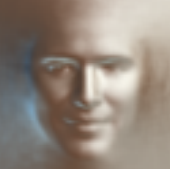}}
\subfloat{\includegraphics[trim={0.5cm 0cm 0.5cm 1cm},clip, height=0.1\linewidth,width=0.1\linewidth]{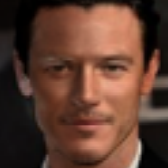}}
&

\subfloat{\includegraphics[trim={0.88cm 1cm 0cm 0cm},clip,height=0.1\linewidth]{test35}}
\subfloat{\includegraphics[trim={0.2cm 0cm 0.5cm 1cm},clip, height=0.1\linewidth,width=0.1\linewidth]{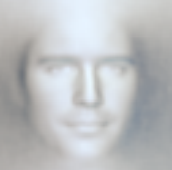}}
\subfloat{\includegraphics[trim={0.2cm 0cm 0.5cm 1cm},clip, height=0.1\linewidth,width=0.1\linewidth]{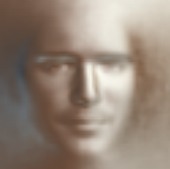}}
\subfloat{\includegraphics[trim={0.2cm 0cm 0.5cm 1cm},clip, height=0.1\linewidth,width=0.1\linewidth]{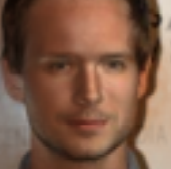}}
\\
\end{tabular}
}
\caption{We relight target faces using illumination from the source image. We compare against results presented in~\cite{shu2017neural}.}
    \label{relight2}
\end{figure*}

\begin{figure*}[t!]
\centering
\captionsetup[subfigure]{labelformat=empty, justification=centering,position=top}
{\def\arraystretch{0.5}\tabcolsep=1pt
\begin{tabular}{ cc } 
\vspace{-9pt}
\subfloat[Original Image]{\includegraphics[width=0.1\linewidth]{FF2321_orig}} 
\subfloat[Expression]{\includegraphics[width=0.1\linewidth]{FF2321_exp}}
\subfloat[Our Recon]{\includegraphics[width=0.1\linewidth]{FF2321_rec}}
\subfloat[Our Exp Edit]{\includegraphics[width=0.1\linewidth]{FF2321_modexp}}&
\subfloat[B \& W]{\includegraphics[width=0.1\linewidth]{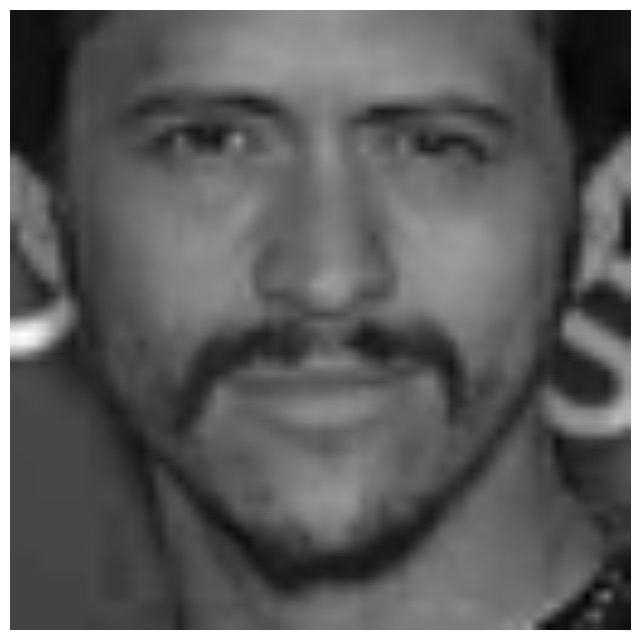}}
\subfloat[\cite{wang2017learning}]{\includegraphics[width=0.1\linewidth]{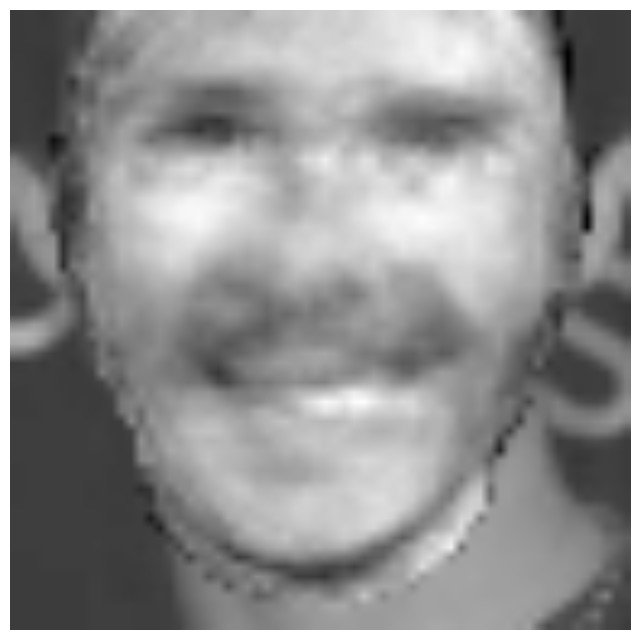}}
\\ \vspace{-9pt}
\subfloat{\includegraphics[width=0.1\linewidth]{A370_orig}} 
\subfloat{\includegraphics[width=0.1\linewidth]{A370_exp}}
\subfloat{\includegraphics[width=0.1\linewidth]{A370_rec}}
\subfloat{\includegraphics[width=0.1\linewidth]{A370_modexp}}&
\subfloat{\includegraphics[width=0.1\linewidth]{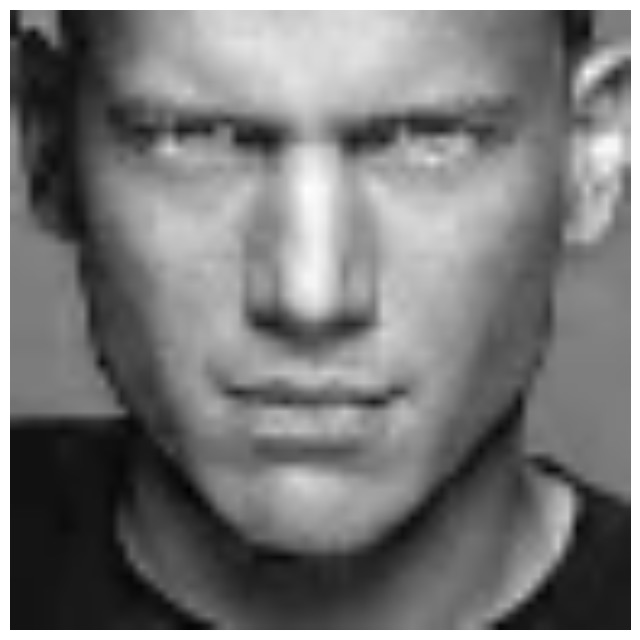}}
\subfloat{\includegraphics[width=0.1\linewidth]{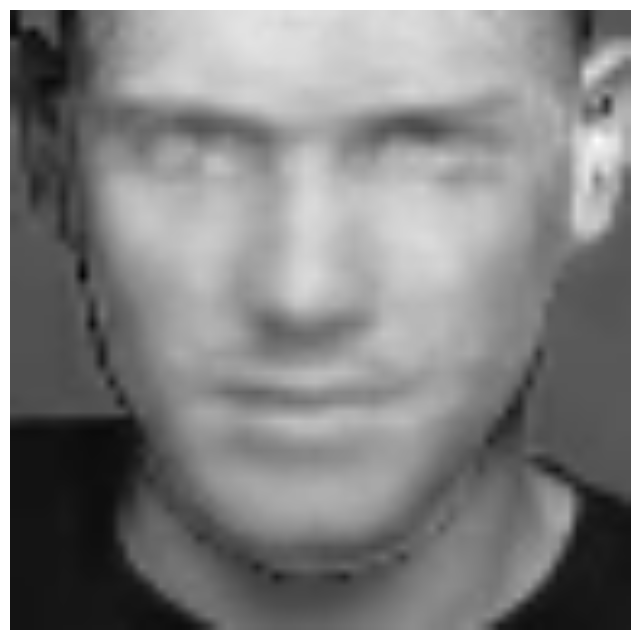}}
\\ \vspace{-9pt}
\subfloat{\includegraphics[trim={0 0.1cm 0.1cm 0},clip, width=0.1\linewidth, height=0.1\linewidth]{T7153_orig}} 
\subfloat{\includegraphics[trim={0.1cm 0 0.1cm 0.1cm},clip, width=0.1\linewidth, height=0.1\linewidth]{T7153_exp}}
\subfloat{\includegraphics[trim={0.1cm 0.1cm 0.1cm 0},clip, width=0.1\linewidth, height=0.1\linewidth]{T7153_rec2}}
\subfloat{\includegraphics[trim={0.1cm 0.1cm 0.1cm 0.1cm},clip, width=0.1\linewidth, height=0.1\linewidth]{T7153_modexp2}}&
\subfloat{\includegraphics[width=0.1\linewidth, height=0.1\linewidth]{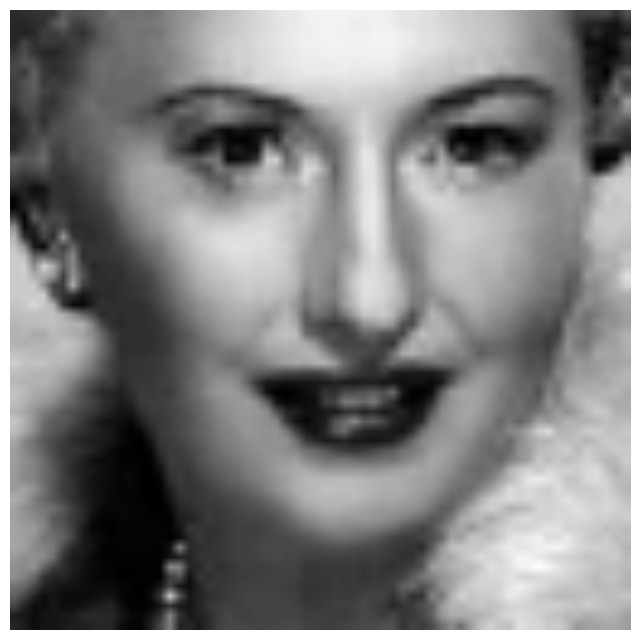}}
\subfloat{\includegraphics[width=0.1\linewidth, height=0.1\linewidth]{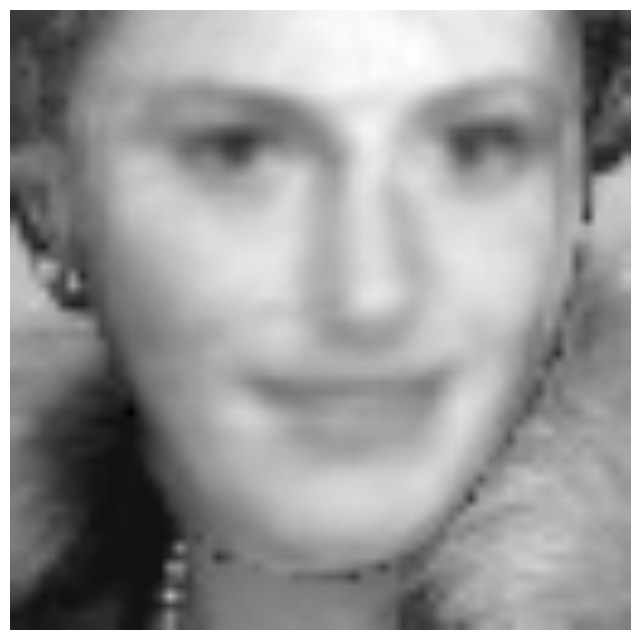}}
\\ \vspace{-9pt}
\subfloat{\includegraphics[width=0.1\linewidth, height=0.1\linewidth]{T7327_orig}} 
\subfloat{\includegraphics[width=0.1\linewidth, height=0.1\linewidth]{T7327_exp}}
\subfloat{\includegraphics[width=0.1\linewidth, height=0.1\linewidth]{T7327_rec}}
\subfloat{\includegraphics[width=0.1\linewidth, height=0.1\linewidth]{T7327_modexp}}&
\subfloat{\includegraphics[width=0.1\linewidth, height=0.1\linewidth]{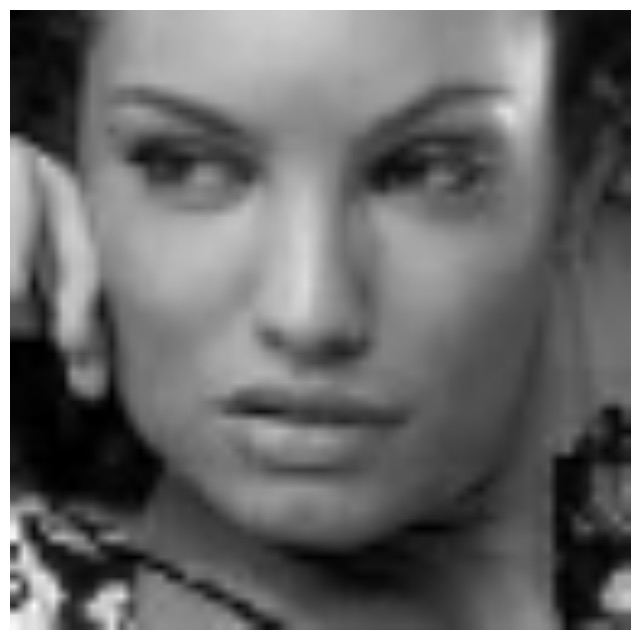}}
\subfloat{\includegraphics[width=0.1\linewidth, height=0.1\linewidth]{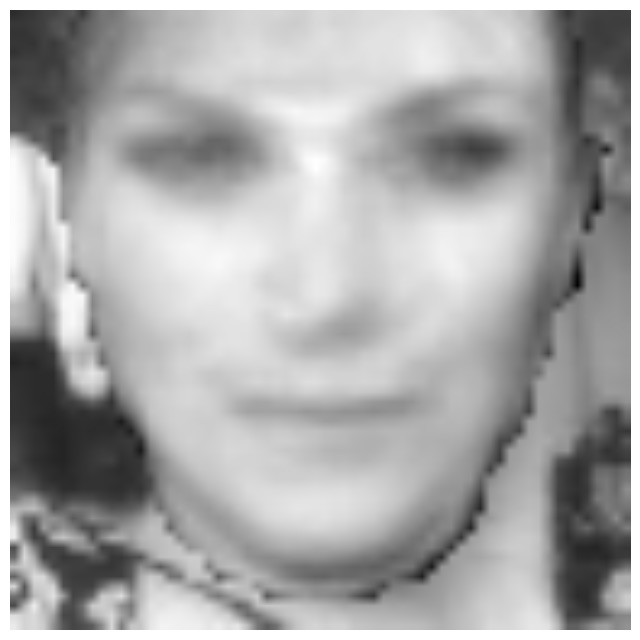}}
\\ \vspace{-9pt}
\subfloat{\includegraphics[width=0.1\linewidth, height=0.1\linewidth]{FV159_orig}}  
\subfloat{\includegraphics[width=0.1\linewidth, height=0.1\linewidth]{FV159_exp}}
\subfloat{\includegraphics[width=0.1\linewidth, height=0.1\linewidth]{FV159_rec}}
\subfloat{\includegraphics[width=0.1\linewidth, height=0.1\linewidth]{FV159_modexp}}&
\subfloat{\includegraphics[width=0.1\linewidth, height=0.1\linewidth]{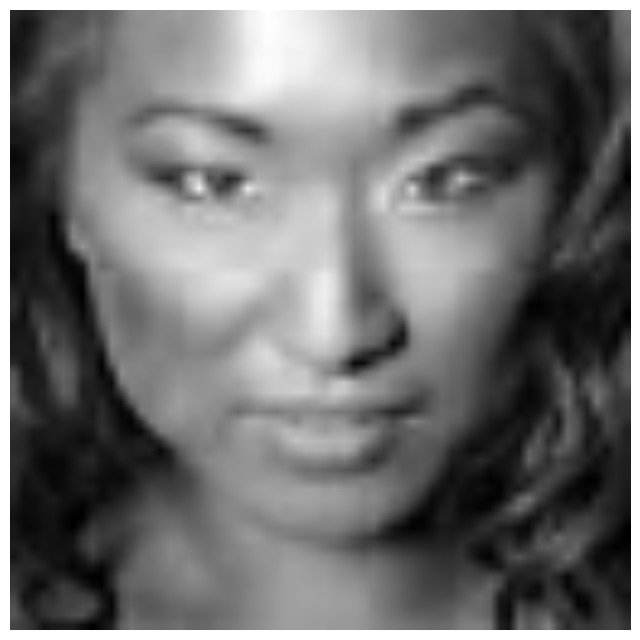}}
\subfloat{\includegraphics[width=0.1\linewidth, height=0.1\linewidth]{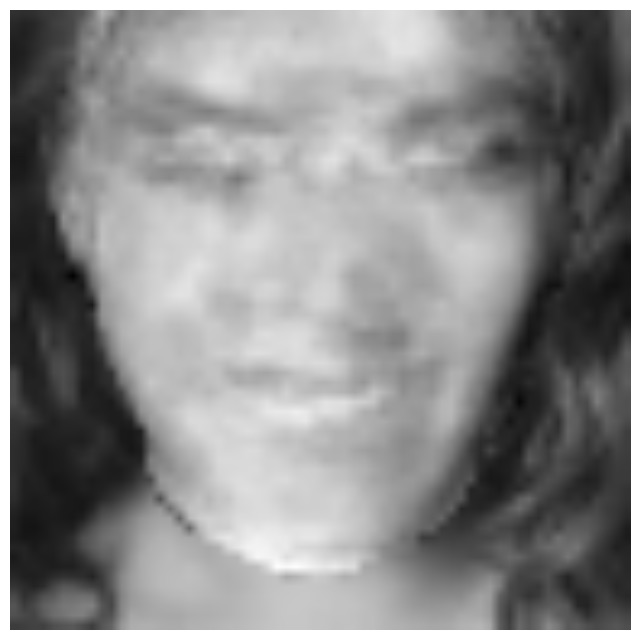}}
\end{tabular}
}
\caption{We compare our expression editing results with \cite{wang2017learning}. As \cite{wang2017learning} requires frontalisation of the face, applying it on our aligned input data does not achieve good results.}
    \label{expcomp2}
\end{figure*}

\newpage
\section{Interpolation Results}
 We interpolate $\bm{z}_{exp}^i$ / $\bm{z}_{id}^i$ of the input image $\bm{x}^i$ on the right-hand side to the $\bm{z}_{exp}^t$ / $\bm{z}_{id}^t$ of the target image $\bm{x}^t$ on the left-hand side. The interpolation is linear and at 0.1 interval. For the interpolation we do not modify the background so the background remains that of image $\bm{x}^i$. 

For expression interpolation, we expect the identity and pose to stay the same as the input image $\bm{x}^i$ and only the expression to change gradually from  the expression of the input image to the expression of the target image $\bm{x}^t$. Figure~\ref{expinter} shows the expression interpolation. We can clearly see the change in expression while pose and identity remain constant. 

For identity interpolation, we expect the expression and pose to stay the same as the input image $\bm{x}^i$ and only the  identity to change gradually from the identity of the input image to the identity of the target image $\bm{x}^t$.
Figure~\ref{idinter} shows the identity interpolation. We can clearly observe the change in identity while other variations remain limited. 

\section{Expression Transfer from Video}
We conducted another challenging experiment to test the potential of our method. Can we transfer facial expressions from an ``in-the-wild" video to a given template image (also ``in-the-wild" image)? For this experiment, we split the input video into frames and extract the expression component $\bm{z}_{exp}$ of each frame. Then we replace the expression component of the template image with the $\bm{z}_{exp}$ of the video frames and decode them. The decoded images form a new video sequence where the person in the template image has taken on the expression of the input video at each frame.
The result can be seen here: \url{https://youtu.be/tUTRSrY_ON8}. The original video is shown on the left side while the template image is shown on the right side. The result of the expression transfer is the 2nd video from the left. We compare against a baseline (3rd video from the left) where the template image has been warped to the landmarks of the input video. We can clearly see that our method is able to disentangle expression from pose and the change is only at the expression level. The baseline though is only able to transform expression and pose together. Our result video also displays expressions that are more natural to the person in the template image. To conclude, we are able to animate a template face using the disentangled facial expression components of a video sequence. 

\section{Relighting}
Figure~\ref{relight2} shows more relighting comparison results with~\cite{shu2017neural}. Here we compare directly with images provided by \cite{shu2017neural} in their paper.

\section{Further Expression Editing Comparison}
Figure~\ref{expcomp2} shows further expression editing comparison results with~\cite{wang2017learning}. The method proposed in \cite{wang2017learning} does not disentangle pose and hence requires a ``frontalisation" of the face to work optimally. Our proposed method on the other hand is able to edit expressions directly on aligned images. To visualise the difference, we run \cite{wang2017learning} directly on our aligned test images to compare with our proposed method. As expected the results returned by \cite{wang2017learning} does not perform well given this setup.
So given the same input (aligned images from CelebA) our proposed method is able to edit expression directly whereas \cite{wang2017learning} requires further ``frontalisation" transformations to obtain good results. This is due to \cite{wang2017learning} not being able to disentangle pose.

\end{document}